\documentclass[10pt]{article}

\usepackage{mathrsfs} 

\usepackage[]{graphicx}    
\usepackage[gen]{eurosym}
\usepackage{verbatim}
\usepackage{amsthm}
\usepackage{amsmath}
\usepackage{dsfont}
\usepackage{amssymb}
\usepackage{latexsym}		
\usepackage{tabularx}
\usepackage{setspace}
\usepackage{listings}
\usepackage{stmaryrd}
\usepackage{hhline}
\usepackage{booktabs}
\usepackage{caption}
\usepackage{subcaption}
\usepackage{bbm}
\usepackage{arxiv}

\usepackage[utf8]{inputenc} 
\usepackage[T1]{fontenc}    
\usepackage{hyperref}       
\usepackage{url}            
\usepackage{booktabs}       
\usepackage{amsfonts}       
\usepackage{nicefrac}       
\usepackage{microtype}      
\usepackage{natbib}
\usepackage{xcolor, colortbl}
\usepackage[capposition=top]{floatrow}
\usepackage{multirow,lscape,longtable}

\newtheorem{theorem}{Theorem}[]

\newtheorem{definition}{Definition}[]
\newtheorem{proposition}{Proposition}[]
\newcommand{\indep}{\perp \!\!\! \perp}

\usepackage{bm}

\title{AI and ethics in insurance: a new solution to mitigate proxy discrimination in risk modeling}

\author{
 Marguerite Saucé \\
  Diot-Siaci \\                         
  Paris, France\\
  \texttt{TOFILL} \\ 
   \And
  Antoine Chancel \\
  Data Analytics Solutions \\
  SCOR\\
  Paris, France\\
  \texttt{achancel@scor.com} \\
  \And
 Antoine Ly \\
  Data Analytics Solutions \\                         
  SCOR\\
  Paris, France \\
  \texttt{aly@scor.com} \\
}

\begin{document}
\maketitle

\begin{abstract}
The development of Machine Learning is experiencing growing interest from the general public, and in recent years there have been numerous press articles questioning its objectivity: racism, sexism, \dots  Driven by the growing attention of regulators on the ethical use of data in insurance, the actuarial community must rethink pricing and risk selection practices for fairer insurance. Equity is a philosophy concept that has many different definitions in every jurisdiction that influence each other without currently reaching consensus. In Europe, the Charter of Fundamental Rights defines guidelines on discrimination, and the use of sensitive personal data in algorithms is regulated. If the simple removal of the protected variables prevents any so-called `direct' discrimination, models are still able to `indirectly' discriminate between individuals thanks to latent interactions between variables, which bring better performance (and therefore a better quantification of risk, segmentation of prices, and so on). After introducing the key concepts related to discrimination, we illustrate the complexity of quantifying them. We then propose an innovative method, not yet met in the literature, to reduce the risks of indirect discrimination thanks to mathematical concepts of linear algebra. This technique is illustrated in a concrete case of risk selection in life insurance, demonstrating its simplicity of use and its promising performance.
\end{abstract}


{\Large\textbf{General introduction}}

\section{Motivation and scope of this paper}

\paragraph{} In many sectors, Machine Learning models have been exposed as unintentionally discriminatory, leading to unfair decisions that can have drastic consequences. The insurance industry has always been under close scrutiny when it comes to the use of personal data and discrimination issues, but in the light of recent denunciations in all industries, the attention on fairness issues has grown. \par
Fairness is a complex philosophical matter, and there is no single definition for it. Researchers have managed to define it statistically, but the definition depends on underlying assumptions and ideologies. Furthermore, as of today, many methods have been proposed to come closer to some definitions of fairness, but none can ensure perfect fairness. \par
Regarding previous points, we wonder to what extent fairness can be approximated when applying a Machine Learning model to insurance mortality data, and more specifically to what extent proxy discrimination can be avoided. This thesis aims at illustrating the complexity of quantifying discrimination and finding a way to mitigate it. For this purpose, we propose a simple and promising method that relies on linear algebra to mitigate indirect discrimination. \par
All along the thesis, we have provided the reader with summaries and key points in the form of light blue frames. \par


\newpage

\section{Understanding key concepts and regulations around discrimination}
\subsection{The life insurance industry}

\paragraph{} Historically, life insurers have used classic statistical models to assess risks for their products, covering Mortality, Critical Illness\footnote{Critical illness insurance compensates insureds with a lump sum payment upon diagnosis in order to cover treatment costs \cite{scor22}.}, Disability, Longevity and Medical Expenses. But with the development of Machine Learning and the now richer than ever data sources, new techniques are becoming more and more popular. These methods imply new challenges for actuaries, mostly due to the richness of information and the complexity of algorithms. \par

\subsubsection{Actuarial fairness}

\paragraph{} One of the challenges concerns the notion of equity, which has always been a key issue for insurers. `Actuarial fairness' means that risky insureds should contribute more and pay a higher premium. Actuaries have to determine how to classify policyholders between risky and non-risky, and more specifically what attributes are good indicators of risk. They then rely on historical data to estimate losses \cite{grari22}. It is sometimes complicated to know if an attribute is directly related to risk or not. As we will see later on, there are country-specific regulations concerning the use of certain attributes for risk assessment. \par

\subsubsection{Segmentation, pooling and adverse selection}
\label{segmentation}
\paragraph{} Before insurance, the only way of hedging against risk was individual prudence. Pooling offered a new way to deal with uncertainty: losses were the collective responsibility of the pool. This created insurance solidarity with an understanding of fairness \cite{frees21}. Today, insurers offer contracts at large levels, counting on the compensation between policyholders who file claims and those who do not. In order to ask for premiums accordingly to a policyholder's risk profile, insurers do a segmentation of the insureds \cite{charpentier22}. Segmentation consists in creating homogeneous classes and estimating the risk on average \cite{barry22}, so that premiums are adapted to the risk profile. \par
The pure premium is the expected loss of the insured over the coverage period. Since pooling is based on the law of large numbers, risks have to be homogeneous, which is why insurers need to classify the risks properly. The classification is based on observable factors, which should indicate what the risk is \cite{grari22}. \par
If groups are heterogeneous, policyholders could cross-subsidy. This leads to adverse selection: lower risks are asked to pay more than their expected loss because they are not classified in the right pool, so they are attracted by a competitor who will offer lower prices. \par

\begin{figure}[H]
    \centering
    \includegraphics[scale=0.6]{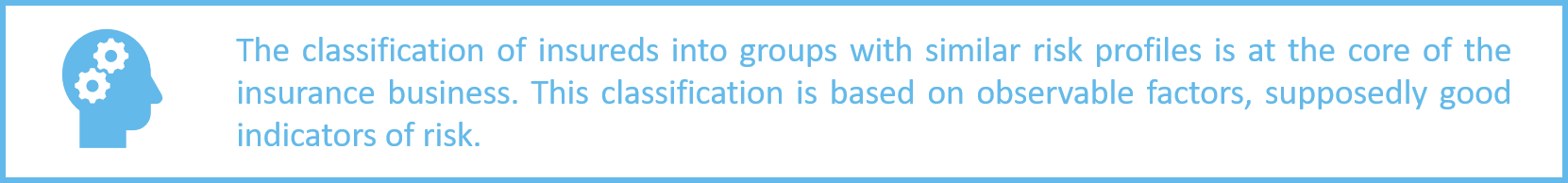}
\end{figure}



\subsubsection{Risk modeling}

\paragraph{} Life insurers model biometric risks, which relate to human life conditions, and more specifically the duration until the occurrence of an event. In our case, we will be studying the mortality of cancer patients. \par
Individuals with cancer history are considered `aggravated risks' and are often offered coverage with deterrent premiums, based on limited and imprecise criteria. But each cancer is specific and the evolution of treatment possibilities has tremendously increased survival odds in the past few years. At SCOR, the medical underwriting team is in charge of providing inclusive solutions for cancer patients. By using all available information about individuals, we can precisely estimate mortality rates thanks to Machine Learning models. These precise rates help insurers offer fair premiums to individuals with a history of cancer. \par
For cancer patients, insurers model the duration before death in the context of mortality or longevity products. Information about these individuals can also be used for critical illness products, to model the duration before the occurrence of a cancer. Once the R\&D department has studied the rates linked to the covered condition, the pricing team takes over to put the add-on cover or product on the market. The underwriters can also benefit from the study to better assess risk profiles. \par

\paragraph{}  To do this modeling, we need to take into account a few constraints: \label{constraints}
\begin{itemize}
    \item Underwriting constraints: the variables used by the model need to be available to the underwriter, i.e. be included in the medical file of the individual. Depending on the local legislation, models must not be discriminatory against certain population, so as a `solution', some variables are simply deleted. The insurer also has business constraints as he is in a competitive market.
    \item Medical constraints: the variables and the coefficients they are attributed need to be coherent with medical literature. For example, variables that are not medically relevant, such as the address, cannot be used by the model. Another example of variable coherence is if a larger tumor size implies a shorter life expectancy, it must be reflected by the model.
    \item Modeling constraints: the variables must be statistically relevant, not too numerous nor strongly correlated with each other.
\end{itemize}

\begin{figure}[H]
    \centering
    \includegraphics[scale=0.6]{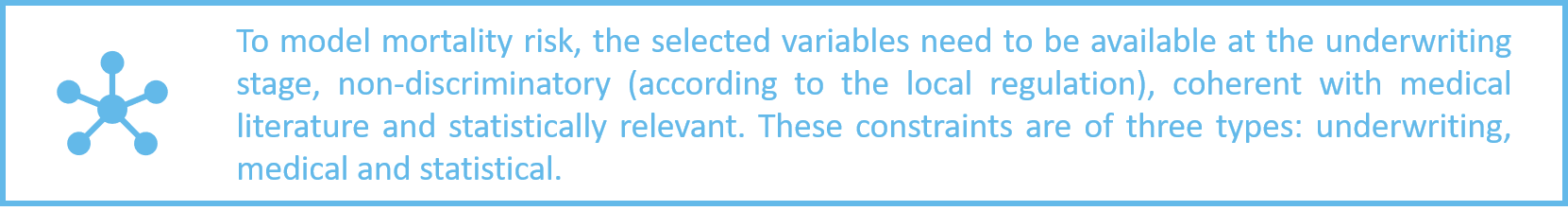}
\end{figure}

\newpage

\subsection{Fairness and bias}
\paragraph{} As we saw with the constraints of variable selection, depending on the local regulation, selected variables must not lead to discriminatory results. Fairness and bias is not a new subject in Machine Learning, but it has received increasing attention these past years, as numerous examples of unfair and biased outcomes were revealed in various fields. The most famous one is the COMPAS Recidivism Algorithm, which was proved biased against black defendants \cite{propublica}. As actuaries use high dimensional data and complex models, to price contracts for example, they need to check that the outcomes are not biased.  We need to define fairness and bias, see how it impacts insurance and find a method to detect it.

\subsubsection{The need for fairness in insurance}
\paragraph*{Reputation}
Fairness is a key issue in insurance, because actuaries need to explain to underwriters how their models work, so that in turn policy applicants can understand and trust the process to be fair. Insurance is not a well-seen sector in the public opinion and by the authorities in general, which is why fairness and transparency are crucial subjects.

\paragraph*{Regulation}
In the EU, the GDPR (General Data Protection Regulation) \label{gdpr} gives individuals the right to control their personal data, and specifically `the right not to be subject to a decision based solely on automated processing, including profiling, which produces legal effects concerning him or her or similarly significantly affects him or her' \cite{gdpr_article_22}. This means that a human intervention is required in any automated decision process. Individuals also have a right to erasure, or `right to be forgotten' \cite{gdpr_article_17}, which means that datasets might not be complete, independently of the collection process. \par
But there are specific regulations for the insurance sector. The French Supervisory Authority, the ACPR, requires appropriate data management, with ethical considerations such as fairness of processing and absence of discriminatory bias \cite{acpr_govAI}. \par
In April 2021, the EU proposed the first-ever legal framework on Artificial Intelligence (AI), the AI Act. The proposal is to define four levels of risk, from unacceptable to none, concerning AI systems. Each category will be subject to requirements and specific obligations such as conformity assessments and registration in a database. It could become applicable as soon as the second half of 2024 \cite{eu_ai_framework}. Insurers do not know how they will be impacted by this new regulation, but if their AI systems are classified as risky, they will be under strict regulation. \par
All in all, there is a regulation stacking that needs to be understood. The real question is: which regulation will be predominant?

\begin{figure}[H]
    \centering
    \includegraphics[scale=0.6]{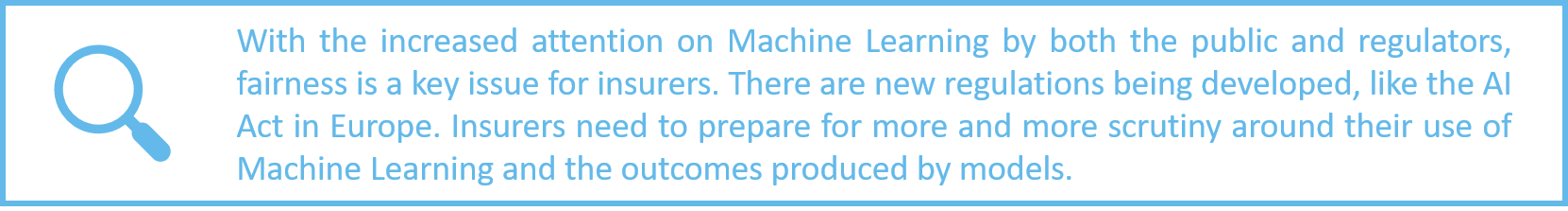}
\end{figure}

\subsubsection{A multitude of points of view}
\paragraph{}
Fairness is a relative notion that has many different definitions. A. Narayanan gave 21 definitions for fairness in classification tasks \cite{narayanan_yt}. It is a legal requirement, but also an ethical concept \cite{charpentier22}. Definitions can be categorized into individual or group fairness. Individual fairness aims to treat similar individuals similarly and group fairness treats different groups equally \cite{mehrabi19}. But group fairness might appear unfair at the individual level and a generalization based on group membership may be wrong \cite{binns19}. There are therefore two opposite worldviews regarding fairness with respect to a specific task: We're All Equal (WAE), meaning all groups have similar abilities, and What You See Is What You Get (WYSIWYG), meaning observations reflect similar abilities \cite{AIF360}. From this, many statistical definitions have been created to measure fairness for model outcomes, mostly for classification. We will study some of them in detail later on. \par
Machine Learning is a statistical discrimination by nature, but it becomes objectionable when there is a systematic advantage for some privileged group. All definitions of fairness cannot be reconciled and cannot satisfy all aspects. There needs to be a legal study to provide an official definition of fairness and an associated metric. \par

\begin{figure}[H]
    \centering
    \includegraphics[scale=0.6]{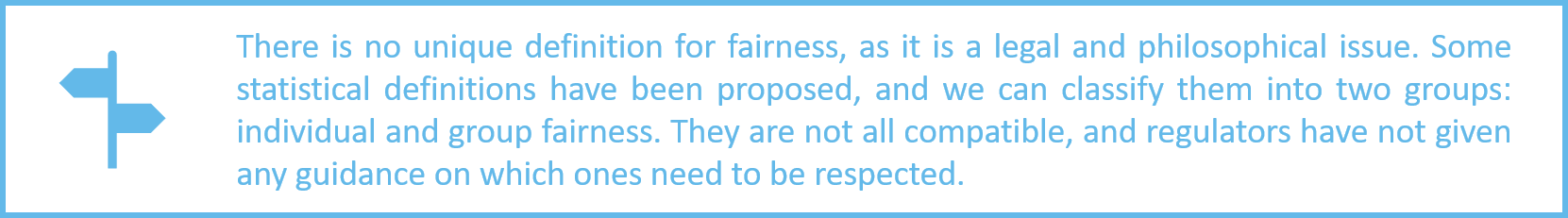}
\end{figure}

\subsubsection{The different types of bias}
\paragraph{} Statistical fairness is strongly related to bias, which can be defined differently depending on the sector. In statistics, it is a systematic error in prediction outcomes \cite{AIF360}. For Machine Learning models, it can be introduced by users, come from the data or be amplified by algorithms. And it is a vicious circle, as algorithms learning from biased data give biased outcomes which will be fed into and amplified by future algorithms \cite{mehrabi19}. \par
There are three types of biases that appear in classification problems \cite{barry22}:
\begin{itemize}
    \item Type 1: classes do not reflect the reality of the risk,
    \item Type 2: classes reflect a correlation with risk that is non-causal,
    \item Type 3: classes reflect a causal statistical reality, but are unacceptable because of ethical reasons.
\end{itemize}

\begin{figure}[H]
    \centering
    \includegraphics[scale=0.6]{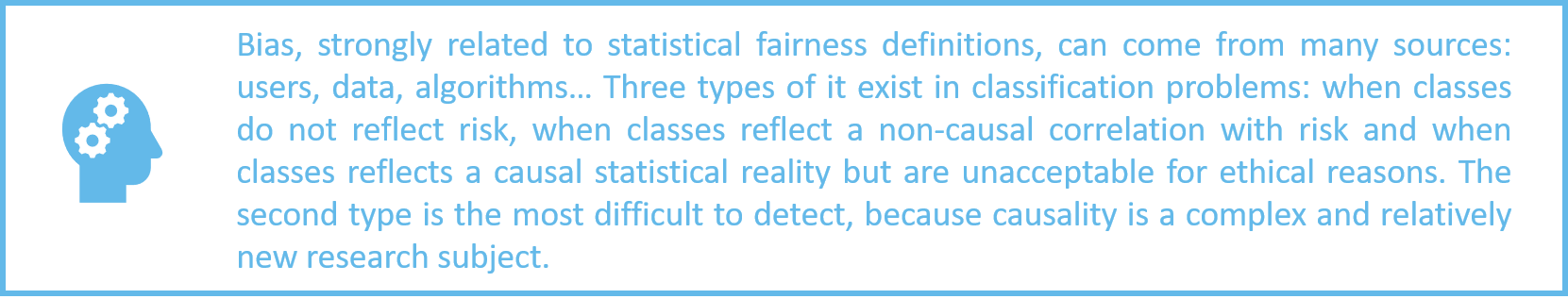}
\end{figure}

\subsubsection{Unfair and fair discrimination}
\paragraph{} To better understand and scope the concept from a scientific point of view, we discussed with different lawyers specialized in data law. Those exchanges helped us focus on the different key concepts exposed below.

\paragraph{Legal definition} Discrimination is the difference in treatment between individuals in similar situations due to prohibited criteria. In France, the Penal Code \cite{penal_code} defines these criteria in the Article 225-1: \label{anti-discrim_law} \begin{quote} `their origin, their sex, their family situation, their pregnancy, their physical appearance, the particular vulnerability resulting from their economic situation, apparent or known to its author, their surname, their place of residence, their state of health, their loss of autonomy, their disability, their genetic characteristics, their morals, their sexual orientation, their gender identity, their age, their political opinions, their trade union activities, their ability to express themselves in a language other than French, their membership or non-membership, real or supposed, of an ethnic group, a nation, a so-called race or a determined religion' \end{quote} These general prohibitions to fight against discrimination are supplemented by the French Insurance Code, and the Article L117-2 states that there can be absolutely no distinction between individuals based on:
\begin{itemize}
    \item age for access to insurance guarantees or termination of insurance benefits, with the exception of pricing for life insurance contracts with mortality tables;
    \item pregnancy and motherhood for premium and benefit computation;
    \item sex for premium and benefit computation, except for mandatory supplementary pension schemes.
\end{itemize}
The Article 16-13 of the Civil Code \cite{civil_code} defines an absolutely prohibited criteria that is applicable to insurance: \begin{quote} `No one may be discriminated against because of their genetic characteristics.' \end{quote} \par
Other criteria can be used if they are a justified business necessity, or to modulate premiums and guarantees.\par

\paragraph{The nature of insurance} As we saw in section~\ref{segmentation}, insurance is based on segmentation and pooling. It is about treating different risks differently. Classification and regression tasks are by definition a form of discrimination: the aim is to distinguish individuals based on a statistical similarity \cite{barocas16}. This is a form of discrimination that is justified and deemed acceptable if it does not systematically put a protected group at a disadvantage. \par

\paragraph{Unfair discrimination} Discrimination is unfair when a certain group is treated unequally based solely on their affiliation to it \cite{kamiran11}. Direct discrimination happens when protected attributes are explicitly used to make the decision. Indirect discrimination happens when the treatment appears neutral and depends on non-protected attributes, but protected groups get treated unjustly \cite{mehrabi19}. Harmful discrimination in insurance can happen at several stages: for the decision to insure, during the underwriting or marketing phases, to renew or cancel policies, for coverage offer or for pricing \cite{frees21}.

\begin{figure}[H]
    \centering
    \includegraphics[scale=0.6]{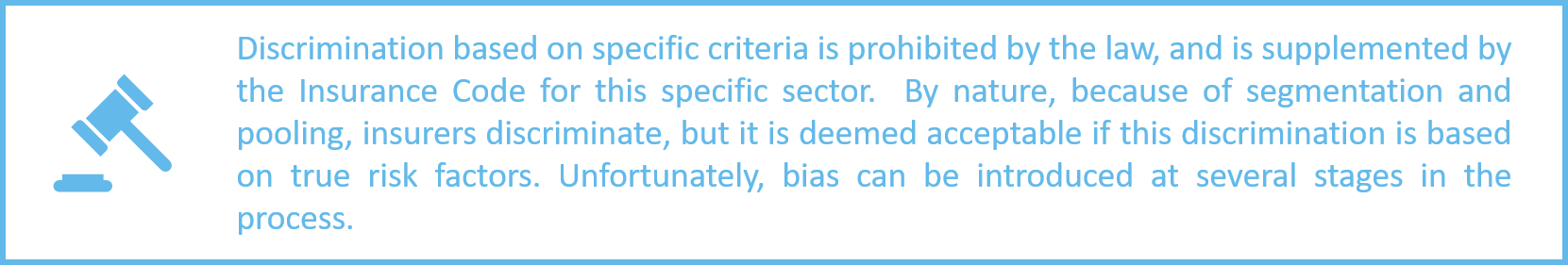}
\end{figure}

\subsubsection{Interpretability: a first step to tackling bias}
\paragraph{} Interpretability is the degree to which a human can understand the cause of a decision. Some models are directly interpretable and others, qualified as `black box' models, need interpretability methods. As the need to explain Machine Learning models become more and more important, a new research field XAI, for eXplainable Artificial Intelligence, was created. Understanding why a model predicted specific outcomes is a first step in bias detection \cite{molnar}.

\paragraph{Interpretable models} Some models are directly interpretable because of their structure. Linear Regressions predict the target as a weighted sum of the feature inputs, but assume linear relationships. Generalized Additive Models take into account non linear effects. Decision trees and decision rules are easy to interpret and capture feature interaction. \par

\paragraph{Model-agnostic explanation methods} Global methods explain how features affect the prediction on average. Techniques include feature interaction detection, prediction function decomposition, feature importance measure and representative data points choice. \par
Local methods explain the individual predictions. Techniques include the description of how changing a feature changes the prediction, of which features anchor a prediction, of which features would need to be changed to change a prediction and of which  individual features are attributed to a prediction.

\paragraph{Detecting bias} Having more information on which variables play an important role in predictions, which variables interact with each other and more generally what causes a prediction can help detect bias. The outcome might be biased if a protected attribute has a great importance in the prediction or if an important variable for the prediction has a strong correlation with a protected attribute. This can help find out where the bias actually comes from.

\begin{figure}[H]
    \centering
    \includegraphics[scale=0.6]{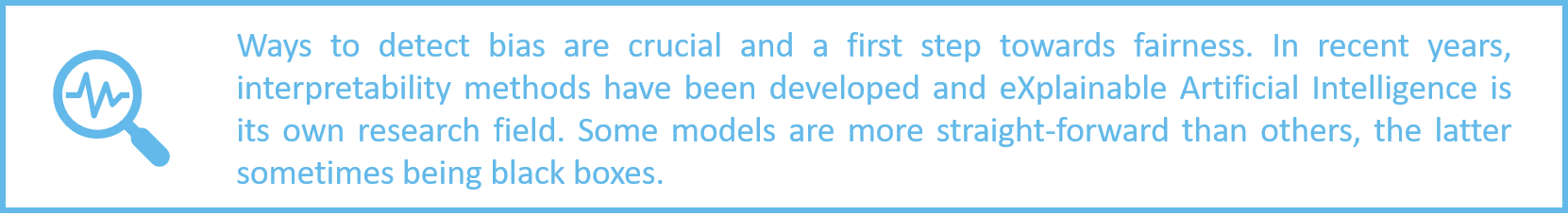}
\end{figure}

\newpage

\subsection{The expression of fairness and bias in data}
\paragraph{} Feature selection is the process of choosing what attributes are observed and taken into account for analysis. They are necessarily a reduction of reality and fail to capture real-world phenomena \cite{barocas16}. But obtaining sufficiently rich information is not always possible, and insurers have to rely on reductive data.

\begin{figure}[H]
    \centering
    \includegraphics[scale=0.6]{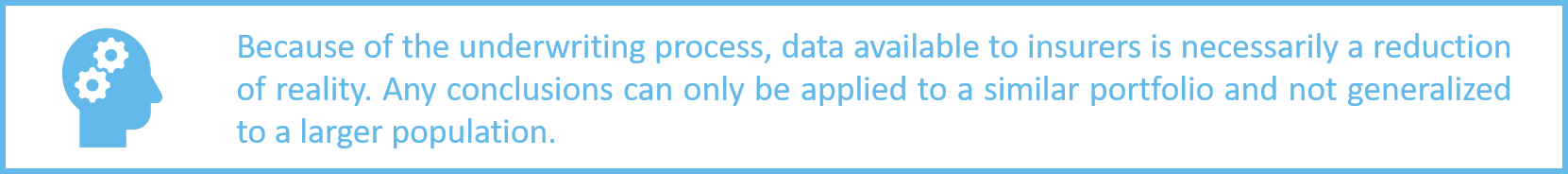}
\end{figure}

\subsubsection{Sensitive variables}
\paragraph{} Insurers are not allowed to use all variables that are available to them, because they are viewed as protected or sensitive, and using them might lead to biased outcomes. The choice of which variables are allowed depends on regulators, but also on society as a whole, because it is both a legal and ethical concern. Often, attributes that are not under the control of individuals are not accepted. Attributes that change over time are accepted, because individuals are on both sides at different stages in their life. Acceptable variables should be good predictors of risk.  If attributes are known to cause a risky event they are accepted \cite{frees21}.

\paragraph{\underline{Regulation}}


\paragraph{EU} The European Union has anti-discrimination legislation, in the Charter of Fundamental Rights, that applies to the insurance sector, but it can have the effect of restricting flexibility of risk management and raising costs and legal insecurity, which means that insurers offer less effective coverage. Lawmakers decide which factors are determining for risk assessment but they are not specialists. With these directives, insurers cannot perfectly prevent adverse selection and moral hazard, and as a consequence, consumers can end up penalized, paying higher premiums and deductibles \cite{petkantchin10}.

\paragraph{France} There is an evolving conflict between insurance and anti-discrimination standards: on the one hand, insurers classify risks and on the other, the law prohibits the differentiation of individuals based on criteria that deny equal dignity. Insurers can select risks as long as they demonstrate the objectivity and statistical foundations of the data they rely on to do so. The Insurance Code completely forbids discrimination based on pregnancy and motherhood, risk selection for supplementary pension and the use of genetic information \cite{robineau10}. In 2011, the EU court ruled that insurers offering different prices to men and women violated gender equality laws, and this affected car, term life, health insurance and annuities \cite{test-achats}. This means that there is no possible discrimination on these criteria, even though there are some technical and pragmatic arguments for using them.

\paragraph{US} Recently, unfair discrimination has become closely connected to disparate impact (see section~\ref{DI}). It is a measure of how a practice affects a group more than another, even when it appears neutral. But as of 2009, no court had applied the disparate impact standard to evaluate insurance rates \cite{miller09}. Unfair discrimination in insurance is indeed not exactly equivalent to disparate impact. In State law, unfair discrimination happens when similar risks are treated differently for determining rates, coverage, benefits and terms and conditions of policies. Depending on States, certain factors are prohibited for risk classifications and in underwriting decisions. But laws are not as restrictive as one could believe: in fifteen States, it is only prohibited to use race as the only factor regarding a decision to issue or continue a policy and in four States there are no restrictions on the use of race for underwriting personal automobile insurance \cite{stead20}. \par
In July 2021, the governor of Colorado signed a Senate Bill on insurers' use of external consumer data \cite{senatebill}. Insurers in the State are prohibited from unfairly discriminating on several variables. Unfair discrimination is defined as including \begin{quote} `the use of one or more external consumer data and information sources, as well as algorithms or predictive models using external consumer data and information sources, that have a correlation to race, color, national or ethnic origin, religion, sex, sexual orientation, disability, gender identity, or gender expression, and that use results in a \textbf{disproportionately negative outcome} for such classification or classifications, which negative outcome exceeds the reasonable correlation to the underlying insurance practice, including losses and costs for underwriting.' \end{quote} This means that insurers using external consumer data in Colorado will need to provide a disparate impact analysis \cite{krafcheck21}, even though it is not a synonym for unfair discrimination. The following question is: will other States follow in the same direction as Colorado? \par

\paragraph{\underline{What are they?}}
\label{protected_variables}
\paragraph{Forbidden variables} in France are pregnancy, motherhood and genetic information in all processes. For car, term life, health insurance and annuities, gender is a forbidden variable. In the US, there are some rules that come from federal law: insurers cannot consider pre-existing health conditions or gender in the underwriting process, genetic information in coverage availability or premium charging, or housing practices that have a disparate impact on protected classes. There are no other federal laws regulating what criteria can be taken into account. Historically, States are responsible for regulating insurance discrimination. These regulations strongly depend on which State and insurance line are in question: nine States completely prohibit the use of race and national origin in all lines, 7 States religion, one State gender and five States sexual orientation. Louisiana explicitly allows the use of race for life insurance. No State completely bans the use of age, credit score, genetic testing or ZIP Code \cite{avraham13}.

\paragraph{Sensitive variables} in France are those defined in the anti-discrimination law that are not forbidden, as we saw in section~\ref{anti-discrim_law}. They can only be used if statistical data proves their relevance and objectivity for risk analysis. Information on ethnicity, religion, sex, gender, sexual orientation, disability, and age can be viewed as sensitive. Less obvious sensitive variables include parenthood, military service, political party, socioeconomic status, or involvement in the criminal justice system. \\

\begin{figure}[H]
    \centering
    \includegraphics[scale=0.6]{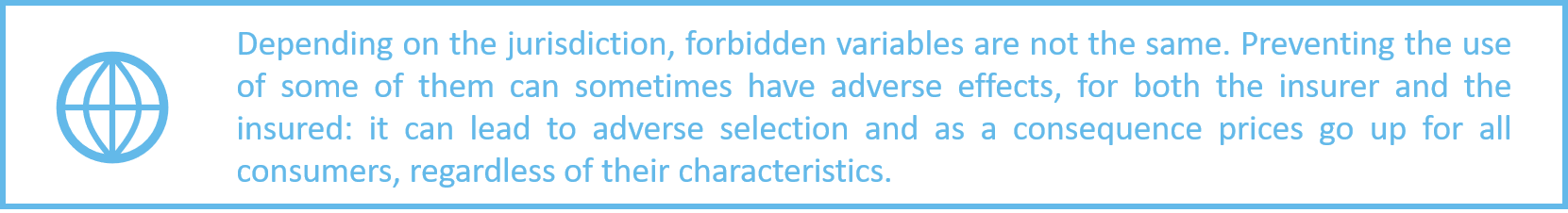}
\end{figure}

\subsubsection{Proxy variables}
\paragraph{} Proxies are unprotected variables that are strongly correlated to protected variables but also contain strong predictive information. Using proxy variables may result in indirect discrimination \cite{barocas16}. Insurers seek good segmentation of risk and profit maximization, but this might lead to perpetuating inequalities in society if outcomes are biased because of the variables that are used for prediction.

\paragraph{Name and surname} Accuracy of guessing the national origin of names varies significantly by the individual perception of national origin. Numerous characteristics matter, such as gender, popularity and the average level of educations of mothers who gave that name \cite{gaddis17}. A name depends on the culture of group, trends, the social level and time. With a name, you could identify the sex and the national origin of a person. You could also estimate the average age of people who have that name based on trends and popularity, but it would be difficult to infer the exact age \cite{charpentier22}.

\paragraph{Address} In the US, Black and Hispanic segregation and spacial isolation is still very active in some metropolitan areas \cite{rugh14}. This means that addresses are good proxies for ethnicity, and classifiers using this variable will exhibit discriminatory behavior \cite{kamiran19a}. The term redlining comes from the 1930s when residential security maps were created to indicate which parts f a city were safe to invest in: neighborhoods outlined in red were the riskiest \cite{martin17}. From the 1960s, in the context of the struggle for Black Civil Rights, the use of redlining for risk classification was strongly criticized. The address is a non-causal variable, but it is strongly correlated with non-observable risk factors and with ethnicity \cite{barry22}. \par
There is a strong correlation between income and ethnicity and between income inequality and income segregation in the US. This is partly due to housing discrimination after World War II, forcing Black families with lower incomes to live in proximity in urban areas \cite{reardon11}. Addresses are therefore strongly correlated with income, which is strongly correlated with ethnicity. \par
Night lighting and wealth are correlated, and it is possible to approximately estimate how wealthy a neighborhood is from satellite imagery with a strong predictive power \cite{jean16}. Address (and its associated satellite images) is a proxy for wealth. \par
With Google Street View, it is also possible to use the number and type of cars to infer wealth, ethnicity, education level and political preferences. It is also easy to detect the presence of handicap access ramps \cite{hara14} or a flag indicating national origins, political preferences or sexual orientations \cite{charpentier22}. \par
 
\paragraph{Occupation} Despite efforts for parity in the workplace, numerous occupations are still dominated by one of the two genders. In 2016 in a representative French region, sectors such as social working, healthcare and teaching are mostly feminine and industry, construction and transportation are mostly masculine. Another visible trend is that very feminine sectors tend to become even more so \cite{faure20}.\par
In 2018 in France, 18\% of employees worked part-time, 78\% of which were women \cite{dares20}. Knowing an individual works part-time means it is three times more likely that it is a woman.\par

\paragraph{Credit score} This variable cannot be measured directly, as it comes from the problem definition of creditworthiness. It is a non-arbitrary definition, not a given. The definition process can already itself be biased \cite{barocas16}, as it relies on measurable attributes that are available. The choice of which variables to use can introduce discrimination. Credit scores also create a vicious circle in terms of poverty, and using this variable introduces a disparate impact on racial minorities and low-income households, who then have to pay a higher premium \cite{charpentier22}.

\paragraph{Face} With the boom of facial recognition softwares, there are numerous opportunities of application. It is now possible to use facial recognition tools for health assessment by using measurements and proportions of facial attributes. They may be considered biometric data, so there are ethical issues surrounding their use \cite{boczar21}. Facial recognition can accurately predict gender and ethnicity, which raises moral questions. \par

\paragraph{Speech} The way an individual talks can indicate origin if he or she has difficulties with pronunciation, has a strong accent or speaks in a dialect. Linguistic profiling is the identification of an individual's ethnicity based on how their voice sounds and using that information for discrimination \cite{squires06}.\par
Current Natural Language Processing tools are trained on traditional written sources, which are different from spoken language, and even more from dialectal spoken language. The latter are more likely to be incorrectly classified, so bias can arise, with an incorrect representation of ideas and opinions from minority groups \cite{blodgett17}.\par
Chatbots are rule-based, information retrieval or learning-based systems that are widely used today. In March 2016, a Microsoft chatbot was supposed to improve its small-talk capabilities by learning from conversations with human users. In less than a day, it was displaying racist and sexist abusive content \cite{schlesinger18}. This shows that technical difficulties must be tackled in order to avoid such outcomes, especially when black box algorithms are used for treating voices.\par
In insurance, writing or speaking chatbots are used to report insurance claims. For example, Izzy Constat is a tool for amicable reports after a car accident. The chatbot asks for drivers' identities and context and generates a sketch representing the incident \cite{calvo22}. What if this chatbot were biased against a group that had a specific dialect? This could result in understating the severity of impact and lower benefits.\par

\paragraph{Network} Who you know either gets you access to resources or makes you guilty by association. Recommendation systems are based on similarity between individuals, and if insurers had access to this kind of information, they could find customers and limit their financial risks. But this kind of practice may not be ethical, as people who are already marginalized can be even more affected \cite{boyd14}. 


\paragraph{} There are many other variables with more or less predictive power that could be used as proxies for protected attributes. We will not be able to list them all, but the conclusion is that a study of their meaning and how they are linked to protected attributes is essential before using them as inputs in a prediction model.

\begin{figure}[H]
    \centering
    \includegraphics[scale=0.6]{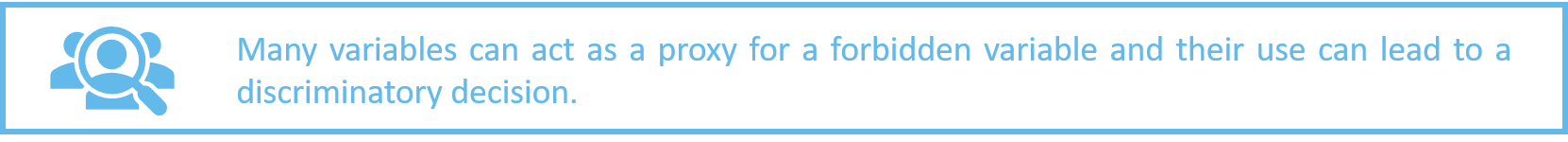}
\end{figure}







\newpage

\subsection{How to measure fairness}
\label{fairnessmetrics}
\paragraph{} Numerous metrics have been created to measure bias and fairness, the two notions not being distinct in most works.
The general setting is the following:
\begin{itemize}
    \item $X \in \mathcal{X}\subset\mathbb{R}^{n}$ is the set of n non-protected variables,
    \item $A \in \mathcal{A}$ is the protected or sensitive variable,
    \item $Y \in \mathcal{Y}$ is the true outcome,
    \item $f$ is the predictor, classifier or regressor,
    \item $\hat{Y}=f(X,A) \in \mathcal{Y}$ is the predicted outcome.
\end{itemize}

\subsubsection{Binary classification}
\label{binary_classif}
\paragraph{} In the binary classification setting, $\mathcal{Y} = \{y^-,y^+\}$0. $Y=y^-$ means the outcome is negative and $Y=y^+$ means the outcome is positive. In this problem, it is possible to compute the confusion matrix between the true and predicted outcomes for each protected group, see table \ref{table:confusion_matrix}.

\begin{table}[H]
    \centering
    \begin{tabular}{l|l|c|c|}
        \multicolumn{2}{c}{} & \multicolumn{2}{c}{Predicted outcome} \\
        \cline{3-4}
        \multicolumn{2}{c|}{} & Positive & Negative \\
        \cline{2-4}
        \multirow{2}{*}{True outcome} & Positive & $TP$ & $FN$ \\
        \cline{2-4}
        & Negative & $FP$ & $TN$ \\
        \cline{2-4}
    \end{tabular}
    \captionsetup{justification=centering}
    \caption{Confusion matrix \\ TP: True Positive, FN: False Negative, FP: False Positive, TN: True Negative}
    \label{table:confusion_matrix}
\end{table}

The most common metrics in the literature are the following \cite{alves22}. \par

\paragraph{Statistical parity}\label{stat_parity} (or demographic parity) requires the likelihood of a positive outcome to be the same for all protected groups i.e. $\hat{Y} \indep A$:
\begin{equation*}
    \forall a \in \mathcal{A}, \mathbb{P}(\hat{Y}=y^+|A=a)=p
\end{equation*}
This is equivalent to having the same predicted acceptance rates AR for all protected groups:
\begin{equation*}
    \mbox{AR}=\frac{\mbox{TP}+\mbox{FP}}{\mbox{TP}+\mbox{TN}+\mbox{FP}+\mbox{FN}}
\end{equation*}

\paragraph{Equalized odds} requires all protected groups to have the same probabilities of being correctly assigned a positive outcome and of being incorrectly assigned a positive outcome i.e. to have the same true and false positive rates i.e. $\hat{Y} \indep A | Y$:
\begin{equation*}
    \forall (y,a) \in \mathcal{Y}\times\mathcal{A}, \mathbb{P}(\hat{Y}=y^+|Y=y,A=a)=p
\end{equation*}
This is equivalent to having the same true positive rates TPR and false positive rates FPR for all protected groups:
\begin{equation*}
    \begin{split}
        \mbox{TPR} & =\frac{\mbox{TP}}{\mbox{TP}+\mbox{FN}} \\
        \mbox{FPR} & =\frac{\mbox{FP}}{\mbox{FP}+\mbox{TN}}
    \end{split}
\end{equation*}
Remark: the true positive rate is also called recall or sensitivity.

\paragraph{Equal opportunity} is the same as equalized odds but only requires all protected groups to have the same probability of being correctly assigned a positive outcome i.e. requires the same true positive rates TPR for all protected groups:
\begin{equation*}
    \forall a \in \mathcal{A}, \mathbb{P}(\hat{Y}=y^+|Y=y^+,A=a)=p
\end{equation*}

\paragraph{Disparate Impact} \label{DI} is a popular metric in the US to measure bias. It is defined as the ratio in probability of favorable outcomes between groups $A=0$ and $A=1$ \cite{AIF360}:
\begin{equation*}
    \mbox{DI} = \frac{\mathbb{P}(\hat{Y}=1|A=0)}{\mathbb{P}(\hat{Y}=1|A=1)}
\end{equation*}
It is a consequence of the statistical parity definition, in the case where the probabilities are non-null. A Disparate Impact of 1 would mean that the model is fair, lower than 1 that the model is unfair to group $A=0$ and above 1 that the model is unfair to group $A=1$0. This Disparate Impact is only defined in the case of a binary classification with a binary protected variable. Its estimation is not as easy as we could think, because of its definition as a ratio: we can have robust estimators for both probabilities, but the estimator of a ratio is not the ratio of estimators. This is why we decided not to use this fairness metric and to keep the original definition of statistical parity which is not as restrictive. \par

\begin{figure}[H]
    \centering
    \includegraphics[scale=0.6]{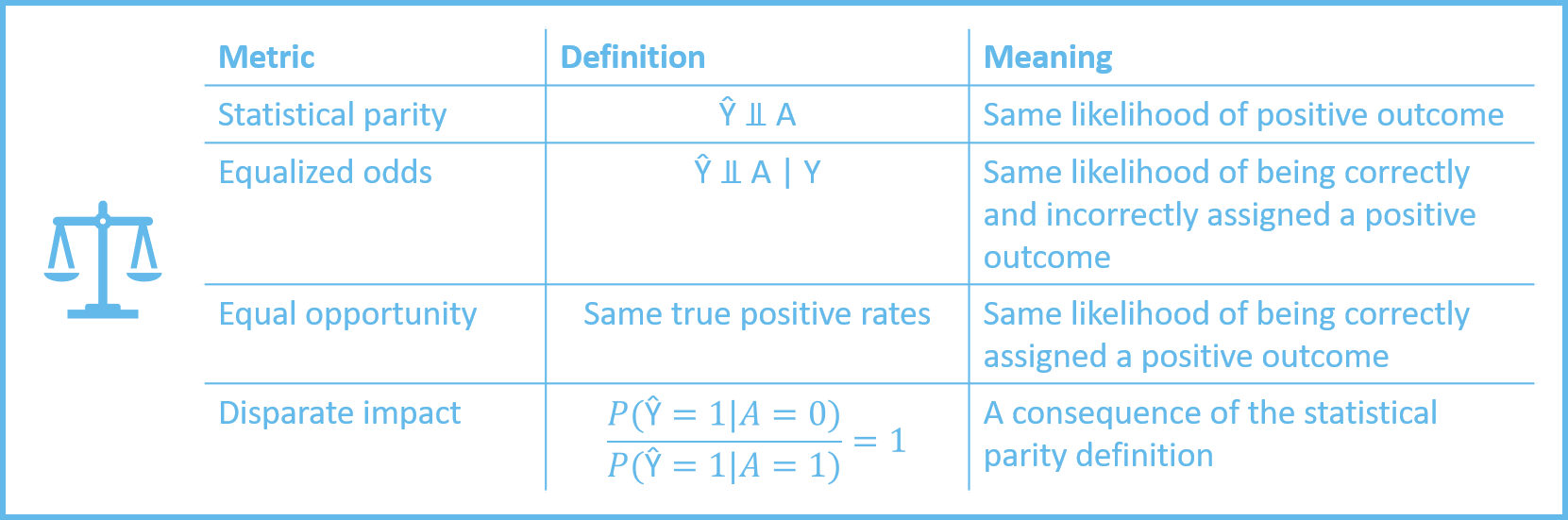}
\end{figure}

\begin{figure}[H]
    \centering
    \includegraphics[scale=0.6]{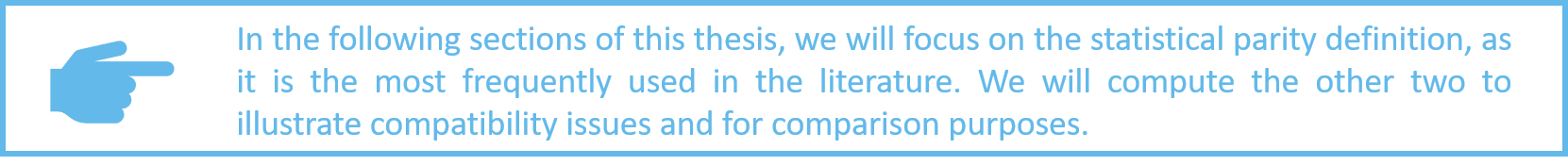}
\end{figure}

\subsubsection{Extension of the metrics to other settings}
\paragraph{} Most of these definitions can be extended to a multi-class classification ($\mathcal{Y}\subset\mathbb{N}$) or a regression ($\mathcal{Y}\subset\mathbb{R}$) setting: by defining a subset $\mathcal{Y}^+\subset\mathcal{Y}$ (respectively $\mathcal{Y}^-\subset\mathcal{Y}$) of values reflecting a positive (respectively negative) outcome, we adapt the definitions from the previous section:
\begin{equation*}
    \begin{split}
         & \mbox{Statistical parity: } \forall a \in \mathcal{A}, \mathbb{P}(\hat{Y}\in\mathcal{Y}^+|A=a)=p \\
         & \mbox{Equalized odds: } \forall a \in \mathcal{A}, \mathbb{P}(\hat{Y}\in\mathcal{Y}^+|Y\in\mathcal{Y}^+,A=a)=\mathbb{P}(\hat{Y}\in\mathcal{Y}^+|Y\in\mathcal{Y}^-,A=a)=p \\
         & \mbox{Equal opportunity: } \forall a \in \mathcal{A}, \mathbb{P}(\hat{Y}\in\mathcal{Y}^+|Y\in\mathcal{Y}^+,A=a)=p
    \end{split}
\end{equation*}
This supposes that we can categorize every value of output as either positive or negative.


\subsubsection{Model evaluation}
\paragraph{} We will compare models using the confusion matrix and metrics deriving from it:
\begin{itemize}
    \item The accuracy is the proportion of correct classifications among all classifications.
    \begin{equation*}
        \mbox{Accuracy}=\frac{\mbox{TP}+\mbox{TN}}{\mbox{TP}+\mbox{TN}+\mbox{FP}+\mbox{FN}}
    \end{equation*}
    \item The true positive rate is the proportion of correct positive classifications among actual positive values.
    \begin{equation*}
        \mbox{TPR}=\frac{\mbox{TP}}{\mbox{TP}+\mbox{FN}}
    \end{equation*}
    \item The false positive rate is the proportion of wrong positive classifications among actual negative values. It is sometimes called the probability of false alarm.
    \begin{equation*}
        \mbox{FPR}=\frac{\mbox{FP}}{\mbox{FP}+\mbox{TN}}
    \end{equation*}
    \item The acceptance rate is the proportion of positive classifications among all classifications.
    \begin{equation*}
        \mbox{Acceptance rate}=\frac{\mbox{TP}+\mbox{FP}}{\mbox{TP}+\mbox{TN}+\mbox{FP}+\mbox{FN}}
    \end{equation*}
\end{itemize}
In order to evaluate a model, we need to take into account the facts that:
\begin{itemize}
    \item An acceptable accuracy threshold depends on the context of the prediction: do we need to be perfectly accurate in order for the decision to be accepted by society? Do mistakes cost a lot to the company? Furthermore, if there is a large class imbalance, accuracy is not the best metric as it can be very high while the model only fits the majority population.
    
\begin{figure}[H]
    \centering
    \includegraphics[scale=0.6]{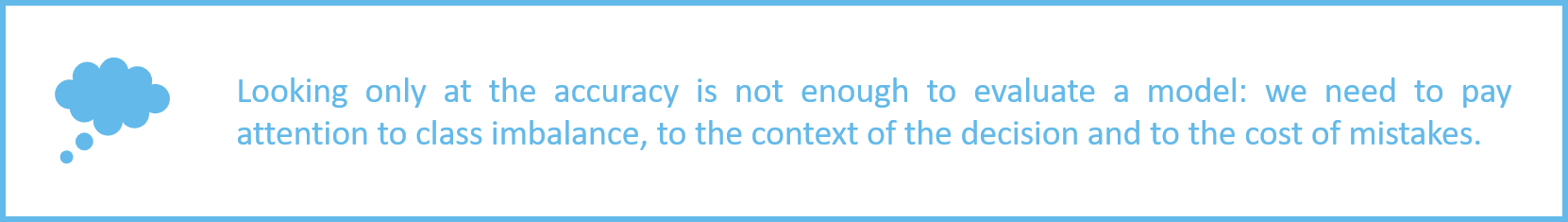}
\end{figure}

    \item Missclassification errors are more or less acceptable depending on the context. For example, in a risk selection decision, underwriters want to select `good' risks. It is generally more acceptable and less costly to falsely reject good risks than falsely accept bad risks.
    
\begin{figure}[H]
    \centering
    \includegraphics[scale=0.6]{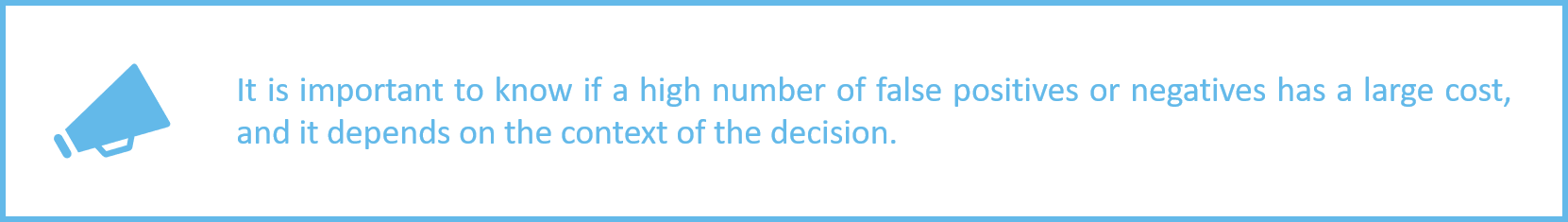}
\end{figure}
\end{itemize}

{\Large\textbf{Simulated data}}
\section{Simulated data}
\paragraph{} We will apply discrimination detection and correction methods to simulated data before tackling a real use case. The reason for this is that we wish to know the answer to the question `Is my prediction discriminatory towards a group?' while knowing how the outcome was computed, so that we can look for a solution to a well-defined problem, instead of having to make assumptions. \par
We will begin by generating our data, which consists of two sensitive variables $A$ and $B$, a set of non-protected variables $X = \{X_i\}_{i=3,...n}$, $n \in \mathbb{N}$ and a variable of interest $Y$0. All variables can be correlated with each other, but depending on the country and its regulation, insurers are not always allowed to use the sensitive variables as explanatory variables, and sometimes, with the GDPR for example, cannot even collect the information. In this section, however, we will suppose that the variable is available, because it is the only way to measure discrimination. \par
To link with a real-life example, if we were in a pricing context for automobile insurance, $A$ could represent gender, $B$ marital status, the $X_i$ other variables such as age or car value and $Y$ the claim occurrence. Gender is not the cause of an accident, but statistically we observe that gender and claim occurrence are correlated. Intuitively, we should not discriminate based on gender, because it would be unfair as it is a stereotype, but we do not have access to a `fairer' variable, which could be driving behavior. \par

\begin{figure}[H]
    \centering
    \includegraphics[scale=0.6]{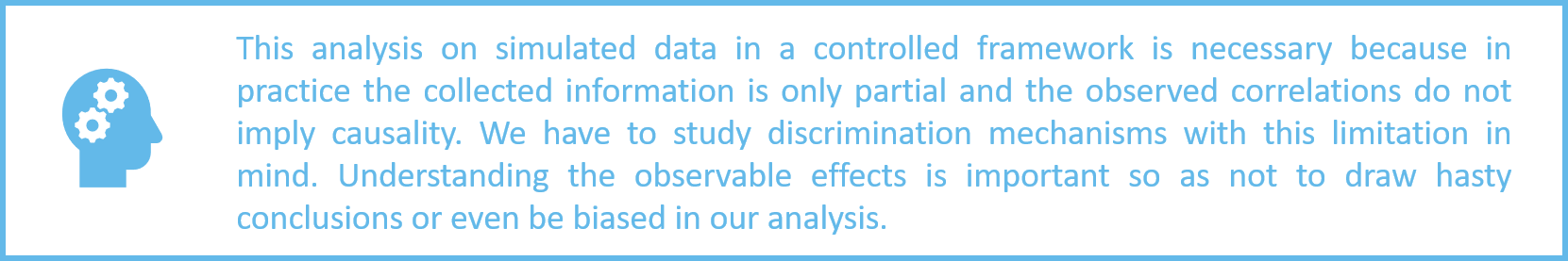}
\end{figure}

For simplification reasons, we will suppose that the protected variables and the outcome are binary. We will begin by giving the theoretical framework, then illustrate the simulation process with two variables, and finally create the dataset. \par

\subsection{A reminder on statistical tools used to set the framework}
\paragraph{} In this section, we will pose important mathematical concepts that will help understand the observed effects on the fairness metric. \par
We want to generate the variables while controlling the relationship they have with each other. To do so, we will use the theory of copulas. \par

\begin{figure}[H]
    \centering
    \includegraphics[scale=0.6]{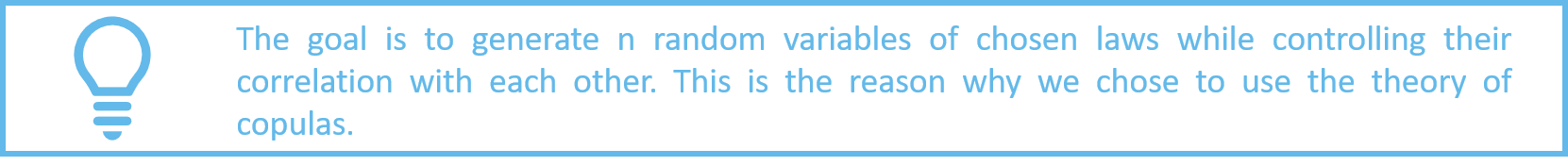}
\end{figure}

\vspace{5mm}
\noindent \underline{\textbf{Notations}} \par
\vspace{5mm}
\noindent First, we need to introduce some notations:
\begin{itemize}
    \item[-] $f$ is a probability density function and $F$ the associated cumulative distribution function: $F'=f$
    \item[-] $\mathbb{E}$ is the expected value and $Var$ the variance
    \item[-] $\Sigma$ is the covariance matrix: $\Sigma_{i,j}=cov(X_i,X_j)$
    \item[-] $R$ is the correlation matrix: $R_{i,j}=corr(X_i,X_j)$
    \item[-] $\phi$ is the probability density function and $\Phi$ the cumulative distribution function of a standard normal distribution $\mathcal{N}(0,1)$
    \item[-] $\phi_R$ is the probability density function and $\Phi_R$ the cumulative distribution function of a standard multivariate normal distribution $\mathcal{N}_n(0,R)$
\end{itemize}

\vspace{5mm}
\noindent \underline{\textbf{Probabilities}} \cite{charpentier10}
\begin{definition} A $X=(X_1,\dots,X_d)^T$ d-dimensional real random vector is a standard normal random vector if all of its components $X_i, i=1,\dots,d$ are independent and follow a standard normal distribution. \end{definition}

\begin{definition} A $X=(X_1,\dots,X_d)^T$ d-dimensional real random vector follows a multivariate normal distribution if there exists a random k-vector $Z$, which is a standard normal random vector, a d-vector $\mu$ and a $k\times d$ matrix $A$ of full rank such that $X=AZ+\mu$0. \\
We denote $X \sim \mathcal{N}_d(\mu,\Sigma)$ where $\mu=\mathbb{E}[X]=(\mathbb{E}[X_1],\dots,\mathbb{E}[X_d])^T$ is the mean vector and $\Sigma=AA^T$ is the covariance matrix. \end{definition}

\begin{definition} \label{def:multiv normal} The multivariate normal distribution is non-degenerate when the covariance matrix $\Sigma$ is positive definite, in which case it has the following density
\begin{equation*}
    f_{X}(x_1,\dots,x_d)=\frac{exp(-\frac{1}{2}(x-\mu)^T\Sigma^{-1}(x-\mu))}{\sqrt{(2\pi)^k|\Sigma|}}
\end{equation*}
where $x=(x_1,\dots,x_d)^T \in \mathbb{R}^d$ and $|\Sigma|=det(\Sigma)$ is the determinant of $\Sigma$0.
\end{definition}

\begin{proposition} \label{prop:sim proc} For any univariate cumulative distribution function $F$, 
\begin{equation*}
    U \sim \mathcal{U}([0,1]) \Rightarrow F^{-1}(U):=X \sim F
\end{equation*}
\end{proposition}

\begin{definition} Set $X$ and $Y$ two continuous random variables. $X$ and $Y$ are independent if and only if $F_{X,Y}(x,y)=F_X(x)F_Y(y)$0. \end{definition}

\begin{definition} \label{def:pearson corr} Set $X$ and $Y$ two continuous random variables with finite variance. Pearson's linear correlation coefficient is defined as 
\begin{equation*}
    corr(X,Y)=\frac{cov(X,Y)}{\sqrt{Var(X)Var(Y)}}
\end{equation*}
\end{definition}

\noindent \underline{Remark:} Pearson's correlation coefficient is linear:
\begin{equation*}
    corr(X,Y)=\pm1  \Longleftrightarrow \exists \mbox{ } a, b,  Y=aX+b \mbox{ a.s.}
\end{equation*}

\begin{proposition}
    \label{prop:corr indep}
    \begin{align*}
        X \indep Y & \Rightarrow corr(X,Y)=0 \\
        corr(X,Y)=0 & \nRightarrow X \indep Y
    \end{align*}
\end{proposition}

\noindent \underline{Remark:} This proposition means that uncorrelated data is not independent in general. Furthermore, correlation does not imply causation, meaning that we cannot deduce a cause-and-effect relationship between variables solely based on their correlation. However, in the case of a multivariate normally distributed random vector, variables that are uncorrelated are independent.

\begin{proposition} \label{prop:square-integrable} Set $X$ and $Y$ two random variables. For any measurable function $\Psi$ such that $\Psi(Y)$ is square-integrable,
\begin{equation*}
    \mathbb{E}[\Psi(Y)X]=\mathbb{E}[\Psi(Y)\mathbb{E}[X|Y]]
\end{equation*}
\end{proposition}

\vspace{5mm}
\noindent \underline{\textbf{Matrices}} \cite{horn13}
\begin{definition} A real symmetrical matrix $A$ is positive definite (respectively semi-definite)  if and only if for any non zero real column vector $z$, $z^TAz$ is positive (respectively non negative). \end{definition}

\begin{proposition} \label{prop:eigen pos def} A matrix is positive definite (resp. semi-definite) if and only if all of its eigenvalues are positive (resp. non-negative). \end{proposition}

\begin{definition} \label{def:cholesky} The Cholesky decomposition of a symmetrical real positive-definite matrix is a unique decomposition of the form $A=LL^T$ where $L$ is a lower triangular matrix with real and positive diagonal entries. \end{definition}

\begin{proposition} \label{prop:positive semi-definite cholesky} If $A$ is positive definite (resp. semi-definite), it can be written as $A=LL^T$ with $L$ a lower triangular matrix with a positive (resp. non negative) diagonal. \end{proposition}

\noindent \underline{Remark:} This is the unique (resp. non unique) Cholesky decomposition.

\begin{proposition} \label{prop:eigen invert} If a matrix can be eigendecomposed and all its eigenvalues are non null then it is invertible. \end{proposition}

\vspace{5mm}
\noindent \underline{\textbf{Copulas}} \cite{charpentier10}
\paragraph{} Copulas are used to study multi-hazard risks, ie random vectors $X=(X_1,\dots,X_n)$0. Most of the time, the marginal laws are known. Copulas are then used to get the joint law and model the dependence between variables \cite{fermanian}.

\begin{definition} A d-dimensional copula is a cumulative distribution function $C:[0,1]^d \rightarrow [0,1]$ whose margins are uniform on $[0,1]$0. \end{definition}

\begin{theorem}[Sklar's theorem] \label{th:sklar} For any multivariate cumulative distribution function $H$ with marginals $F_1,\dots,F_d$, there exists a d-dimensional copula $C$ such that
\begin{center} $H(x_1\dots,x_d)=C(F_1(x_1),\dots,F_d(x_d))$ \end{center}
If $F_1,\dots,F_d$ are all continuous, then $C$ is unique. \end{theorem}

\begin{definition} The independence copula is defined as \begin{equation*} C^\perp(u_1,\dots,u_n)=u_1 \times \dots \times u_n \end{equation*} \end{definition}

\noindent \underline{Remark:} We will note $X^\perp$ a random vector which has copula $C^\perp$0.

\begin{definition} For some correlation matrix $R$, the n-dimensional Gaussian copula with parameter $R$ is defined as \begin{equation*} C_R(u) = \Phi_R(\Phi^{-1}(u_1),\dots,\Phi^{-1}(u_n)), u\in[0,1]^n \end{equation*} \end{definition}

\begin{figure}[H]
    \centering
    \includegraphics[scale=0.6]{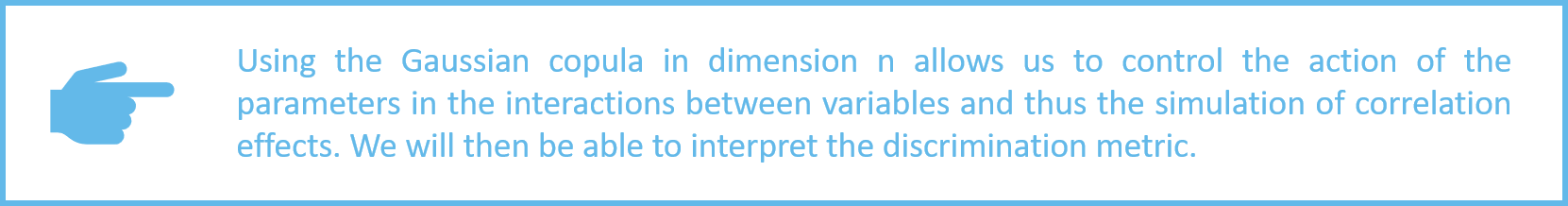}
\end{figure}

\subsection{Simulation process}

\paragraph{} We will now see how to simulate copula models. It all relies on proposition~\ref{prop:sim proc}. Our goal is to simulate $X=(X_1,\dots,X_n)$, which is characterized by its marginal distributions $F_1,\dots,F_n$ and copula $C_{R_Z}$, chosen to be the Gaussian copula. The procedure is the following:
\begin{enumerate}
    \item Draw $Z\sim \mathcal{N}_n(0,R_Z)$
    \item Compute $U=(\Phi(Z_1),\dots,\Phi(Z_n))$
    \item Compute $X=(F_1^{-1}(U_1),\dots,F_n^{-1}(U_n)=(F_1^{-1}(\Phi(Z_1)),\dots,F_n^{-1}(\Phi(Z_n)))$
\end{enumerate}


\begin{figure}[H]
    \centering
    \includegraphics[scale=0.6]{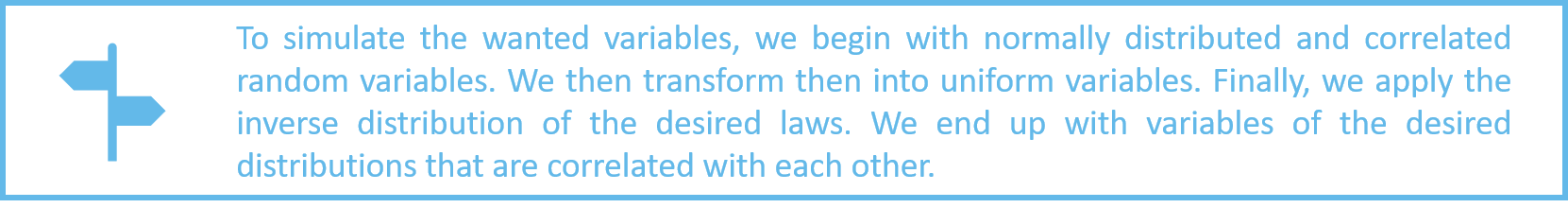}
\end{figure}

\subsubsection{Illustration in dimension 2}

\paragraph{} We will look at the results in dimension n=2 before creating our final dataset. This will allow us to give a visual illustration. First, we generate $Z=(Z_1,Z_2)$ $\sim \mathcal{N}_2(0,R)$ with $R=\begin{bmatrix}
        1 & \rho \\
        \rho & 1
    \end{bmatrix}$
and $\rho=corr(Z_1,Z_2)$0. Then, we compute $X=(X_1,X_2)=\bigl(F_1^{-1}(\Phi(Z_1)),F_2^{-1}(\Phi(Z_2))\bigr)$0. \par
The correlation between the $Z_i$ is
\begin{equation*}
    corr(Z_1,Z_2)=\rho=\frac{\mathbb{E}(Z_1Z_2)-\mathbb{E}(Z_1)\mathbb{E}(Z_2)}{\sqrt{Var(Z_1)Var(Z_2)}}=\mathbb{E}(Z_1Z_2)
\end{equation*} because $Z_i\sim \mathcal{N}(0,1)$, $i=1,2$0. And the correlation between the $X_i$ is
\begin{equation*}
    corr(X_1,X_2)=\frac{\mathbb{E}(X_1X_2)-\mathbb{E}(X_1)\mathbb{E}(X_2)}{\sqrt{Var(X_1)Var(X_2)}}
\end{equation*}

\paragraph{\underline{First case:}} $X_1 \sim \mathcal{N}(\mu_1,\sigma_1^2)$ and $X_2 \sim \mathcal{N}(\mu_2,\sigma_2^2)$ with $\mu_i \in \mathbb{R}, \sigma_i^2 \in \mathbb{R}^+_{\ast}, i=1,2$ \par
\noindent Then
\begin{equation*}
    F_i^{-1}(p)=\mu_i+\sigma_i\Phi^{-1}(p), p \in [0,1], i=1,2
\end{equation*}
In order to find the correlation between the $X_i$, we need to compute:
\begin{equation*}
    \begin{split}
    \mathbb{E}(X_1X_2) & =\mathbb{E}[F_1^{-1}(\Phi(Z_1))F_2^{-1}(\Phi(Z_2))] \\
     & =\mathbb{E}[(\mu_1+\sigma_1\Phi^{-1}(\Phi(Z_1)))(\mu_2+\sigma_2\Phi^{-1}(\Phi(Z_2)))] \\
     & =\mathbb{E}[\mu_1 \mu_2+\mu_1 \sigma_2 Z_2+\mu_2 \sigma_1 Z_1+\sigma_1 \sigma_2 Z_1 Z_2] \\
     & =\mu_1 \mu_2 +\mu_1 \sigma_2 \mathbb{E}(Z_2)+\mu_2 \sigma_1 \mathbb{E}(Z_1)+\sigma_1 \sigma_2 \mathbb{E}(Z_1Z_2) \\
     & =\mu_1 \mu_2 + \sigma_1 \sigma_2 \rho
    \end{split}
\end{equation*}
And so,
\begin{equation}
    corr(X_1,X_2)=\frac{\mu_1 \mu_2 + \sigma_1 \sigma_2 \rho -\mu_1 \mu_2}{\sigma_1 \sigma_2}=\rho
    \label{eq:corr normal normal}
\end{equation}
\vspace{5mm}
\noindent \underline{Illustration:} We will take $X_1 \sim \mathcal{N}(0,2)$ and $X_2 \sim \mathcal{N}(3,1)$0. \par
\begin{figure}[H]
    \centering
    \begin{subfigure}[b]{0.3\textwidth}
        \centering
        \includegraphics[width=\textwidth]{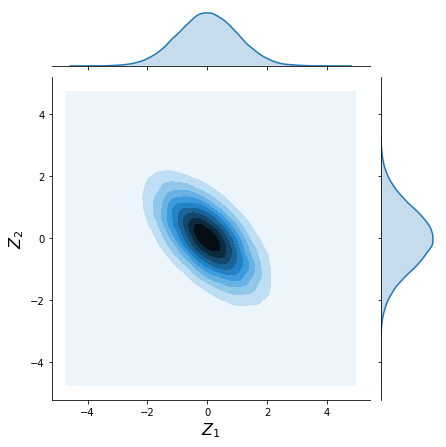}
        \caption{$\rho=-0.6$}
    \end{subfigure}
    \begin{subfigure}[b]{0.3\textwidth}
        \centering
        \includegraphics[width=\textwidth]{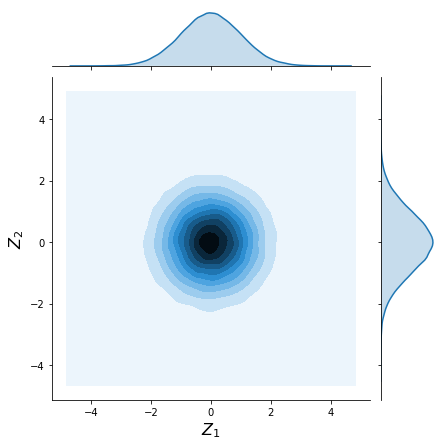}
        \caption{$\rho=0$}
    \end{subfigure}
    \begin{subfigure}[b]{0.3\textwidth}
        \centering
        \includegraphics[width=\textwidth]{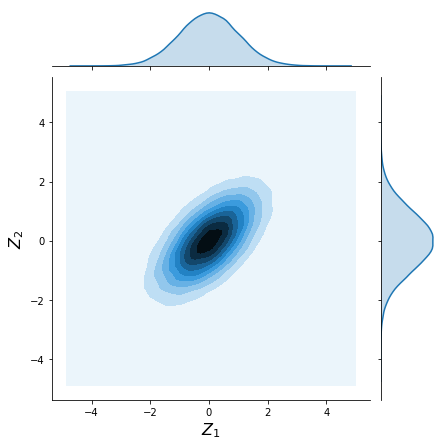}
        \caption{$\rho=0.6$}
    \end{subfigure}
    \caption{Step 1: joint distributions of $(Z_1,Z_2)$}
    \label{fig:Jointplot_1_Z}
\end{figure}

As specified by step 1, we first draw
\begin{equation*}
    Z \sim \mathcal{N}_2 \Bigl(
    \begin{bmatrix}
        0\\
        0
    \end{bmatrix},
    \begin{bmatrix}
            1 & \rho \\
            \rho & 1
    \end{bmatrix} \Bigr)
\end{equation*}
of size $2\times 100,000$0. We will follow the rest of the simulation process for three values of $\rho$: $-0.6$, $0$ and $0.6$, so as to compare the results for different correlations.
Figure~\ref{fig:Jointplot_1_Z} represents the joint distribution of $(Z_1,Z_2)$0. We see that increasing the degree of correlation positively (resp. negatively) between the marginal distributions concentrates the joint distribution around the line $y=x$ (resp. $y=-x$).

\begin{figure}[H]
    \centering
    \begin{subfigure}[b]{0.3\textwidth}
        \centering
        \includegraphics[width=\textwidth]{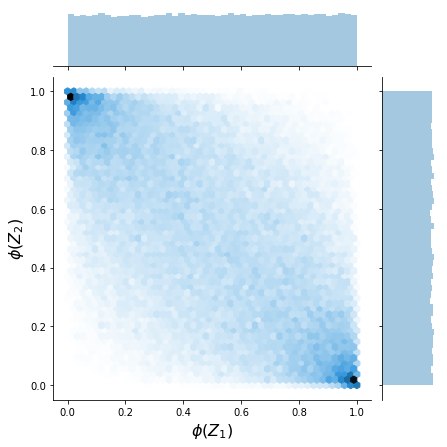}
        \caption{$\rho=-0.6$}
    \end{subfigure}
    \begin{subfigure}[b]{0.3\textwidth}
        \centering
        \includegraphics[width=\textwidth]{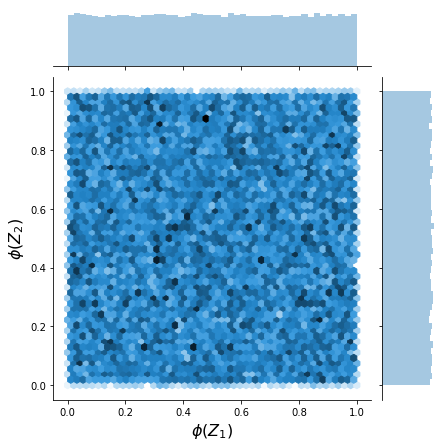}
        \caption{$\rho=0$}
    \end{subfigure}
    \begin{subfigure}[b]{0.3\textwidth}
        \centering
        \includegraphics[width=\textwidth]{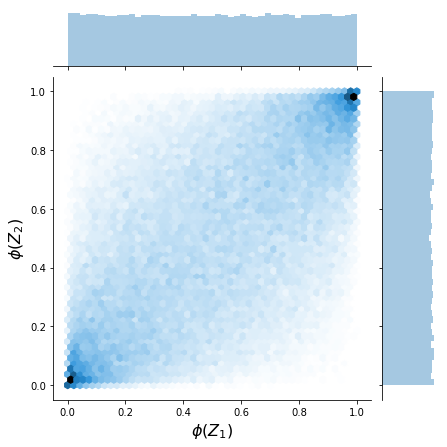}
        \caption{$\rho=0.6$}
    \end{subfigure}
    \caption{Step 2: joint cumulative distributions of $(Z_1,Z_2)$}
    \label{fig:Jointplot_2_phi(Z)}
\end{figure}

We then apply $\Phi$ to $Z_1$ and $Z_2$ (step 2) and get their joint cumulative distribution, as plotted in figure~\ref{fig:Jointplot_2_phi(Z)}. As $Z_i\sim\mathcal{N}(0,1)$, $\Phi(Z_i)\sim\mathcal{U}([0,1])$, $i=1,2$, which is another formulation of proposition~\ref{prop:sim proc}. So we get two uniform random variables that are correlated. Once again, the joint cumulative distribution concentrate around the line $y=\pm x$ when the correlation varies.

\begin{figure}[H]
    \centering
    \begin{subfigure}[b]{0.3\textwidth}
        \centering
        \includegraphics[width=\textwidth]{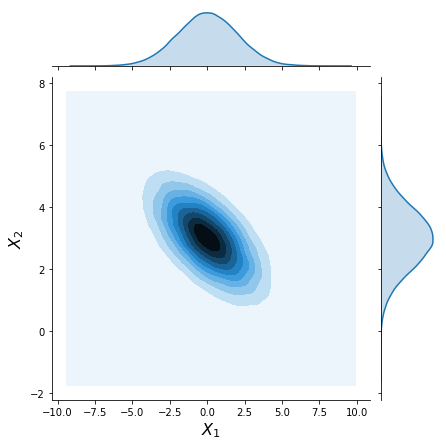}
        \caption{$\rho=-0.6$}
    \end{subfigure}
    \begin{subfigure}[b]{0.3\textwidth}
        \centering
        \includegraphics[width=\textwidth]{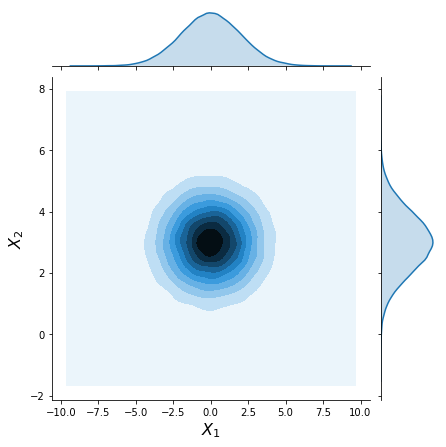}
        \caption{$\rho=0$}
    \end{subfigure}
    \begin{subfigure}[b]{0.3\textwidth}
        \centering
        \includegraphics[width=\textwidth]{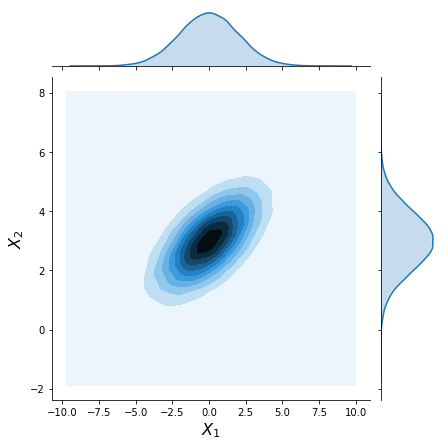}
        \caption{$\rho=0.6$}
    \end{subfigure}
    \caption{Step 3: joint distributions of $(X_1$, $X_2)$}
    \label{fig:Jointplot_3_X}
\end{figure}

We finally apply $F_i^{-1}$ to $\Phi(Z_i)$ to obtain $X_i$, $i=1,2$ (step 3), as plotted in figure~\ref{fig:Jointplot_3_X}. We have the same conclusion as before on the correlation. We are well-aware that we could have directly generated a multivariate normal vector with the desired expected values and standard deviations, but this way allowed us to give an easy example of how to follow the simulation process.

We wrote down the values of the estimated Pearson correlation coefficients in table~\ref{tab:corr_norm_norm}. As computed in equation~\ref{eq:corr normal normal}, we have $corr(X_1,X_2)=corr(Z_1,Z_2)$ for the three values of $\rho$0.
\begin{table}[H]
    \centering
    \begin{tabular}{|c|c|c|c|}
        \hline
        $\rho$ & -0.6 & 0 & 0.6 \\ \hline
        $corr(Z_1,Z_2)$ & -0.6018 & -0.0009 & 0.6005 \\ \hline
        $corr(X_1,X_2)$ & -0.6018 & -0.0009 & 0.6005 \\ \hline
    \end{tabular}
    \caption{$corr(Z_1,Z_2)$ and $corr(X_1,X_2)$}
    \label{tab:corr_norm_norm}
\end{table}

\begin{figure}[H]
    \centering
    \includegraphics[scale=0.6]{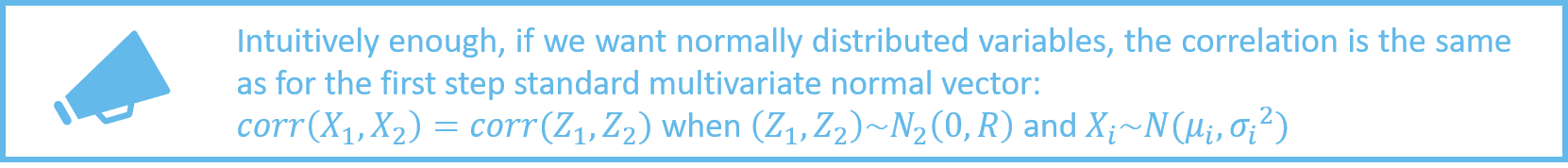}
\end{figure}

\paragraph{\underline{Second case:}} $X_1 \sim \mathcal{U}([a,b])$ and $X_2 \sim \mathcal{N}(\mu,\sigma^2)$ with $(a,b) \in \mathbb{R}^2, a<b, \mu \in \mathbb{R}, \sigma^2 \in \mathbb{R}^+_{\ast}$ \par
\noindent The computation of $corr(X_1,X_2)$ is not as direct as in the first case. As $X_1$ has a uniform distribution, we have
\begin{equation*}
    \begin{split}
         & \mathbb{E}(X_1)=\frac{a+b}{2} \\
         & Var(X_1)=\frac{(b-a)^2}{12} \\
         & F_1^{-1}(p)=a+p(b-a), p \in (0,1)
    \end{split}
\end{equation*}
We can compute
\begin{equation*}
    \begin{split}
        \mathbb{E}(X_1 X_2) & =\mathbb{E}[F_1^{-1}(\Phi(Z_1))F_2^{-1}(\Phi(Z_2))] \\
         & =\mathbb{E}[(a+(b-a)\Phi(Z_1))(\mu+\sigma Z_2)] \\
         & =a\mu+a\sigma\mathbb{E}(Z_2)+(b-a)\mu\mathbb{E}[\Phi(Z_1)]+(b-a)\sigma \mathbb{E}[\Phi(Z_1)Z_2] \\
         & =a\mu+(b-a)\frac{\mu}{2}+(b-a)\sigma \mathbb{E}[\Phi(Z_1)Z_2] \\
         & =\frac{a+b}{2}\mu+(b-a)\sigma \mathbb{E}[\Phi(Z_1)Z_2]
    \end{split}
\end{equation*}
$\mathbb{E}[\Phi(Z_1)]=\frac{1}{2}$ because $Z_1\sim\mathcal{N}(0,1)$ and so $\Phi(Z_1)\sim\mathcal{U}([0,1])$ (consequence of proposition~\ref{prop:sim proc}). \\
So we have
\begin{equation*}
    \begin{split}
        corr(X_1,X_2) & =\frac{\mathbb{E}(X_1X_2)-\mathbb{E}(X_1)\mathbb{E}(X_2)}{\sqrt{Var(X_1)Var(X_2)}} \\
         & =\frac{\frac{a+b}{2}\mu+(b-a)\sigma \mathbb{E}[\Phi(Z_1)Z_2] - \frac{a+b}{2}\mu}{\frac{b-a}{\sqrt{12}}\sigma} \\
         & =2\sqrt{3}\mathbb{E}[\Phi(Z_1)Z_2] \\
         & =2\sqrt{3}\mathbb{E}[\Phi(Z_1)\mathbb{E}[Z_2|Z_1]]
    \end{split}
\end{equation*}
because of proposition~\ref{prop:square-integrable}, with $\Phi(Z_1)$ square-integrable. \\
And $\mathbb{E}[Z_2|Z_1]=\rho Z_1$, because
\begin{equation}
    Z_2|Z_1=z_1\sim\mathcal{N}(\rho z_1, 1-\rho^2)
    \label{eq:Z2|Z1}
\end{equation}
\begin{proof}
    \begin{equation*}
        \begin{split}
            f_{Z_2|Z_1=z_1}(z_1,z2) & =\frac{f_{Z_1,Z_2}(z_1,z_2)}{f_{Z_1}(z_1)} \\
             & =\frac{\frac{1}{2\pi \sqrt{1-\rho^2}}exp(-\frac{1}{2(1-\rho)^2}(z_1^2-2\rho z_1z_2+z_2^2))}{\frac{1}{\sqrt{2\pi}}exp(-\frac{z_1^2}{2})} \\
             & =\frac{1}{\sqrt{2\pi (1-\rho^2)}}exp(-\frac{1}{2(1-\rho^2)}(z_1^2-2\rho z_1z_2+z_2^2-(1-\rho^2)z_1^2)) \\
             & =\frac{1}{\sqrt{2\pi (1-\rho^2)}}exp(-\frac{1}{2(1-\rho^2)}(z_2^2-2\rho z_1z_2+(\rho z_1)^2)) \\
             & =\frac{1}{\sqrt{2\pi (1-\rho^2)}}exp(-\frac{(z_2-\rho z_1)^2}{2(1-\rho^2)}) \\
        \end{split}
    \end{equation*}
    By identification,
    \begin{equation*}
        \begin{split}
            \mathbb{E}[Z_2|Z_1=z_1]=\rho z_1 \\
            Var(Z_2|Z_1=z_1)=1-\rho^2
        \end{split}
    \end{equation*}
\end{proof}
\noindent So we have
\begin{equation*}
    \begin{split}
        corr(X_1,X_2) & =2\sqrt{3}\mathbb{E}[\Phi(Z_1)\rho Z_1] \\
        & =2\sqrt{3}\rho \mathbb{E}[\Phi(Z_1)Z_1] \\
        & =2\sqrt{3}\rho \int_\mathbb{R} z\Phi(z) \phi(z) \,dz
    \end{split}
\end{equation*}
We will do an integration by parts:
\begin{equation*}
    \int_a^b u'(x)v(x) \,dx = [u(x)v(x)]_a^b-\int_a^b u(x)v'(x) \,dx
\end{equation*}
We can set 
\begin{equation*}
    \begin{split}
        u'(x) & =x\phi(x)=x\frac{e^{-x^2/2}}{\sqrt{2\pi}} \\
        v(x) & =\Phi(x)
    \end{split}
\end{equation*}
So
\begin{equation*}
    \begin{split}
        u(x) & =-\phi(x)=-\frac{e^{-x^2/2}}{\sqrt{2\pi}} \\
        v'(x) & =\Phi'(x)=\phi(x)
    \end{split}
\end{equation*}
So we have
\begin{equation*}
    \begin{split}
        corr(X_1,X_2) & =2\sqrt{3}\rho \bigl[[-\phi(z)\Phi(z)]_\mathbb{R} - \int_\mathbb{R} -\phi^2(z) \,dz\bigr]
    \end{split}
\end{equation*}
$\lim_{z\to-\infty} \Phi(z)=0$ and $\lim_{z\to+\infty} \Phi(z)=1$ because $\Phi$ is a cumulative distribution function \\
$\phi(z)=\frac{e^{-z^2/2}}{\sqrt{2\pi}}$ so $\lim_{z\to-\infty} \phi(z)=\lim_{z\to+\infty} \phi(z)=0$ \\
So we have $[-\phi(z)\Phi(z)]_\mathbb{R}=0$
and
\begin{equation*}
    \begin{split}
        \int_\mathbb{R} \phi^2(z) \,dz & =\int_\mathbb{R} \frac{e^{-z^2}}{2\pi} \,dz \\
        & =\int_\mathbb{R} \frac{e^{-y^2/2}}{2\pi \sqrt{2}} \,dy \\
        & =\frac{1}{2\sqrt{\pi}}
    \end{split}
\end{equation*}
because $\int_\mathbb{R} \frac{e^{-y^2/2}}{\sqrt{2\pi}} \,dy=1$, as it is the integral on the entire support ($\mathbb{R}$) of a probability density function (of the standard normal distribution). In the end,
\begin{equation*}
    corr(X_1,X_2)=2\sqrt{3}\rho \frac{1}{2\sqrt{\pi}}
\end{equation*}
\begin{equation}
    corr(X_1,X_2)=\sqrt{\frac{3}{\pi}} \rho
    \label{eq: corr normal uniform}
\end{equation}
To conclude, as $\sqrt{\frac{3}{\pi}}\simeq0.997$, we almost keep the same correlation between $X_1$ and $X_2$ as between $Z_1$ and $Z_2$, in the case where $X_1\sim \mathcal{U}([a,b])$ and $X_2 \sim \mathcal{N}(\mu,\sigma^2)$0. \par
\begin{figure}[H]
    \centering
    \begin{subfigure}[b]{0.3\textwidth}
        \centering
        \includegraphics[width=\textwidth]{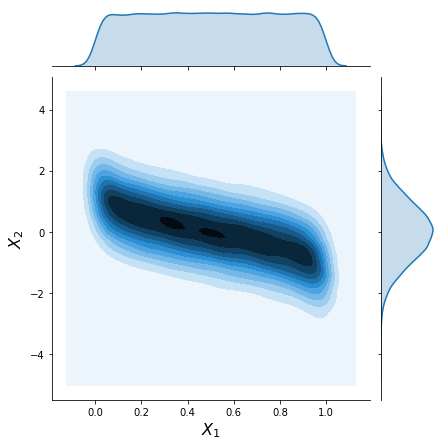}
        \caption{$\rho=-0.6$}
    \end{subfigure}
    \begin{subfigure}[b]{0.3\textwidth}
        \centering
        \includegraphics[width=\textwidth]{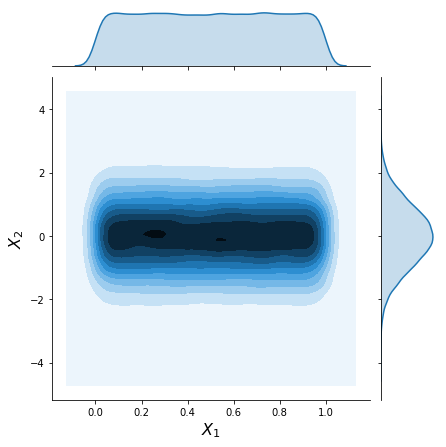}
        \caption{$\rho=0$}
    \end{subfigure}
    \begin{subfigure}[b]{0.3\textwidth}
        \centering
        \includegraphics[width=\textwidth]{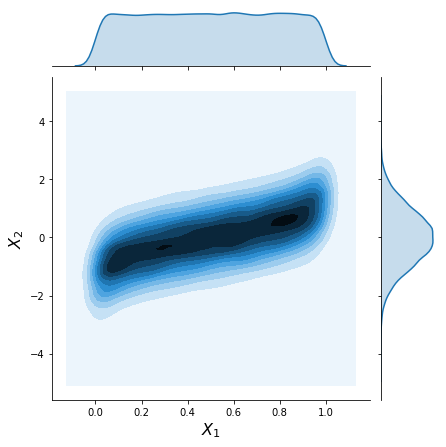}
        \caption{$\rho=0.6$}
    \end{subfigure}
    \caption{Joint distributions of $(X_1$, $X_2)=\bigl(F_1^{-1}(\Phi(Z_1)),F_2^{-1}(\Phi(Z_2))\bigr)$}
    \label{fig:Jointplot_3_X_norm_unif}
\end{figure}

\vspace{5mm}
\noindent \underline{Illustration:} We will take $X_1\sim\mathcal{U}([0,1])$ and $X_2\sim\mathcal{N}(0,1)$0. The joint distributions are plotted in figure~\ref{fig:Jointplot_3_X_norm_unif}. Like in the previous case, increasing the degree of correlation positively (respectively negatively) between the marginal distributions shifts the joint distribution on the line $y=x$ (respectively $y=-x$).

We computed the estimated Pearson correlation coefficients between the simulated $Z_1$ and $Z_2$, and $X_1$ and $X_2$ in table~\ref{tab:corr_norm_unif}. As found with equation~\ref{eq: corr normal uniform}, $corr(X_1,X_2)=\sqrt{\frac{3}{\pi}}\rho$ for the three values of $\rho$0. \par
\vspace{5mm}
\begin{table}[H]
    \centering
    \begin{tabular}{|c|c|c|c|}
        \hline
        $\rho$ & -0.6 & 0 & 0.6 \\ \hline
        $corr(Z_1,Z_2)$ & -0.5983 & -0.0031 & 0.6006 \\ \hline
        $\sqrt{3/\pi}\rho$ & -0.586 & 0 & 0.586 \\ \hline
        $corr(X_1,X_2)$ & -0.5838 & -0.0029 & 0.5874 \\ \hline
    \end{tabular}
    \caption{$corr(X_1,X_2)$}
    \label{tab:corr_norm_unif}
\end{table}

\begin{figure}[H]
    \centering
    \includegraphics[scale=0.6]{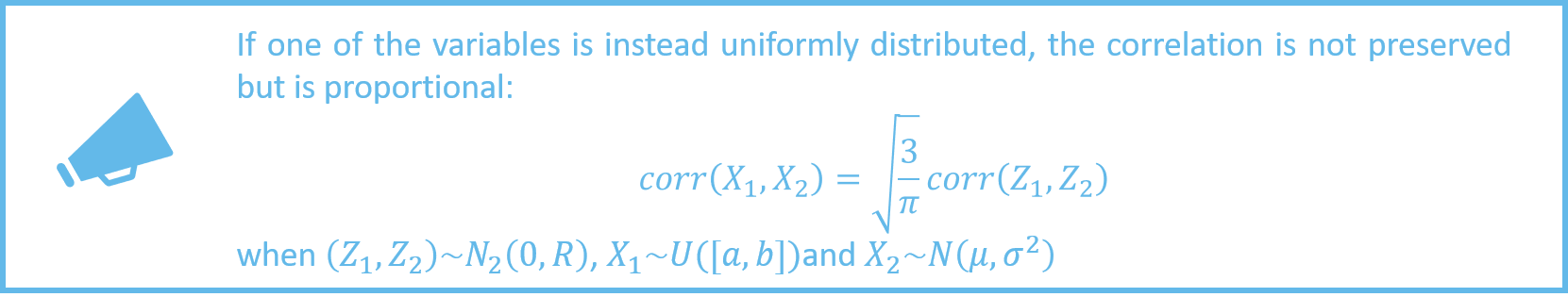}
\end{figure}

\paragraph{\underline{Third case:}} $X_1 \sim \mathcal{B}(p)$ and $X_2 \sim \mathcal{N}(\mu,\sigma^2)$, $p \in (0,1),\mu \in \mathbb{R}, \sigma^2 \in \mathbb{R}^+_{\ast}$ \par
\noindent As $F_1$ is not continuous, we are not under the assumptions of Sklar's theorem (theorem~\ref{th:sklar}), so the copula $C$ might not be unique. \par
\begin{figure}[H]
    \centering
    \includegraphics[scale=0.5]{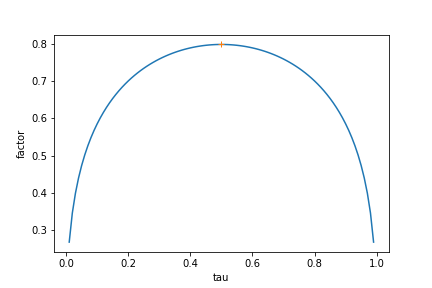}
    \caption{$corr(h(X_1),X_2)$ for $\tau\in(0,1)$, with a maximum of $0.798$ for $\tau=0.5$}
    \label{fig:factor_p}
\end{figure}

We will place ourselves in the second case and simulate $X_1 \sim \mathcal{U}([0,1])$ and $X_2 \sim \mathcal{N}(\mu,\sigma^2)$ with $\mu \in \mathbb{R}, \sigma^2 \in \mathbb{R}^+_{\ast}$, then transform $X_1$ into a Bernoulli random variable using the function
\begin{equation*}
    \begin{split}
        h: & [0,1] \rightarrow \{0,1\} \\
         & u \mapsto h(u)=\left\{
    \begin{array}{ll}
        0 & \mbox{if } u<\tau \\
        1 & \mbox{if } u\geq \tau
    \end{array}
\right. =\mathbb{1}_{u \geq \tau}
    \end{split}
\end{equation*}
We then have $h(X_1) \sim \mathcal{B}(p)$0.
\begin{proof}
    \begin{equation*}
        \mathbb{P}(h(X_1)=1)=\mathbb{P}(\mathbb{1}_{X_1 \geq \tau}=1)=\mathbb{P}(X_1 \geq \tau)=1-F_1(\tau)=1-\tau
    \end{equation*}
    \begin{equation*}
        \mathbb{P}(h(X_1)=0)=\mathbb{P}(\mathbb{1}_{X_1 \geq \tau}=0)=1-\mathbb{P}(\mathbb{1}_{X_1 \geq \tau}=1)=\tau
    \end{equation*}
    By identification, $h(X_1) \sim \mathcal{B}(p)$ with $p=1-\tau$
\end{proof}
As the cumulative distribution function of a Bernoulli random variable is not invertible, we cannot go back to the generative copula from the set $(h(X_1),X_2)$0. In a sense, we have broken the correlation structure. But our transformation $h$ allows the computation of the correlation as a function of $\rho$:
\begin{equation*}
    corr(h(X_1),X_2)=\frac{\mathbb{E}[h(X_1)X_2]-\mathbb{E}[h(X_1)]\mathbb{E}[X_2]}{\sqrt{Var(h(X_1))Var(X_2)}}
\end{equation*}
with
\begin{equation*}
    \begin{split}
        \mathbb{E}[h(X_1)X_2] & =\mathbb{E}[h(X_1)F_2^{-1}(\Phi(Z_2))] \\
         & =\mathbb{E}[h(X_1)(\mu+\sigma Z_2)] \\
         & =\mu \mathbb{E}[h(X_1)]+\sigma \mathbb{E}[h(X_1)Z_2]
    \end{split}
\end{equation*}
and
\begin{equation*}
    \begin{split}
        \mathbb{E}[h(X_1)Z_2] & =\mathbb{E}[\mathbb{1}_{X_1 \geq \tau}Z_2] \\
         & =\mathbb{E}[\mathbb{1}_{F_1^{-1}(\Phi(Z_1)) \geq \tau}Z_2] \\
         & =\mathbb{E}[\mathbb{1}_{Z_1 \geq \Phi^{-1}(F_1(\tau))}Z_2] \\
         & =\mathbb{E}[\mathbb{1}_{Z_1 \geq \Phi^{-1}(\tau)}Z_2] \\
         & =\mathbb{E}[\mathbb{1}_{Z_1 \geq \Phi^{-1}(\tau)}\mathbb{E}[Z_2|Z_1]] \mbox{ (proposition~\ref{prop:square-integrable})} \\
         & =\mathbb{E}[\mathbb{1}_{Z_1 \geq \Phi^{-1}(\tau)}\rho Z_1] \mbox{ (equation~\ref{eq:Z2|Z1})} \\
         & =\rho \int_\mathbb{R} \mathbb{1}_{z \geq \Phi^{-1}(\tau)}z \phi(z) \,dz \\
         & =\rho \int_{\Phi^{-1}(\tau)}^{+\infty} z \phi(z) \,dz \\
         & =\rho[-\phi(z)]_{\Phi^{-1}(\tau)}^{+\infty} \\
         & =\rho \phi(\Phi^{-1}(\tau))
    \end{split}
\end{equation*}
So
\begin{equation*}
    \begin{split}
        corr(h(X_1),X_2) & =\frac{\mu \mathbb{E}[h(X_1)]+\sigma \rho \phi(\Phi^{-1}(\tau))-\mathbb{E}[h(X_1)]\mu}{\sqrt{\tau(1-\tau)\sigma^2}} \\
         & =\frac{\rho \phi(\Phi^{-1}(\tau))}{\sqrt{\tau(1-\tau)}} \\
         & =\frac{\rho \frac{1}{\sqrt{2\pi}}e^{-\frac{(\Phi^{-1}(\tau))^2}{2}}}{\sqrt{\tau(1-\tau)}}
    \end{split}
\end{equation*}
So we have
\begin{equation}
    corr(h(X_1),X_2)=\frac{\rho e^{-\frac{(\Phi^{-1}(\tau))^2}{2}}}{\sqrt{\tau(1-\tau)2\pi}}
    \label{eq:corr bern norm}
\end{equation}
As the correlation coefficient is in $[-,1]$,
\begin{equation*}
    \begin{split}
        \min_{\rho \in (-1,1)} corr(h(X_1),X_2) & =\min_{\rho \in (-1,1)} \frac{\rho e^{-\frac{(\Phi^{-1}(\tau))^2}{2}}}{\sqrt{\tau(1-\tau)2\pi}}=-\frac{e^{-\frac{(\Phi^{-1}(\tau))^2}{2}}}{\sqrt{\tau(1-\tau)2\pi}} \\
        \max_{\rho \in (-1,1)} corr(h(X_1),X_2) & =\max_{\rho \in (-1,1)} \frac{\rho e^{-\frac{(\Phi^{-1}(\tau))^2}{2}}}{\sqrt{\tau(1-\tau)2\pi}}=\frac{e^{-\frac{(\Phi^{-1}(\tau))^2}{2}}}{\sqrt{\tau(1-\tau)2\pi}}
    \end{split}
\end{equation*}
So $corr(h(X_1),X_2)$ has a minimal and maximal value:
\begin{equation*}
    corr(h(X_1),X_2) \in \Bigl[-\frac{e^{-\frac{(\Phi^{-1}(\tau))^2}{2}}}{\sqrt{\tau(1-\tau)2\pi}},\frac{e^{-\frac{(\Phi^{-1}(\tau))^2}{2}}}{\sqrt{\tau(1-\tau)2\pi}}\Bigr]
\end{equation*}
Figure~\ref{fig:factor_p} represents $\frac{e^{-\frac{(\Phi^{-1}(\tau))^2}{2}}}{\sqrt{\tau(1-\tau)2\pi}}$ for $\tau\in(0,1)$0.
\begin{equation*}
    \max_{\tau\in(0,1)}(\frac{e^{-\frac{(\Phi^{-1}(\tau))^2}{2}}}{\sqrt{\tau(1-\tau)2\pi}})=0.798 \mbox{ for } \tau=\frac{1}{2}
\end{equation*}
So for all $\tau\in(0,1),|corr(h(X_1),X_2)|\leq0.798$0.

\noindent \underline{Illustration:} We will take $X_1\sim \mathcal{B}(0.8)$ and $X_2 \sim \mathcal{N}(2,0.8)$0. \par
\begin{figure}[H]
    \centering
    \begin{subfigure}[b]{0.3\textwidth}
        \centering
        \includegraphics[width=\textwidth]{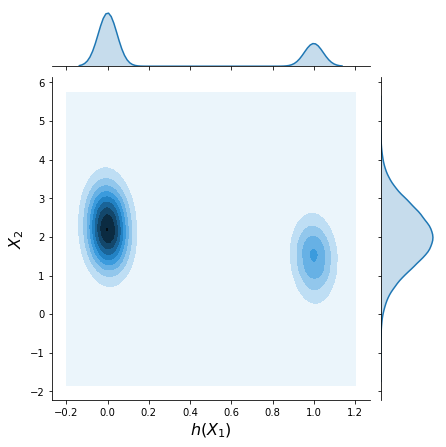}
        \caption{$\rho=-0.6$}
    \end{subfigure}
    \begin{subfigure}[b]{0.3\textwidth}
        \centering
        \includegraphics[width=\textwidth]{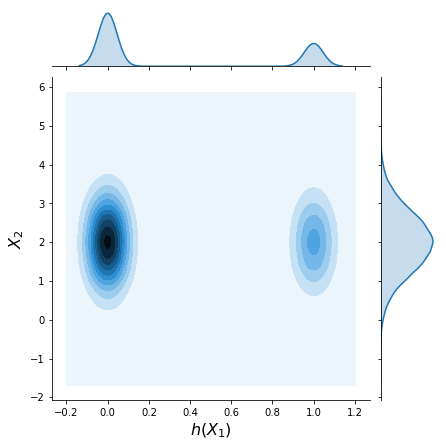}
        \caption{$\rho=0$}
    \end{subfigure}
    \begin{subfigure}[b]{0.3\textwidth}
        \centering
        \includegraphics[width=\textwidth]{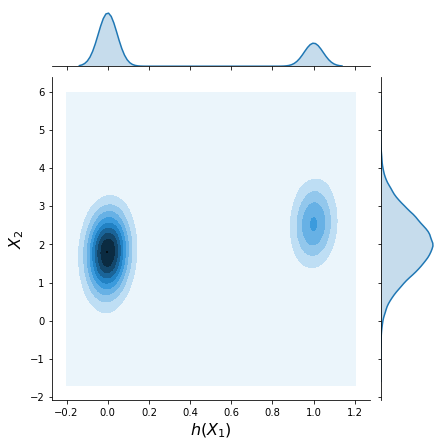}
        \caption{$\rho=0.6$}
    \end{subfigure}
    \caption{Joint distributions of $(h(X_1)$, $X_2)$}
    \label{fig:Jointplot_bern_norm}
\end{figure}

\noindent Then $\Phi^{-1}(\tau)=\Phi^{-1}(1-0.8)=\Phi^{-1}(0.2)\simeq-0.842$ (numeric computation) and
\begin{equation*}
    \frac{e^{-\frac{(\Phi^{-1}(\tau))^2}{2}}}{\sqrt{\tau(1-\tau)2\pi}}=\frac{e^{-\frac{(-0.842)^2}{2}}}{\sqrt{(1-0.8)\times 0.8\times 2\pi}}\simeq 0.759
\end{equation*}
So
\begin{equation*}
    corr(h(X_1),X_2)=0.759\rho \in(-0.759,0.759)
\end{equation*}
The joint distributions are plotted in figure~\ref{fig:Jointplot_bern_norm}. Like in the previous cases, increasing the degree of correlation positively (respectively negatively) between the marginal distributions shifts the joint distribution on the line $y=x$ (respectively $y=-x$).

We computed the estimated Pearson correlation coefficients between the simulated $h(X_1)$ and $X_2$ in table~\ref{tab:h(x1)x2}. We can see that $corr(h(X_1),X_2)=\frac{\rho e^{-\frac{(\Phi^{-1}(\tau))^2}{2}}}{\sqrt{\tau(1-\tau)2\pi}}$ for the three values of $\rho$, as we computed in equation~\ref{eq:corr bern norm}.

\begin{table}[H]
    \centering
    \begin{tabular}{|c|c|c|c|}
        \hline
        $\rho$ & -0.6 & 0 & 0.6 \\ \hline
        $\frac{\rho e^{-\frac{(\Phi^{-1}(\tau))^2}{2}}}{\sqrt{\tau(1-\tau)2\pi}}$ & -0.455 & -0.001 & 0.455 \\ \hline
        $corr(h(X_1),X_2)$ & -0.454 & 0.000 & 0.455 \\ \hline
    \end{tabular}
    \caption{$corr(h(X_1),X_2)$}
    \label{tab:h(x1)x2}
\end{table}

\begin{figure}[H]
    \centering
    \includegraphics[scale=0.6]{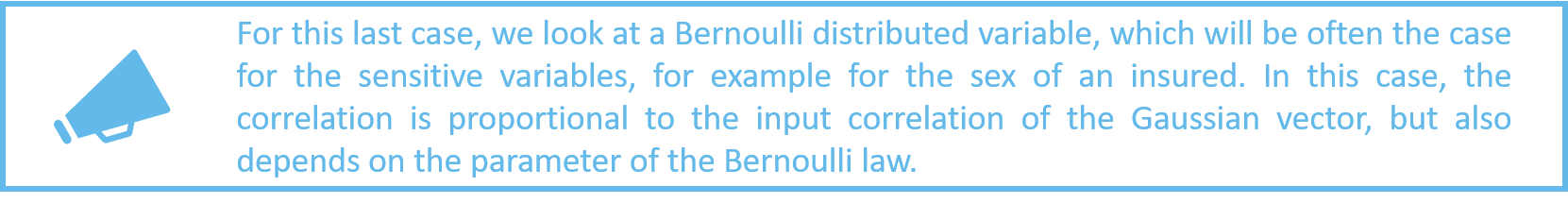}
\end{figure}

\subsubsection{Illustration in dimension $n$: creating the dataset}
\paragraph{} We can now create the simulated dataset. For simplicity reasons, we will suppose that we are in the setting of section~\ref{binary_classif}, with two binary protected variables and a binary outcome. We have
\begin{itemize}
    \item a set of non-protected variables $\bm{X} = \{X^{(i)}\}_{i=1,...n-3}$, $n \in \mathbb{N}$ 
    with $X^{(i)}\sim\mathcal{N}(\mu_i,\sigma_i^2)$, $\mu_i \in \mathbb{R}$, $\sigma_i \in \mathbb{R}_\ast^+$
    \item two protected variables $A\sim\mathcal{B}(p_a),p_a\in(0,1)$ and $B\sim\mathcal{B}(p_b),p_b\in(0,1)$
    \item the output variable $Y\sim\mathcal{B}(p_y),p_y\in(0,1)$
    \item $\bm{Z}\sim \mathcal{N}_n(0,R_Z)$0. For $(i,j)\in\llbracket 1,n\rrbracket^2$,
    \begin{equation*}
        R_{Z,ij}=\left\{
        \begin{array}{ll}
            1 & \mbox{if } i=j \\
            corr(Z_i,Z_j) & \mbox{if } i\neq j
        \end{array}
        \right.
    \end{equation*}
    \item[] with $corr(X^{(i)},X^{(j)})=corr(Z_i,Z_j)$, $(i,j)\in\llbracket 1,n-3\rrbracket$ (equation~\ref{eq:corr normal normal})
    \item[] $corr(X^{(i)},A)=\frac{e^{-\frac{(\Phi^{-1}(1-p_a))^2}{2}}}{\sqrt{p_a(1-p_a)2\pi}}corr(Z_i,Z_{n-2}),i\in\llbracket 1,n-3\rrbracket$ (equation~\ref{eq: corr normal uniform})
    \item[] $corr(X^{(i)},B)=\frac{e^{-\frac{(\Phi^{-1}(1-p_b))^2}{2}}}{\sqrt{p_b(1-p_b)2\pi}}corr(Z_i,Z_{n-1}),i\in\llbracket 1,n-3\rrbracket$ (equation~\ref{eq: corr normal uniform})
    \item[] $corr(X^{(i)},Y)=\frac{e^{-\frac{(\Phi^{-1}(1-p_y))^2}{2}}}{\sqrt{p_y(1-p_y)2\pi}}corr(Z_i,Z_n),i\in\llbracket 1,n-3\rrbracket$ (equation~\ref{eq: corr normal uniform})
\end{itemize}

\noindent For the choice of $R_Z$, there are constraints. It must be a square matrix and:
\begin{itemize}
    \item symmetrical, because $corr(Z_i,Z_j)=corr(Z_j,Z_i)$
    \item with a unity diagonal, because $corr(Z_i,Z_i)=1$
    \item all values must be in $[-1,1]$ by definition of the correlation
    \item positive semi-definite
    \item invertible, otherwise the distribution is degenerate and does not have a density, as we saw in definition~\ref{def:multiv normal}
\end{itemize}
\noindent In order to have an invertible and positive semi-definite matrix, it needs to be positive definite. Indeed, proposition~\ref{prop:eigen invert} gives the equivalence between an invertible matrix and its eigenvalues being non null, and proposition ~\ref{prop:eigen pos def} gives the equivalence between a positive definite (respectively semi-definite) matrix and its eigenvalues being positive (respectively positive or null). \par
Using proposition~\ref{prop:positive semi-definite cholesky} (Cholesky decomposition), we will compute $R_Z=LL^T$, choosing $L$ as a lower triangular matrix with a positive diagonal. That way we will have a square symmetrical, positive definite (so positive semi-definite and invertible) matrix.
\begin{equation*}
    \begin{split}
        R_Z & =
        \begin{bmatrix}
            L_{11} & 0 & 0 & \dots \\
            L_{21} & L_{22} & 0 & \dots \\
            \vdots & & \ddots &
        \end{bmatrix}
        \times
        \begin{bmatrix}
            L_{11} & L_{21} & \dots \\
            0 & L_{22} & \dots \\
            \vdots & 0 & \ddots
        \end{bmatrix} \\
         & =\begin{bmatrix}
            L_{11}^2 & L_{11}L_{21} & \dots \\
            L_{11}L_{21} & L_{21}^2+L_{22}^2 & \\
            \vdots & & \ddots
        \end{bmatrix}
    \end{split}
\end{equation*}
We have
\begin{equation*}
    \begin{split}
        R_{Z,ii} & =\sum_{k=1}^i L_{ik}^2 \\
        R_{Z,ij} & =R_{Z,ji}=\sum_{k=1}^{i} L_{ik}L_{jk}, i>j
    \end{split}
\end{equation*}
As we want the diagonal of $R_Z$ to be unity, we need
\begin{equation*}
    \begin{split}
        R_{Z,ii}=1 & \text{ ie } \sum_{k=1}^i L_{ik}^2=1 \\
         & \text{ ie } L_{ii}=\sqrt{1-\sum_{k=1}^{i-1}L_{ik}^2} \mbox{ (the diagonal has to be positive)}
    \end{split}
\end{equation*}
The computation of the $L_{ii}$ implies constraints on the values of the $L_{ij}$:
\begin{equation*}
    \forall i=1,\dots,n, 1-\sum_{k=1}^{i-1}L_{ik}^2>0 \mbox{ ie } \sum_{k=1}^{i-1}L_{ik}^2<1
\end{equation*}
Finally, to set $R_Z$, we set values for the $L_{ij},i>j$, check that the constraint above is verified, compute the $L_{ii}$ accordingly and then $R_Z$0. We also need to check that we have $R_{Z,ij}\in[-1,1]$0. \par
Finally, the procedure is:
\begin{itemize}
    \item set the matrix $L$ with the constraints mentioned above
    \item compute $R_Z=LL^T$
    \item generate $Z\sim \mathcal{N}_n(0,R_Z)$
    \item set the parameters of the laws of \begin{itemize}
        \item[] $X^{(i)}\sim\mathcal{N}(\mu_i,\sigma_i^2)$
        \item[] $A\sim\mathcal{B}(p_a)$
        \item[] $B\sim\mathcal{B}(p_b)$
        \item[] $Y\sim\mathcal{B}(p_y)$
    \end{itemize}
    \item compute \begin{itemize}
        \item[] $\bm{X}=(F_1^{-1}(\Phi(Z_1)),\dots,F_{n-3}^{-1}(\Phi(Z_{n-3})))$
        \item[] $A=h_a(F_{n-3}^{-1}(\Phi(Z_{n-3})))$
        \item[] $B=h_b(F_{n-2}^{-1}(\Phi(Z_{n-2})))$
        \item[] $Y=h_y(F_{n}^{-1}(\Phi(Z_{n})))$
    \end{itemize}
    Reminder: $h_a(u) =\mathbb{1}_{u \geq 1-p_a}$
\end{itemize}

\paragraph{Final dataset} We set $n=7$ and generated 100 datasets with the same parameters:
\begin{equation*}
    \begin{split}
        \label{def x_i}
        X^{(1)} & \sim\mathcal{N}(2,0.6) \\
        X^{(2)} & \sim\mathcal{N}(0.2,0.3) \\
        X^{(3)} & \sim\mathcal{N}(-0.3,2) \\
        X^{(4)} & \sim\mathcal{N}(0.7,0.4) \\
        A & \sim\mathcal{B}(0.3) \\
        B & \sim\mathcal{B}(0.9) \\
        Y & \sim\mathcal{B}(0.2)
    \end{split}
\end{equation*}
\begin{equation*}
    \label{theoretical corr R_xa}
    R_{X,A,B,Y}=\begin{bmatrix}
        1 & 0.395 & -0.018 & 0.297 & -0.230 & 0.350 & 0.139 \\
        0.395 & 1 & -0.501 & 0.103 & 0.226 & 0.111 & 0.209 \\
        -0.018 & -0.501 & 1 & 0.294 & 0.076 & 0.294 & -0.066 \\
        0.297 & 0.103 & 0.294 & 1 & -0.227 & 0.348 & -0.208 \\
        -0.230 & 0.226 & 0.076 & -0.227 & 1 & 0.043 & 0.105 \\
        0.350 & 0.111 & 0.294 & 0.348 & 0.043 & 1 & 0.039 \\
        0.139 & 0.209 & -0.066 & -0.208 & 0.105 & 0.039 & 1
    \end{bmatrix}
\end{equation*}

\begin{figure}[H]
    \centering
    \includegraphics[scale=0.5]{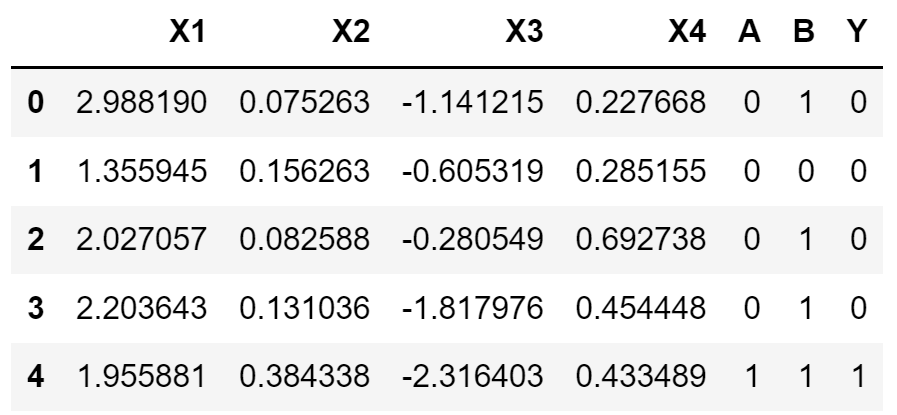}
    \caption{Head of a simulated dataset}
    \label{fig:table_sim_data}
\end{figure}

The reason for that is that we want to take into account the instability of results. Instability can come from the generative process: not all datasets will have the exact same variable distribution because of the limited sample size (100,00 lines here). It can also be caused later on by the train test split or sampling which depend strongly on the execution. Generating 100 datasets allows us to average and get confidence intervals on metrics and results. In reality, we often do not have access to the data generator so, to construct confidence intervals, we use bootstraping, which estimates the sampling distribution of statistics such as sample mean thanks to random sampling with replacement. Figure~\ref{fig:table_sim_data} shows the head of one of the datasets. \par

As mentioned previously, we will take advantage of having access to the data generation process. It means we can produce confidence intervals for every value: the mean of a variable, the number of observations of a certain class, the weights of the regression etc. This is the reason why we generated 100 datasets: we will look at the average and standard deviation over these datasets to produce the confidence interval of the value we are looking at. We gave a reminder on confidence intervals in appendix~\ref{appendix:confidence intervals}. 

\begin{figure}[H]
    \centering
    \includegraphics[scale=0.6]{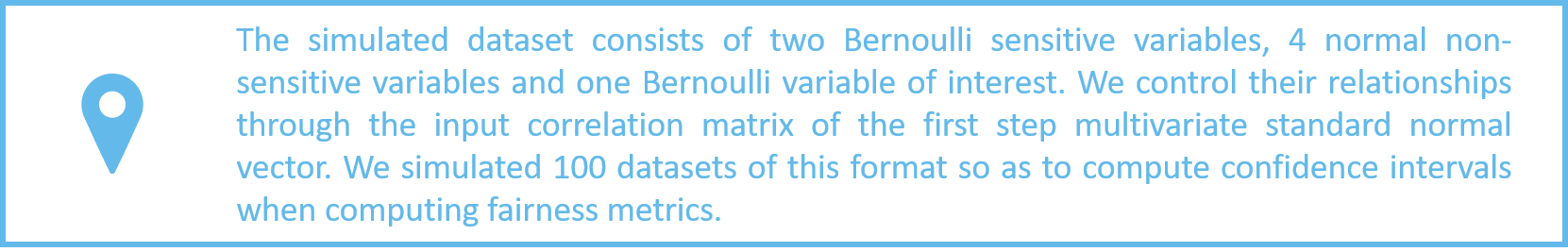}
\end{figure}

\subsection{Descriptive statistics}
\paragraph{} Before heading into applying methods, we need to prepare and explore our data. As we have built our datasets ourselves, we already know a lot about them.  \par

\subsubsection{Variable identification and univariate analysis}
\paragraph{} The explanatory variables are the $X^{(i)},i=1,\dots,4$, $A$ and $B$, and the target variable is $Y$0. As expected, we have 100,000 non-null values for each variable. The $X^{(i)}$ are continuous, and $A$, $B$ and $Y$ are categorical, taking values in $\{0,1\}$0. \par

\paragraph{X} By construction, the $X^{(i)}$ have means and standard derivations as defined in section~\ref{def x_i}, and it is verified by the computation of the sample means and standard derivations in table~\ref{tab:X_i describe}. We can notice that the confidence intervals are of size 0. This means that over our 100 datasets, we have a probability of 95\% that they all have the means and standard deviations as defined in data generation process.

\begin{table}[H]
    \centering
    \begin{tabular}{|c|c|c|c|c|}
        \hline
        i & 1 & 2 & 3 & 4 \\ \hline
        mean & $2.00\pm0.00$ & $0.20\pm0.00$ & $-0.30\pm0.00$ & $0.70\pm0.00$ \\ \hline
        std & $0.60\pm0.00$ & $0.30\pm0.00$ & $2.00\pm0.00$ & $0.40\pm0.00$  \\ \hline
    \end{tabular}
    \caption{Means and standard derivations of the $X^{(i)}$}
    \label{tab:X_i describe}
\end{table}

\paragraph{A} We computed it to follow a Bernoulli distribution of parameter $p_a=0.3$, so we find as planned that
\begin{equation*}
    \begin{split}
        \mbox{mean(A)} & =0.30\pm0.00=p_a \\
        \mbox{std(A)} & =0.46\pm0.00=\sqrt{0.3(1-0.3)}=\sqrt{p_a(1-p_a)}
    \end{split}
\end{equation*}
Table~\ref{tab:A counts} gives the number of observations for each value of $A$0. By construction, as $A \sim \mathcal{B}(0.3)$, there is an imbalance: about $30\%$ individuals have $A=1$ and $70\%$ individuals have $A=0$0. So the group imbalance ratio is
\begin{equation*}
    IR_A = \frac{\mbox{Number of majority observations}}{\mbox{Number of minority observations}} = 2.33\pm0.00
\end{equation*}
meaning there are 2.33 times more observations of $A=0$ than $A=1$0.

\begin{table}[H]
    \centering
    \begin{tabular}{|c|c|}
        \hline
        A & Observations \\ \hline
        0 & $70,002.22\pm290.73$ \\ \hline
        1 & $29,997.78\pm290.73$ \\ \hline
    \end{tabular}
    \caption{Number of observations by value of $A$}
    \label{tab:A counts}
\end{table}

\paragraph{B} We computed it to follow a Bernoulli distribution of parameter $p_b=0.9$, so we find as planned that
\begin{equation*}
    \begin{split}
        \mbox{mean(B)} & =0.90\pm0.00=p_a \\
        \mbox{std(B)} & =0.30\pm0.00=\sqrt{0.9(1-0.9)}=\sqrt{p_b(1-p_b)}
    \end{split}
\end{equation*}
Table~\ref{tab:B counts} gives the number of observations for each value of $B$0. By construction, as $B \sim \mathcal{B}(0.9)$, there is an imbalance: about $90\%$ individuals have $B=1$ and $10\%$ individuals have $B=0$0. So the group imbalance ratio is
\begin{equation*}
    IR_A = 8.99\pm0.02
\end{equation*}
meaning there are almost 9 times more observations of $B=0$ than $B=1$0.

\begin{table}[H]
    \centering
    \begin{tabular}{|c|c|}
        \hline
        B & Observations \\ \hline
        0 & $89,980.70\pm207.11$ \\ \hline
        1 & $10,019.30\pm207.11$ \\ \hline
    \end{tabular}
    \caption{Number of observations by value of $B$}
    \label{tab:B counts}
\end{table}

\paragraph{Y} We computed it to follow a Bernoulli distribution of parameter $p_y=0.2$, so as expected
\begin{equation*}
    \begin{split}
        \mbox{mean(Y)} & =0.20\pm0.00=p_y \\
        \mbox{std(A)} & =0.40\pm0.00=\sqrt{0.2(1-0.2)}=\sqrt{p_y(1-p_y)}
    \end{split}
\end{equation*}
This results in an imbalance too: table~\ref{tab:Y counts} gives the number of observations for each value of $Y$, and the imbalance ratio is
\begin{equation*}
    IR_Y = 4.00\pm0.01
\end{equation*}
meaning there are 4 times more observations of $Y=0$ than $Y=1$0.

\begin{table}[H]
    \centering
    \begin{tabular}{|c|c|}
        \hline
        Y & Observations \\ \hline
        0 & $80,010.49\pm262.28$ \\ \hline
        1 & $19,989.51\pm262.28$ \\ \hline
    \end{tabular}
    \caption{Number of observations by value of $Y$}
    \label{tab:Y counts}
\end{table}

\paragraph{Going back on imbalance} In real life, imbalance can be explained either by the way the data was collected or by the natural domination of one class. The collection of data can lead to an imbalance if the sampling is biased or if mistakes are made, for examples writing down the wrong labels on observations. In insurance, there are sampling biases: conclusions about risks only concern individuals who have been accepted at the underwriting stage. As underwriters aim at selecting `good' risks, most of the time, the insurer's portfolio will be very specific and predictions on claims frequency, for example, cannot be generalized to a different population.

\begin{figure}[H]
    \centering
    \includegraphics[scale=0.6]{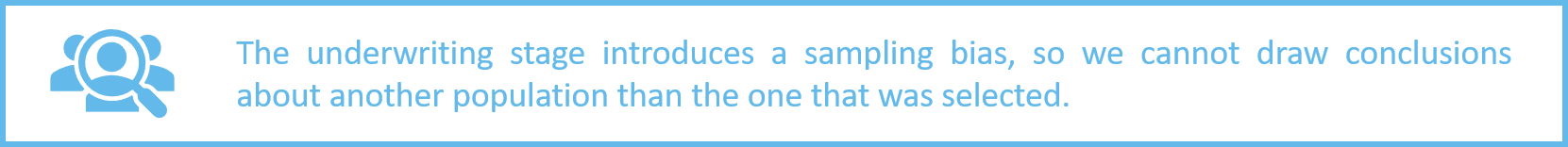}
\end{figure}

\subsubsection{Multivariate analysis}
\paragraph{} An important part of data exploration consists in studying the relationships between variables. To do so, we will analyze how they are correlated to each other. As a reminder, correlation is how linearly related two variables are, and is only the first order approximation of dependence, as seen previously.

\begin{figure}[H]
    \centering
    \includegraphics[scale=0.7]{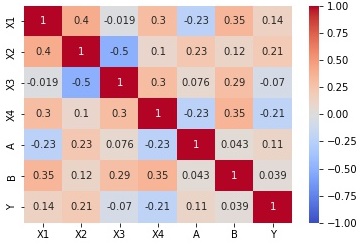}
    \caption{Heatmap of correlations}
    \label{fig:heatmap corr}
\end{figure}

\paragraph{Correlations with Y} A heatmap of the correlations can help understand which variables are correlated with each other. Figure~\ref{fig:heatmap corr} shows the heatmap of correlations. The strongest positive correlations are colored in bright red and the strongest negative correlations are colored in bright blue. The strongest correlations to $Y$ in absolute value are with $X^{(4)}$, $X^{(2)}$, $X^{(1)}$, $A$, $X^{(2)}$, $X^{(3)}$ and then $B$0. This is coherent with the theoretical values of the correlation matrix $R_{X,A,B,Y}$ in section~\ref{theoretical corr R_xa}.

\paragraph{Correlations with A} We saw in the previous paragraph that $A$ has the third strongest absolute value correlation to $Y$0. It is also strongly correlated to the $X^{(i)}$, as we see in figure~\ref{fig:heatmap corr}. We noted in the previous section that there are imbalances in the number of observations for both $A$ and $Y$, and we see in table~\ref{tab:A Y counts} that among group $A$ there is an imbalance in output values too. If we compute the imbalance ratio for the output, we get
\begin{equation*}
    \begin{split}
        \mbox{A=0: } & IR_Y=4.86\pm0.01 \\
        \mbox{A=1: } & IR_Y=2.73\pm0.01
    \end{split}
\end{equation*}
meaning that within group $A=0$, there are 4.86 times more observations of $Y=0$ than $Y=1$ and within group $A=1$, there are 2.73 times more observations of $Y=0$ than $Y=1$0. In general there are a lot more $Y=0$ outputs than $Y=1$ ones, and when we zoom in on protected groups, the imbalance ratio is larger for group $A=0$ than for $A=1$, meaning that the former has a larger proportion of $Y=0$ outputs than the latter.

\begin{table}[H]
    \centering
    \begin{tabular}{c|c|c|}
        \cline{2-3}
         & Y=0 & Y=1 \\ \hline
        \multicolumn{1}{|c|}{A=0} & $58,048.36\pm33.08$ & $11,953.86\pm21.76$ \\ \hline
        \multicolumn{1}{|c|}{A=1} & $21,962.13\pm26.37$ & $8,035.65\pm15.49$ \\ \hline
    \end{tabular}
    \caption{Number of observations by values of $A$ and $Y$}
    \label{tab:A Y counts}
\end{table}

\paragraph{Correlations with B} We saw that $B$ has the weakest absolute value correlation to $Y$, but it is still strongly correlated to the $X^{(i)}$0. Within groups, as for $A$, there can be imbalances, as we see in table~\ref{tab:B Y counts}. The imbalance ratios by group are
\begin{equation*}
    \begin{split}
        \mbox{B=0: } & IR_Y=5.53\pm0.02 \\
        \mbox{B=1: } & IR_Y=3.88\pm0.01
    \end{split}
\end{equation*}
meaning that within group $B=0$, there are 5.53 times more observations of $Y=0$ than $Y=1$ and within group $B=1$, there are 3.88 times more observations of $Y=0$ than $Y=1$0. The imbalance ratio is larger for group $B=0$ than for $B=1$0.

\begin{table}[H]
    \centering
    \begin{tabular}{c|c|c|}
        \cline{2-3}
         & Y=0 & Y=1 \\ \hline
        \multicolumn{1}{|c|}{B=0} & $8,484.30\pm19.82$ & $1,535.00\pm9.06$ \\ \hline
        \multicolumn{1}{|c|}{B=1} & $71,526.19\pm29.28$ & $18,454.51\pm24.25$ \\ \hline
    \end{tabular}
    \caption{Number of observations by values of $B$ and $Y$}
    \label{tab:B Y counts}
\end{table}

\begin{figure}[H]
    \centering
    \includegraphics[scale=0.4]{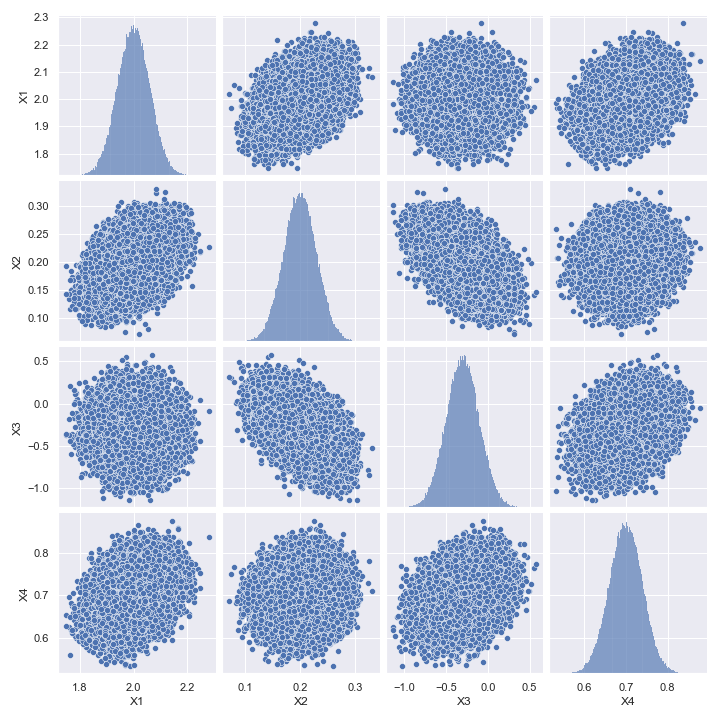}
    \caption{Pairplots of the $X^{(i)}$}
    \label{fig:pairplots xi}
\end{figure}

\paragraph{Correlations between the $\mathbf{X^{(i)}}$} By construction the $X^{(i)}$ are related to each other, through the correlation matrix. Figure~\ref{fig:pairplots xi} gives the pairwise relationships between the $X^{(i)}$, and the diagonal is their marginal distribution. We observe that $X^{(1)}$ is positively correlated with $X^{(2)}$ and $X^{(3)}$, $X^{(2)}$ is negatively correlated with $X^{(3)}$, and $X^{(3)}$ is negatively correlated with $X^{(4)}$0. Correlations between other variables are less obvious. \par

\subsubsection{Outliers}
\paragraph{} As the $X^{(i)}$ were computed to follow normal distributions, and $A$, $B$ and $Y$ to take values in $\{0,1\}$, we do not expect to have any outliers. This is verified in our datasets: there are no unexpected values for the $X^{(i)}$ and we have values of $A$, $B$ and $Y$ in $\{0,1\}$0.

\newpage
{\Large\textbf{Discrimination mitigation applied to the simulated data}}
\section{Discrimination mitigation applied to the simulated data}
\paragraph{} The goal of the section is to compare different pre-processing steps and see how they influence our fairness metrics. After this, we will apply a logistic regression model to the simulated explanatory variables predict the variable of interest. Appendix~\ref{appendix:logistic regression} gives a reminder on the logistic regression. \par
The reason why we chose the logistic regression is that it is interpretable, which is a major issue with Machine Learning, and an important characteristic to simplify our study on fairness. It also presents other advantages, such as its simplicity with a low number of parameters. The main drawback is that a lot of preprocessing must be done: we need to select variables that are not strongly correlated with each other, and to transform most continuous variables into categorical variables, or only their general effect will be captured by the coefficients. For this simulated dataset, we will keep all variables as there are only 5 of then, and they are very simple. \par

\begin{figure}[H]
    \centering
    \includegraphics[scale=0.6]{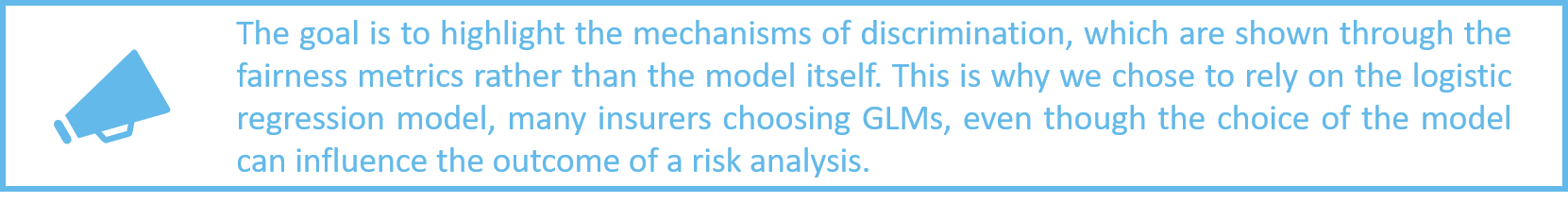}
\end{figure}

\subsection{Regression model with no pre-processing step}
\label{baseline}
\paragraph{} In this section, we will simply predict the outcome with all the variables. We standardized the variables to have mean zero and standard deviation 1, then randomly separated the dataset into a train (80\%) and a test dataset (20\%), and applied a logistic regression model. \par
Table~\ref{tab:weights_reg all variables} gives confidence intervals of the weights of this logistic regression, their standard errors and the associated p-values. All variables have p-values below 0.05, so they are all significant to the model. \par

\begin{table}[H]
    \centering
    \begin{tabular}{c|c|c|c}
        Variable & Coefficient & Standard Error & P-value \\ \hline
        Intercept & $-1.82\pm0.01$ & $0.05\pm0.00$ & $0.00\pm0.00$ \\
        $X_1$ & $0.55\pm0.00$ & $0.02\pm0.00$ & $0.00\pm0.00$ \\
        $X_2$ & $2.81\pm0.01$ & $0.05\pm0.00$ & $0.00\pm0.00$ \\
        $X_3$ & $0.25\pm0.00$ & $0.01\pm0.00$ & $0.00\pm0.00$ \\
        $X_4$ & $-2.53\pm0.01$ & $0.03\pm0.00$ & $0.00\pm0.00$ \\
        A & $-0.17\pm0.01$ & $0.03\pm0.00$ & $0.00\pm0.00$ \\
        B & $0.41\pm0.01$ & $0.04\pm0.00$ & $0.00\pm0.00$ \\
    \end{tabular}
    \caption{Coefficients of the model using all variables}
    \label{tab:weights_reg all variables}
\end{table}

\begin{figure}[H]
    \centering
    \includegraphics{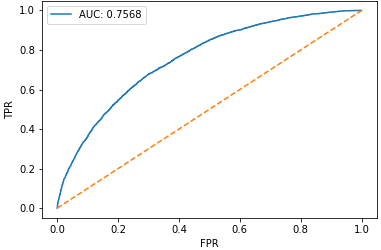}
    \caption{ROC curve for the model with all variables}
    \label{fig:ROC_sim_allvar}
\end{figure}

\paragraph{Performance evaluation}
\begin{itemize}
    \item We have 81.05\% correct classifications on average. The accuracy acceptability depends on the business context. As we saw previously, if mistakes have a high cost then it might not be sufficient. \\
    If there is a large class imbalance, accuracy is not the best metric as it can be very high while the model only fits the majority population. As a reminder, the output imbalance ratio is $IR_Y=4.00$0. So we cannot rely solely on accuracy to evaluate our model.
    \item As a reminder, the ROC (Receiver Operating Characteristic) curve plots the true positive rate against the false positive rate for varying classification thresholds. A random classifier will exhibit a linear ROC curve as the one plotted in a dashed orange line (figure~\ref{fig:ROC_sim_allvar}). Above that line, the model performs better than the random classifier, and below, worse. The perfect classifier has a ROC curve that is confined to the (0,1) point. Here, our model performs a lot better than the random classifier: the AUC (Area Under the ROC Curve) is of 0.7568.
\end{itemize}

\begin{table}[H]
    \centering
    \begin{tabular}{c|c}
         (\%) & Global \\ \hline
        Accuracy & $81.05\pm0.05$
    \end{tabular}
    \caption{Global metrics (all variables)}
    \label{tab:metrics all variables}
\end{table}

\paragraph{Fairness evaluation} We will compare fairness under all three definitions, first comparing the metrics between values of protected groups A and B (table~\ref{tab:fairness metrics all variables}), then between subgroups of combinations of A and B (table~\ref{tab:fairness metrics subgroups all variables}).
\begin{itemize}
    \item Statistical parity requires the same acceptance rates for all protected groups.
    \begin{itemize}
        \item It is a lot higher for group $A=0$ than for group $A=1$, by 8.26 points. So group $A=1$ is disadvantaged by the model under this definition.
        \item It is higher for group $B=0$ than for group $B=1$, by 3.47 points. So group $B=1$ is disadvantaged by the model under this definition.
        \item Looking at subgroups, the most advantaged subgroup is for $(A=0,B=0)$ and the most disadvantaged is for $(A=1,B=1)$0. This shows that looking at combinations of protected variables reveals that groups with a certain combination of characteristics are even more disadvantaged.
    \end{itemize} 
    \item Equal opportunity requires the same true positive rates for all protected groups.
    \begin{itemize}
        \item It is higher for group $A=0$, so group $A=1$ is disadvantaged by the model under this definition.
        \item It is higher for group $B=0$, although not by much, so group $B=1$ is disadvantaged by the model under this definition.
        \item It is highest for subgroup $(A=0,B=0)$ and lowest for group $(A=1,B=1)$0.
    \end{itemize} 
    \item Equalized odds requires the same true and false positive rates for both protected groups.
    \begin{itemize}
        \item The false positive rate is higher for group $A=0$ than group $A=1$0. Group $A=1$ has lower true and false positive rates, so it is disadvantaged by the model under this definition.
        \item The false positive rate is higher for group $B=0$ than $B=1$0. Group $B=1$ has lower true and false positive rates, so it is disadvantaged by the model under this definition.
        \item The false positive rate is highest for subgroup $(A=0,B=0)$ and lowest for $(A=1,B=1)$0.
    \end{itemize}
\end{itemize}
 To conclude, the model is unfair and disadvantages groups $A=1$ and $B=1$ under all three fairness definitions, and the most disadvantaged subgroup is $(A=1,B=1)$0.

\begin{table}[H]
    \centering
    \makebox[\linewidth]{
    \begin{tabularx}{19.5cm}{c|c|ccc|ccc}
        (\%) & Global & A=0 & A=1 & Difference & B=0 & B=1 & Difference \\ \hline
        AR & $94.13\pm0.05$ & $96.60\pm0.04$ & $88.34\pm0.10$ & $8.26\pm0.10$ & $97.25\pm0.09$ & $93.78\pm0.05$ & $3.47\pm0.10$ \\
        TPR & $96.98\pm0.04$ & $98.26\pm0.03$ & $93.60\pm0.09$ & $4.67\pm0.09$ & $98.62\pm0.06$ & $96.79\pm0.04$ & $1.83\pm0.07$ \\
        FPR & $82.69\pm0.15$ & $88.55\pm0.15$ & $73.95\pm0.24$ & $14.60\pm0.26$ & $89.72\pm0.39$ & $82.11\pm0.16$ & $7.62\pm0.42$
    \end{tabularx}
    }
    \caption{Fairness metrics globally and by protected group (all variables)}
    \label{tab:fairness metrics all variables}
\end{table}

\begin{table}[H]
    \centering
    \begin{tabular}{c|cc|cc}
        \multirow{2}{*}{(\%)} & \multicolumn{2}{c|}{A=0} & \multicolumn{2}{c}{A=1} \\ \cline{2-5}
         & \multicolumn{1}{c|}{B=0} & B=1 & \multicolumn{1}{c|}{B=0} & B=1 \\ \hline
        AR & \multicolumn{1}{c|}{$98.47\pm0.07$} & $96.39\pm0.04$ & \multicolumn{1}{c|}{$93.72\pm0.22$} & $87.92\pm0.12$ \\
        TPR & \multicolumn{1}{c|}{$99.21\pm0.05$} & $98.15\pm0.03$ & \multicolumn{1}{c|}{$96.58\pm0.19$} & $93.41\pm0.09$ \\
        FPR & \multicolumn{1}{c|}{$93.65\pm0.36$} & $88.08\pm0.16$ & \multicolumn{1}{c|}{$83.15\pm0.66$} & $73.34\pm0.26$
    \end{tabular}
    \caption{Fairness metrics by protected subgroups (all variables)}
    \label{tab:fairness metrics subgroups all variables}
\end{table}

\begin{figure}[H]
    \centering
    \includegraphics[scale=0.6]{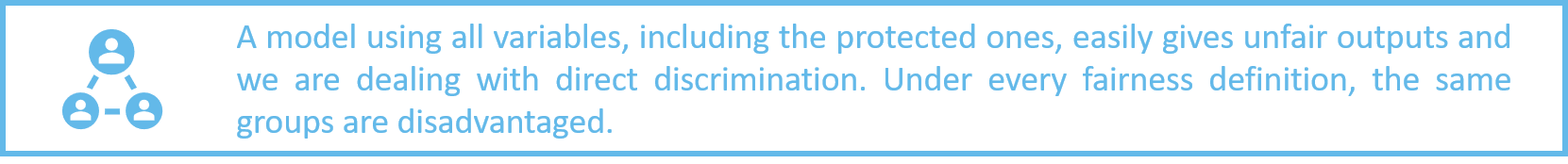}
\end{figure}

\subsection{Removing protected variables to avoid direct discrimination}
\paragraph{} When sensitive variables are omitted, models can still learn stereotypes, because sensitive information is embedded in datasets even if it is not intentional. Leaving out sensitive variables forces the correlated variables to take on a greater importance. This is the omitted variable bias \cite{williams18}. \par
Removing sensitive attributes is problematic, because it becomes impossible to check for bias and discrimination. We cannot see if the most important variables for prediction are strongly correlated with a protected attribute or compute metrics. This is a problem related to data regulations: as we saw in section~\ref{gdpr}, the GDPR requires minimal data collection, but more data is needed to prove discrimination. \par
We will preprocess our data by removing the sensitive attributes, $A$ and $B$, then predict the outcome with a logistic regression model. Table~\ref{tab:weights_reg without A} gives the predicted weights of this logistic regression. As $A$ and $B$ are no longer used as explanatory variables, the weights of regression have changed. For example, the biggest change is for $X^{(3)}$, which had a coefficient of 0.25 and now has a coefficient of 2.46. Remembering the correlation matrix, $X^{(3)}$ is correlated with A and B with relatively high Pearson correlation coefficients: -0.23 and 0.35 respectively. This can indicate that A and B will still indirectly play a part in the model predictions.

\begin{table}[H]
    \centering
    \begin{tabular}{c|c|c|c}
    \multicolumn{4}{c}{With all variables} \\
        Variable & Coefficient & Standard Error & P-value \\ \hline
        Intercept & $-1.82\pm0.01$ & $0.05\pm0.00$ & $0.00\pm0.00$ \\
        $X_1$ & $0.55\pm0.00$ & $0.02\pm0.00$ & $0.00\pm0.00$ \\
        $X_2$ & $2.81\pm0.01$ & $0.05\pm0.00$ & $0.00\pm0.00$ \\
        $X_3$ & $0.25\pm0.00$ & $0.01\pm0.00$ & $0.00\pm0.00$ \\
        $X_4$ & $-2.53\pm0.01$ & $0.03\pm0.00$ & $0.00\pm0.00$ \\
        A & $-0.17\pm0.01$ & $0.03\pm0.00$ & $0.00\pm0.00$ \\
        B & $0.41\pm0.01$ & $0.04\pm0.00$ & $0.00\pm0.00$ \\
        \multicolumn{4}{c}{} \\
        \multicolumn{4}{c}{Without protected variables} \\
        Variable & Coefficient & Standard Error & P-value \\ \hline
        Intercept & $-1.76\pm0.01$ & $0.05\pm0.00$ & $0.00\pm0.00$ \\
        $X_1$ & $0.65\pm0.00$ & $0.02\pm0.00$ & $0.00\pm0.00$ \\
        $X_2$ & $2.68\pm0.01$ & $0.05\pm0.00$ & $0.00\pm0.00$ \\
        $X_3$ & $2.46\pm0.00$ & $0.01\pm0.00$ & $0.00\pm0.00$ \\
        $X_4$ & $-2.41\pm0.01$ & $0.03\pm0.00$ & $0.00\pm0.00$ \\
    \end{tabular}
    \caption{Weights of the logistic regression}
    \label{tab:weights_reg without A}
\end{table}

\begin{figure}[H]
    \centering
    \includegraphics[scale=0.6]{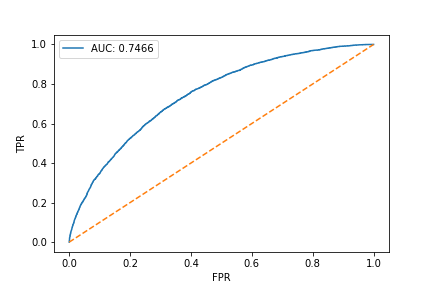}
    \caption{ROC curve for the model without protected variables}
    \label{fig:ROC_sim_withoutprotected}
\end{figure}

\paragraph{Performance evaluation}
\begin{itemize}
    \item As we can see in table~\ref{tab:metrics without A}, the accuracy has only decreased by 0.01 on average, which is negligible, especially considering the width of the confidence interval.
    \item Looking at the ROC curve in figure~\ref{fig:ROC_sim_withoutprotected}, the model still performs quite well, and the AUC has decreased from 0.7568 to 0.7466 compared to the model using all variables.
\end{itemize}

\begin{table}[H]
    \centering
    \begin{tabular}{c|c|c}
         (\%) & With all variables & Without protected variables \\ \hline
        Accuracy & $81.05\pm0.05$ & $81.04\pm0.05$
    \end{tabular}
    \caption{Global metrics}
    \label{tab:metrics without protected}
\end{table}

\paragraph{Fairness evaluation}
\begin{itemize}
    \item Acceptance rate \begin{itemize}
        \item The global acceptance rate has slightly increased.
        \item  For groups A, the gap between acceptance rates has increased by more than one point. This was expected, as $X^{(3)}$, strongly correlated with A, has taken on a great importance in the model prediction.
        \item For groups B, the difference in acceptance rates has decreased.
        \item Looking at protected subgroups, the most advantaged subgroups are the same as in the model with all variables: the most advantaged subgroup is $(A=0,B=0)$ and the most disadvantaged is $(A=1,B=1)$0. We now have a lower gap between acceptance rates for subgroups $(A=0,B=0)$ and $(A=0,B=1)$ and for $(A=1,B=0)$ and $(A=1,B=1)$, which is coherent as acceptance rates between groups B are closer than previously.
    \end{itemize}
    \item For true and false positive rates, we have the same conclusions as for the acceptance rates.
\end{itemize}
For groups A, the gaps between fairness metrics are wider when the protected variables are not used in the model, but for group B, they are smaller, although this phenomenon comes from the structure of the correlation matrix. The same groups remain disadvantaged.
\paragraph{} To conclude, the performance metrics have not deteriorated too much compared to when using all variables. The fairness metrics are worse for groups A but better when looking at groups B. To conclude, simply ignoring the protected variables is not a solution. \par

\begin{table}[H]
    \centering
    \captionsetup{justification=centering}
    \makebox[\linewidth]{
    \begin{tabularx}{19.5cm}{c|c|ccc|ccc}
        \multicolumn{8}{c}{With all variables} \\
        (\%) & Global & A=0 & A=1 & Difference & B=0 & B=1 & Difference \\ \hline
        AR & $94.13\pm0.05$ & $96.60\pm0.04$ & $88.34\pm0.10$ & $8.26\pm0.10$ & $97.25\pm0.09$ & $93.78\pm0.05$ & $3.47\pm0.10$ \\
        TPR & $96.98\pm0.04$ & $98.26\pm0.03$ & $93.60\pm0.09$ & $4.67\pm0.09$ & $98.62\pm0.06$ & $96.79\pm0.04$ & $1.83\pm0.07$ \\
        FPR & $82.69\pm0.15$ & $88.55\pm0.15$ & $73.95\pm0.24$ & $14.60\pm0.26$ & $89.72\pm0.39$ & $82.11\pm0.16$ & $7.62\pm0.42$ \\
        \multicolumn{8}{c}{} \\
        \multicolumn{8}{c}{Without protected variables} \\
        (\%) & Global & A=0 & A=1 & Difference & B=0 & B=1 & Difference \\ \hline
        AR & $94.29\pm0.05$ & $97.05\pm0.04$ & $87.55\pm0.10$ & $9.50\pm0.09$ & $95.38\pm0.10$ & $94.07\pm0.055$ & $1.31\pm0.09$ \\
        TPR & $97.02\pm0.03$ & $98.51\pm0.03$ & $93.07\pm0.09$ & $5.44\pm0.08$ & $97.41\pm0.08$ & $96.97\pm0.04$ & $0.44\pm0.08$ \\
        FPR & $82.92\pm0.12$ & $89.94\pm0.13$ & $72.53\pm0.23$ & $17.41\pm0.26$ & $84.15\pm0.46$ & $82.82\pm0.13$ & $1.33\pm0.48$ \\
    \end{tabularx}
    }
    \caption{Fairness metrics}
    \label{tab:metrics without A}
\end{table}

\begin{table}[H]
    \centering
    \begin{tabular}{c|cc|cc}
        \multicolumn{5}{c}{With all variables} \\
        \multirow{2}{*}{(\%)} & \multicolumn{2}{c|}{A=0} & \multicolumn{2}{c}{A=1} \\ \cline{2-5}
         & \multicolumn{1}{c|}{B=0} & B=1 & \multicolumn{1}{c|}{B=0} & B=1 \\ \hline
        AR & \multicolumn{1}{c|}{$98.47\pm0.07$} & $96.39\pm0.04$ & \multicolumn{1}{c|}{$93.72\pm0.22$} & $87.92\pm0.12$ \\
        TPR & \multicolumn{1}{c|}{$99.21\pm0.05$} & $98.15\pm0.03$ & \multicolumn{1}{c|}{$96.58\pm0.19$} & $93.41\pm0.09$ \\
        FPR & \multicolumn{1}{c|}{$93.65\pm0.36$} & $88.08\pm0.16$ & \multicolumn{1}{c|}{$83.15\pm0.66$} & $73.34\pm0.26$ \\
        \multicolumn{5}{c}{} \\
        \multicolumn{5}{c}{Without protected variables} \\
        \multirow{2}{*}{(\%)} & \multicolumn{2}{c|}{A=0} & \multicolumn{2}{c}{A=1} \\ \cline{2-5}
         & \multicolumn{1}{c|}{B=0} & B=1 & \multicolumn{1}{c|}{B=0} & B=1 \\ \hline
        AR & \multicolumn{1}{c|}{$97.43\pm0.08$} & $97.00\pm0.04$ & \multicolumn{1}{c|}{$88.94\pm0.27$} & $87.43\pm0.10$ \\
        TPR & \multicolumn{1}{c|}{$98.56\pm0.06$} & $98.50\pm0.03$ & \multicolumn{1}{c|}{$93.41\pm0.25$} & $93.04\pm0.09$ \\
        FPR & \multicolumn{1}{c|}{$90.01\pm0.44$} & $89.92\pm0.13$ & \multicolumn{1}{c|}{$72.61\pm0.87$} & $72.53\pm0.24$
    \end{tabular}
    \caption{Fairness metrics by protected subgroups}
    \label{tab:fairness metrics subgroups without protected}
\end{table}

\begin{figure}[H]
    \centering
    \includegraphics[scale=0.6]{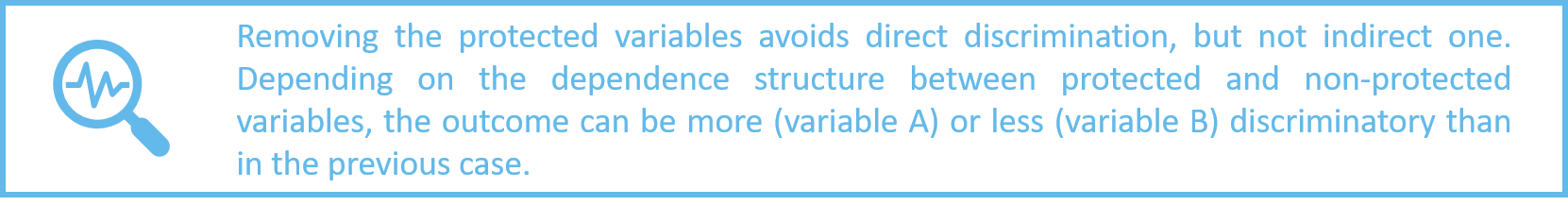}
\end{figure}

\subsection{Transforming the non-protected variables to mitigate indirect discrimination}
\subsubsection{Theory}
\paragraph{The idea} Focusing on the definition of fairness as statistical parity, we can view it as an independence condition. Statistical parity requires $\hat{Y} \indep A,B$0. Since we do not want $A$ or $B$ to impact the predicted output, the goal is firstly not to use them as explanatory variables and secondly to have explanatory variables that are independent of them. Dealing with independence is a complex problem, which is why we will tackle it on a linear level only - with correlation. We will try to obtain transformations of the variables $X^{(i)}$ uncorrelated with $A$ and $B$, and we will then use them as inputs of a logistic regression model to predict $Y$0. \par
Drawing inspiration from the Gram-Schmidt process (a reminder is given in appendix~\ref{appendix:gram schmidt}), the idea is to view the variables as vectors in a n-dimensional space (n being the number of variables in our dataset) and the covariance between them as a scalar product. Let us set the theoretical framework. \par

\begin{figure}[H]
    \centering
    \includegraphics[scale=0.6]{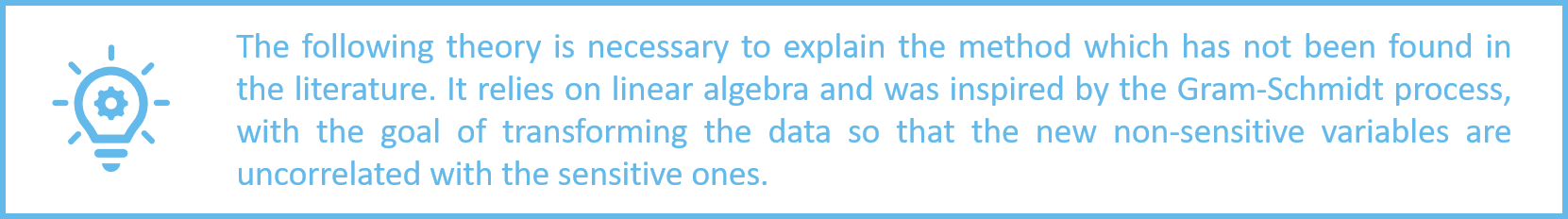}
\end{figure}

\begin{definition} An inner product is a map
\begin{equation*}
    \langle \cdot,\cdot \rangle : V\times V\rightarrow \mathbb{R}
\end{equation*}
with $V$ a real vector space, that is symmetric, bilinear and positive-definite.
\end{definition}

\begin{proposition} Set $X$ and $Y$ real random variables with zero mean and finite variance. Their covariance
\begin{equation*}
    \langle X,Y \rangle =cov(X,Y)
\end{equation*}
is an inner product (on the space of random variables with zero mean and finite variance).
\end{proposition}
\begin{proof}
Set $X,Y,Z$ real random variables with zero mean and finite variance and $(a,b)\in\mathbb{R}^2$0. Then $cov(X,Y)=\mathbb{E}[XY]-\mathbb{E}[X]\mathbb{E}[Y]=\mathbb{E}[XY]$0. We will check the three properties of the inner product:
\begin{itemize}
    \item symmetry: $\mathbb{E}[XY]=\mathbb{E}[YX]$
    \item bilinearity: $\mathbb{E}[(aX+bY)Z]=a\mathbb{E}[XZ]+b\mathbb{E}[YZ]$
    \item positive-definiteness: $\mathbb{E}[XX] =\mathbb{E}[X^2]\geq0$ \\
    and $\mathbb{E}[X^2]=0\iff Var(X)+(\mathbb{E}[X])^2=0\iff \left\{ \begin{array}{l}
        Var(X)=0 \\
        \mathbb{E}[X]=0
    \end{array} \right. \iff X=0 \mbox{ a.s.}$
\end{itemize}
\end{proof}

\begin{definition} Two vectors are orthogonal if their inner product is zero. \end{definition}

As covariance is an inner product on the space of random variables with zero mean and finite variance, two random variables of this space are orthogonal if their covariance is zero ie if their correlation is zero, because as we saw in definition~\ref{def:pearson corr},
\begin{equation*}
    corr(X,Y)=\frac{cov(X,Y)}{\sqrt{Var(X)Var(Y)}}
\end{equation*}
and their variance is finite. So we have an interpretation of random variables of this space as vectors with an inner product. \par
Going back to the initial goal, we wanted a transformation of our variables $X^{(i)}$ that is uncorrelated to both $A$ and $B$0. With the framework we have set, it means that the transformed variables will be orthogonal to $A$ and to $B$0. Suppose the (non-orthogonal) basis of our vector space is the linearly independent set $B=\{u_1,\dots,u_n\}$ such that any vector Z of the space - which corresponds to the characteristic of one individual ie row of the dataset - can be uniquely written as a linear combination of the vectors of the basis:
\begin{equation*}
    Z=\sum_{i=1}^n x_iu_i \mbox{ with } X=\begin{bmatrix}
    x_1 \\
    \dots \\
    x_n
    \end{bmatrix}
\end{equation*}

Our goal is to find a change of basis that gives us the new basis $B'=\{v_1,\dots,v_n\}$0.
We will then be able to write 
\begin{equation*}
    Z=\sum_{j=1}^n x_j'v_j \mbox{ with } X'=\begin{bmatrix}
    x_1' \\
    \dots \\
    x_n'
    \end{bmatrix}
\end{equation*}
$A=(a_{i,j})_{i,j}$ is called the transition matrix and its $\mbox{j}^{th}$ row is formed by the coordinates of $v_j$:
\begin{equation*}
    A=\begin{bmatrix}
    a_{1,1} & a_{1,2} & \dots & a_{1,n} \\
    a_{2,1} & a_{2,2} & \dots & \\
    \dots & & & \\
    a_{n,1} & \dots & & a_{n,n}
    \end{bmatrix}
\end{equation*}
Then, the change-of-basis formula gives in matrix form
\begin{equation*}
    \begin{split}
        X' & =AX \\
         & =\begin{bmatrix}
    a_{1,1} & a_{1,2} & \dots & a_{1,n} \\
    a_{2,1} & a_{2,2} & \dots & \\
    \dots & & & \\
    a_{n,1} & \dots & & a_{n,n}
    \end{bmatrix}\begin{bmatrix}
    x_1 \\
    x_2 \\
    \dots \\
    x_n
    \end{bmatrix} \\
    & = \begin{bmatrix}
    a_{1,1}x_1+a_{1,2}x_2+\dots+a_{1,n}x_n \\
    a_{2,1}x_1+a_{2,2}x_2+\dots+a_{2,n}x_n \\
    \dots \\
    a_{n,1}x_1+a_{n,2}x_2+\dots+a_{n,n}x_n
    \end{bmatrix}
    \end{split}
\end{equation*}
So
\begin{equation*}
    x_i'=\sum_{j=1}^n a_{i,j}x_j
\end{equation*}

\par
Our goal is to have a new basis in which the non-sensitive vectors are orthogonal to the sensitive ones. For this reason, we impose the following constraints for the construction of the new basis:
\begin{enumerate}
    \item we do not want to transform the sensitive variables, so the first s vectors of the new basis will remain the same as in the old basis: $x_1=x_1',\dots,x_s=x_s'$
    \begin{equation*}
        \begin{split}
            x_i=x_i' & \Leftrightarrow x_i=\sum_{j=1}^n a_{i,j}x_j \\
             & \Leftrightarrow \left\{
            \begin{array}{l}
                a_{i,i}=1 \\
                a_{i,j}=0 \mbox{ for } j\neq i
            \end{array}
            \right.
        \end{split}
    \end{equation*}
    As a result,
    \begin{equation*}
        A=\left[ \begin{array}{cccccc}
            \multicolumn{3}{c}{\multirow{3}{*}{$I_s$}} & 0 & \dots & 0 \\
            \multicolumn{3}{c}{} & \vdots &  & \vdots \\
            \multicolumn{3}{c}{} & 0 & \dots & 0 \\
            a_{s+1,1} & \multicolumn{4}{c}{\dots} & a_{s+1,n} \\
            \vdots &  &  &  &  & \vdots \\
            a_{n,1} & \multicolumn{4}{c}{\dots} & a_{n,n} \\
      \end{array} \right]
    \end{equation*}
    with $I_s$ the identity matrix of size $s$0.
    \item the new non-sensitive vectors will be orthogonal to the sensitive vectors:
    $\forall j\in \llbracket1,s\rrbracket, \forall k\in \llbracket s+1,n\rrbracket, \langle v_j,v_k \rangle=0$ \\
    This is equivalent to: for all $k\in \llbracket s+1,n\rrbracket$ a system of $s$ equations with $n$ unknown variables (the $a_{.,k}$):
\begin{equation} \label{eq:syst}
    \left\{
    \begin{array}{ll}
        \langle v_1,v_k \rangle & =0 \\
        \multicolumn{2}{c}{\dots} \\
        \langle v_s,v_k \rangle & =0
    \end{array}
\right.
\end{equation}
The rows of $A$ are the $v_j$, so $v_j=\left\{
    \begin{array}{ll}
        u_j & \mbox{ if } j\in \llbracket1,s\rrbracket \\
        \sum_{i=1}^n a_{j,i}u_i & \mbox{ if } j\in \llbracket s+1,n\rrbracket
    \end{array}
\right.$
\begin{equation*}
    \begin{split}
        (\ref{eq:syst}) & \Leftrightarrow \left\{
        \begin{array}{ll}
            \langle u_1,\sum_{j=1}^n a_{k,j}u_j \rangle & =0 \\
            \multicolumn{2}{c}{\dots} \\
            \langle u_s,\sum_{j=1}^n a_{k,j}u_j \rangle & =0
        \end{array}
        \right. \\
         & \Leftrightarrow \left\{
        \begin{array}{ll}
            \sum_{j=1}^n a_{k,j} \langle u_1,u_j \rangle & =0 \\
            \multicolumn{2}{c}{\dots} \\
            \sum_{j=1}^n a_{k,j} \langle u_s,u_j \rangle & =0
        \end{array}
        \right.
    \end{split}
\end{equation*}
\end{enumerate}

For each row k of A, we are looking for the values of the $a_{k,1},\dots,a_{k,n}$0. The previous constraints give for each row a system of $s$ equations, but we have $n$ unknown variables, so the system is underdetermined, with an infinite number of solutions. To simplify, we can set
\begin{equation*}
    \forall j \in \{s+1,\dots,k-1,k+1,\dots,n\}, a_{k,j}=0 \mbox{ ie } v_k=\sum_{j=1}^{s} a_{k,j}u_j+a_{k,k}u_k
\end{equation*}
Meaning that each non-sensitive vector of the new basis is written as a linear combination of itself and of the sensitive vectors of the old basis. This gives a transition matrix of the shape
\begin{equation*}
    A=\left[ \begin{array}{cccccccc}
        \multicolumn{3}{c}{\multirow{3}{*}{$I_s$}} & 0 & \multicolumn{3}{c}{\dots} & 0 \\
        \multicolumn{3}{c}{} & \vdots &  &  &  & \vdots \\
        \multicolumn{3}{c}{} & 0 & \multicolumn{3}{c}{\dots} & 0 \\
        a_{s+1,1} & \dots & a_{s+1,s} & a_{s+1,s+1} & 0 & 0 & \dots & 0 \\
        a_{s+2,1} & \dots & a_{s+2,s} & 0 & a_{s+2,s+2} & 0 & \dots & 0 \\
        \vdots \\
        a_{n,1} & \dots & a_{n,s} & 0 & 0 & \dots &  0 & a_{n,s+1}
    \end{array} \right]
\end{equation*}

We now have for each row, a system of $s$ linear equations and $s+1$ unknown variables. We need one more constraints in order to have a unique solution. An idea is to minimize the distance between the old and the new basis (non-sensitive) vectors:
\begin{equation*}
    \min_{a_{k,1},\dots,a_{k,n}} d(u_k,v_k)
\end{equation*}
As the distance is positive, it has a lower bound, so this problem has a solution. We have defined the inner product as the covariance between random variables with zero mean and finite variance, so the distance between two such random variables $X$ and $Y$ is
\begin{equation*}
    d(X,Y)=\langle X-Y,X-Y\rangle=cov(X-Y,X-Y)=Var(X-Y)
\end{equation*}
So
\begin{equation*}
    \begin{split}
        d(u_k,v_k) & =d(u_k,\sum_{j=1}^{s} a_{k,j}u_j+a_{k,k}u_k) \\
         & =\langle (1-a_{k,k})u_k-\sum_{j=1}^{s} a_{k,j}u_j-a_{k,k}u_k,(1-a_{k,k})u_k-\sum_{j=1}^{s} a_{j,k}u_j \rangle \\
         & =((1-a_{k,k})u_k-\sum_{j=1}^{s} a_{k,j}u_j)^T((1-a_{k,k})u_k-\sum_{j=1}^{s} a_{k,j}u_j) \\
         & =\sum_{i=1}^n ((1-a_{k,k})u_{i,k}-\sum_{j=1}^{s} a_{k,j}u_{i,j})^2
    \end{split}
\end{equation*}
As we had a system of $s$ linear equations and $s+1$ unknown variables, we can express $a_{k,k}$ as a combination of the other $a_{.,k}$0. So the minimization of this distance gives a unique solution with the previous constraints. To summarize, for each $k=s+1,\dots,n$, we have the following minimization problem under constraints:
\begin{equation*}
        \min_{a_{k,1},\dots,a_{k,s},a_{k,k}} d(u_k,v_k) \mbox{ such that }\left\{
        \begin{array}{ll}
            \langle v_1,v_k \rangle & =0 \\
            \multicolumn{2}{c}{\dots} \\
            \langle v_s,v_k \rangle & =0
        \end{array}
        \right. \mbox{ with }  v_k=\sum_{j=1}^{s} a_{k,j}u_j+a_{k,k}u_k
\end{equation*}
Solving the problem for every $k=s+1,\dots,n$ gives us the transition matrix $A$0. Then, with $X$ the coordinates of a vector in the base $B$ and $X'$ in the base $B'$, we can write 
\begin{equation*}
    X'=AX 
\end{equation*}
Meaning that we will compute, for every observation, the transformation of each vector - corresponding to each individual - in the new basis. \par

\paragraph{Extreme-case scenarii}
We can wonder what would happen in the extreme-case scenario in which the non-sensitive variables are already uncorrelated with the sensitive ones. Then, we do not need to transform the non-sensitive variables: $A=I_n$ We have $\langle v_j,v_k\rangle=0$ for $j=1,\dots,s$ and $k=s+1,\dots,n$ and we have a minimal distance as $d(u_k,v_k)=d(u_k,u_k)=\langle u_k-u_k,u_k-u_k\rangle=0$0. \par
The other extreme-case scenario is the one in which all the variables are the non-sensitive variables are perfectly (positively or negatively) correlated with the sensitive ones. Then it means that the non-sensitive variables are a linear function of the sensitive ones, and consequently, of each other. It is therefore impossible to have non-sensitive vectors uncorrelated with the sensitive ones. Fortunately, in reality, when we have our datasets, we only have samples of `true' distributions, meaning that variables are never perfectly correlated with each other (except if a variables appears twice, but we can delete the duplicate). In the worst case, if the non-sensitive variables are very correlated to the non-sensitive ones, we will end up with transformed non-sensitive variables that have a very low variance, meaning that they will not explain the output very well.

\subsubsection{Results}
\begin{figure}[H]
    \centering
    \begin{subfigure}[b]{0.45\textwidth}
        \centering
        \includegraphics[width=\textwidth]{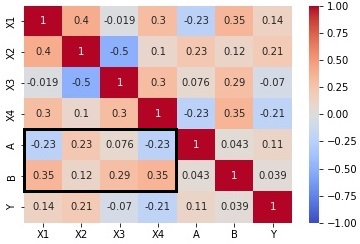}
        \caption{Before}
    \end{subfigure}
    \begin{subfigure}[b]{0.45\textwidth}
        \centering
        \includegraphics[width=\textwidth]{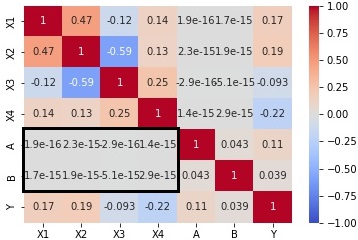}
        \caption{After}
    \end{subfigure}
    \caption{Heatmaps of correlations before and after transformation of the $X^{(i)}$}
    \label{fig:orthog proj heatmap}
\end{figure}

\paragraph{Transition matrix} The average transition matrix obtained on the 100 datasets is
\begin{equation*}
    A=\begin{bmatrix}
    1 & 0 & 0 & 0 & 0 & 0 \\
    0 & 1 & 0 & 0 & 0 & 0 \\
    -0.32 & 0.72 & 1 & 0 & 0 & 0 \\
    0.15 & 0.11 & 0 & 1 & 0 & 0 \\
    0.28 & 1.93 & 0 & 0 & 1 & 0 \\
    -0.21 & 0.48 & 0 & 0 & 0 & 1
    \end{bmatrix}\pm \begin{bmatrix}
    0 & 0 & 0 & 0 & 0 & 0 \\
    0 & 0 & 0 & 0 & 0 & 0 \\
    3\mathrm{e}{-3} & 6\mathrm{e}{-3} & 0 & 0 & 0 & 0 \\
    2\mathrm{e}{-3} & 3\mathrm{e}{-3} & 0 & 0 & 0 & 0 \\
    1\mathrm{e}{-2} & 2\mathrm{e}{-2} & 0 & 0 & 0 & 0 \\
    2\mathrm{e}{-3} & 3\mathrm{e}{-3} & 0 & 0 & 0 & 0
    \end{bmatrix}
\end{equation*}
This means that, on average,
\begin{equation*}
    A^{-1}=\begin{bmatrix}
    1 & 0 & 0 & 0 & 0 & 0 \\
    0 & 1 & 0 & 0 & 0 & 0 \\
    0.32 & -0.72 & 1 & 0 & 0 & 0 \\
    -0.15 & -0.11 & 0 & 1 & 0 & 0 \\
    -0.28 & -1.93 & 0 & 0 & 1 & 0 \\
    0.21 & -0.48 & 0 & 0 & 0 & 1
    \end{bmatrix}
\end{equation*}
\begin{equation*}
    \begin{split}
        X_1' & =A^{-1}X_1=.32A-0.72B+X_1 \\
        X_2' & =A^{-1}X_2=-0.15A-0.11B+X_2 \\
        X_3' & =A^{-1}X_3=-0.28A-1.93B+X_3 \\
        X_4' & =A^{-1}X_4=.21A-0.48B+X_4
    \end{split}
\end{equation*}
\paragraph{Transformed variables} Figure~\ref{fig:orthog proj heatmap} shows the correlations between variables before and after transforming the $X^{(i)}$0. We can see that the correlations between A (respectively B) and the $X{(i)}'$ have been reduced to zero, which was the goal of the procedure. Most correlations between other variables are close to before and after the change of basis, keeping the same signs and orders of magnitude. The most noticeable difference is that $corr(X^{(1)},X^{(3)})$ went from -0.019 to -0.12.

\begin{figure}[H]
    \centering
    \includegraphics[scale=0.6]{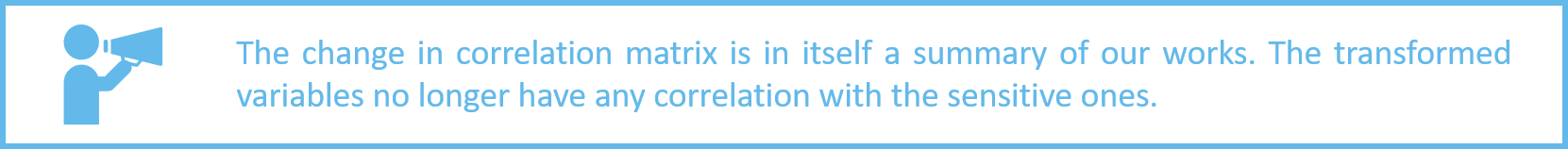}
\end{figure}

\paragraph{Prediction} We then apply the baseline model to predict the output, using only the transformed non-sensitive variables as explanatory variables.

\begin{table}[H]
    \centering
    \begin{tabular}{c|c|c|c}
        \multicolumn{4}{c}{Without protected variables} \\
        Variable & Coefficient & Standard Error & P-value \\ \hline
        Intercept & $-1.76\pm0.01$ & $0.05\pm0.00$ & $0.00\pm0.00$ \\
        $X_1$ & $0.65\pm0.00$ & $0.02\pm0.00$ & $0.00\pm0.00$ \\
        $X_2$ & $2.68\pm0.01$ & $0.05\pm0.00$ & $0.00\pm0.00$ \\
        $X_3$ & $2.46\pm0.00$ & $0.01\pm0.00$ & $0.00\pm0.00$ \\
        $X_4$ & $-2.41\pm0.01$ & $0.03\pm0.00$ & $0.00\pm0.00$ \\
        \multicolumn{4}{c}{} \\
        \multicolumn{4}{c}{Transformed variables} \\
        Variable & Coefficient & Standard Error & P-value \\ \hline
        Intercept & $-1.65\pm0.00$ & $0.01\pm0.00$ & $0.00\pm0.00$ \\
        $X_1$ & $0.55\pm0.00$ & $0.02\pm0.00$ & $0.00\pm0.00$ \\
        $X_2$ & $2.72\pm0.01$ & $0.05\pm0.00$ & $0.00\pm0.00$ \\
        $X_3$ & $0.24\pm0.00$ & $0.01\pm0.00$ & $0.00\pm0.00$ \\
        $X_4$ & $-2.46\pm0.01$ & $0.03\pm0.00$ & $0.00\pm0.00$
    \end{tabular}
    \caption{Weights of the logistic regression}
    \label{tab:weights_reg proj}
\end{table}

\begin{figure}[H]
    \centering
    \includegraphics{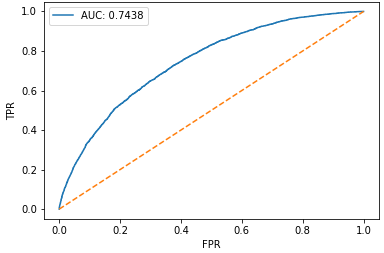}
    \caption{ROC curve for the model with transformed variables}
    \label{fig:ROC_sim_changebasis}
\end{figure}

\paragraph{Performance evaluation}
\begin{itemize}
    \item Table~\ref{tab:metrics basis change} gives the accuracy of the model. Compared to when simply deleting the protected variables, the accuracy decreases by only 0.23 points.
    \item Figure~\ref{fig:ROC_sim_changebasis} gives the ROC curve of the model. As a reminder, the AUC of the model without protected variables was of 0.7466. The AUC for the model with transformed variables has only decreased by 0.0028.
\end{itemize}

\begin{table}[H]
    \centering
    \begin{tabular}{c|c|c}
         (\%) & Without protected variables & With transformed variables \\ \hline
        Accuracy & $81.04\pm0.05$ & $80.81\pm0.05$
    \end{tabular}
    \caption{Global metrics}
    \label{tab:metrics basis change}
\end{table}

\paragraph{Fairness evaluation}
\begin{itemize}
    \item Acceptance rate \begin{itemize}
        \item The global acceptance rate has increased by 0.73 points on average, meaning that globally, more individuals get predicted the outcome $Y=0$ with the model using the transformed variables compared to the model without protected variables.
        \item For groups A: the difference in acceptance rates is now close to zero, which was the goal of the change of basis method. The reason why it is not exactly zero might come from the fact that we have approximated independence to correlation, so there might be some non linear dependence left between the transformed non-sensitive variables and A. It is interesting to note that the acceptance rate for group $A=0$ has decreased and for $A=1$ it has increased, and the advantageous position has shifted: the acceptance rate for group $A=0$ is now slightly lower than the one for group $A=1$0.
        \item For groups B: the difference in acceptance rates is now null, meaning that there we have $X^{(i)}\perp A$ ie independence.
        \item Looking at protected subgroups in figure~\ref{tab:fairness metrics subgroups change basis}, not all subgroups are treated fairly by the model as we have slight gaps between subgroups with $A=0$ and $A=1$0. The order of unfairness has also changed: now, the most disadvantaged subgroup is $(A=0,B=0)$ when it used to be the most advantaged one, and the most advantaged subgroup is $(A=1,B=1)$ when it used to be the most disadvantaged one.
        \end{itemize}
    \item True positive rate \begin{itemize}
        \item The global true positive rate has decreased compared to the model without protected variables.
        \item Groups A: as for the acceptance rate, the difference in true positive rates is now closer to zero, and the sign has changed, meaning that group $A=0$ now has a lower true positive rate than group $A=1$0.
        \item Groups B: the difference in true positive rates is now very close to zero and has also changed signs.
        \item As for the acceptance rate, the most advantaged subgroup now used to be the most disadvantaged and vice versa.
        \end{itemize}
    \item False positive rate \begin{itemize}
        \item Globally, the false positive rate has also decreased.
        \item Groups A: the difference in false positive rates has also decreased and changed signs.
        \item Groups B: surprisingly, the difference in false positive rates has increased and changed signs.
        \item Looking at subgroups, there has also been a shift in which subgroup is the most and least advantaged.
        \end{itemize}
\end{itemize}
To conclude, looking at the variable A, we have almost reached statistical parity, and for B, we have. We are closer to equal opportunity, although there was a shift in the advantage. For equalized odds, we are closer to fairness when looking at variable A but not B, and we have for both a shift in the advantage. All in all, the change of basis method has achieved the removal of linear dependence between the protected and non protected variables. We can also draw the conclusion that all three fairness definitions are not compatible.

\begin{table}[H]
    \centering
    \captionsetup{justification=centering}
    \makebox[\linewidth]{
    \begin{tabularx}{20cm}{c|c|ccc|ccc}
        \multicolumn{8}{c}{Without protected variables} \\
        (\%) & Global & A=0 & A=1 & Difference & B=0 & B=1 & Difference \\ \hline
        AR & $94.29\pm0.05$ & $97.05\pm0.04$ & $87.55\pm0.10$ & $9.50\pm0.09$ & $95.38\pm0.10$ & $94.07\pm0.055$ & $1.31\pm0.09$ \\
        TPR & $97.02\pm0.03$ & $98.51\pm0.03$ & $93.07\pm0.09$ & $5.44\pm0.08$ & $97.41\pm0.08$ & $96.97\pm0.04$ & $0.44\pm0.08$ \\
        FPR & $82.92\pm0.12$ & $89.94\pm0.13$ & $72.53\pm0.23$ & $17.41\pm0.26$ & $84.15\pm0.46$ & $82.82\pm0.13$ & $1.33\pm0.48$ \\
        \multicolumn{8}{c}{} \\
        \multicolumn{8}{c}{Transformed variables} \\
        (\%) & Global & A=0 & A=1 & Difference & B=0 & B=1 & Difference \\ \hline
        AR & $95.02\pm0.04$ & $94.91\pm0.05$ & $95.28\pm0.06$ & $-0.37\pm0.06$ & $95.02\pm0.11$ & $95.02\pm0.04$ & $0.00\pm0.01$ \\
        TPR & $97.38\pm0.03$ & $97.15\pm0.04$ & $97.97\pm0.05$ & $-0.82\pm0.05$ & $97.07\pm0.08$ & $97.41\pm0.03$ & $-0.34\pm0.08$ \\
        FPR & $85.58\pm0.13$ & $84.02\pm0.17$ & $87.92\pm0.17$ & $-3.90\pm0.21$ & $83.62\pm0.51$ & $85.74\pm0.13$ & $-2.13\pm0.50 $
    \end{tabularx}
    }
    \caption{Fairness metrics}
    \label{tab:metrics proj}
\end{table}

\begin{table}[H]
    \centering
    \begin{tabular}{c|cc|cc}
        \multicolumn{5}{c}{Without protected variables} \\
        \multirow{2}{*}{(\%)} & \multicolumn{2}{c|}{A=0} & \multicolumn{2}{c}{A=1} \\ \cline{2-5}
         & \multicolumn{1}{c|}{B=0} & B=1 & \multicolumn{1}{c|}{B=0} & B=1 \\ \hline
        AR & \multicolumn{1}{c|}{$97.43\pm0.08$} & $97.00\pm0.04$ & \multicolumn{1}{c|}{$88.94\pm0.27$} & $87.43\pm0.10$ \\
        TPR & \multicolumn{1}{c|}{$98.56\pm0.06$} & $98.50\pm0.03$ & \multicolumn{1}{c|}{$93.41\pm0.25$} & $93.04\pm0.09$ \\
        FPR & \multicolumn{1}{c|}{$90.01\pm0.44$} & $89.92\pm0.13$ & \multicolumn{1}{c|}{$72.61\pm0.87$} & $72.53\pm0.24$ \\
        \multicolumn{5}{c}{} \\
        \multicolumn{5}{c}{With transformed variables} \\
        \multirow{2}{*}{(\%)} & \multicolumn{2}{c|}{A=0} & \multicolumn{2}{c}{A=1} \\ \cline{2-5}
         & \multicolumn{1}{c|}{B=0} & B=1 & \multicolumn{1}{c|}{B=0} & B=1 \\ \hline
        AR & \multicolumn{1}{c|}{$94.87\pm0.11$} & $94.93\pm0.05$ & \multicolumn{1}{c|}{$95.18\pm0.23$} & $95.24\pm0.06$ \\
        TPR & \multicolumn{1}{c|}{$96.75\pm0.10$} & $97.20\pm0.04$ & \multicolumn{1}{c|}{$97.60\pm0.19$} & $97.99\pm0.04$ \\
        FPR & \multicolumn{1}{c|}{$82.62\pm0.52$} & $82.28\pm0.18$ & \multicolumn{1}{c|}{$86.34\pm0.69$} & $87.92\pm0.17$
    \end{tabular}
    \caption{Fairness metrics by protected subgroups}
    \label{tab:fairness metrics subgroups change basis}
\end{table}

\begin{figure}[H]
    \centering
    \includegraphics[scale=0.6]{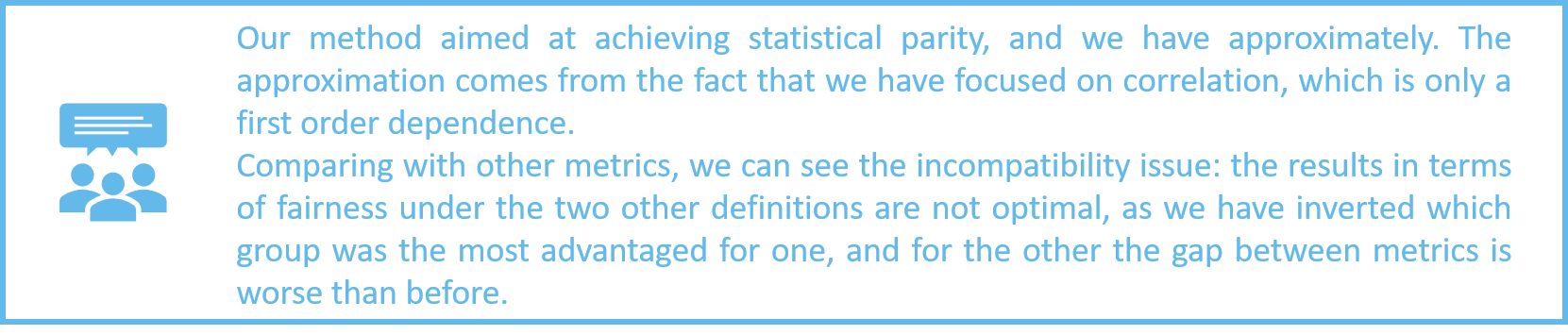}
\end{figure}

\subsection{Conclusion on the methods}

\paragraph{} We tested multiple preprocessing methods to mitigate unfairness. We compared the following: not doing anything, removing the protected variables, and transforming the non-sensitive variables in a way that they become uncorrelated to the sensitive ones. The goal of this last method was to approach fairness under the statistical parity, with equal acceptance rates for each protected group. \par
Comparing the model using all variables and the one using only non-protected variables, we can conclude that simply removing protected variables is not enough to reach fairness, under any definition. Depending on the correlation (and dependence) structure of the data, fairness can improve, with closer values of metrics for the different protected groups, as we saw is the case when looking at the B variable. But it can also magnify unfairness as we saw is the case when looking at the A variable. All in all, it is not a solution. \par
With our change of basis method, we ensure that the variables that will be used by the model are uncorrelated with the protected variables. Of course, uncorrelatedness is only an approximation of independence and there could still be non-linear relationships between the protected and transformed variables, which is why acceptance rates are only approximately equal for protected groups. \par
With this method, we have a slight decrease in accuracy, because we have no access to the information about the protected attributes, which were correlated with the outcome. \par
Now that we have studied a simulated dataset in which the correlation structure was known, we will take on a real dataset.

\newpage

{\Large\textbf{Use case: mortality of individuals with melanoma of the skin}}

\section{Use case: mortality of individuals with melanoma of the skin}
\paragraph{} In this section, we will study a real-life use case: the mortality of individuals with melanoma skin cancer. It is the 17th most common cancer worldwide, with over 300,000 new cases in 2020 \cite{WHO20}. In the US, the 5-year survival rate is of 93.7\% over the period of 2012 to 2018 \cite{SEER}. \par

\subsection{Some information about skin cancer}
\begin{figure}[H]
    \centering
    \includegraphics[scale=0.4]{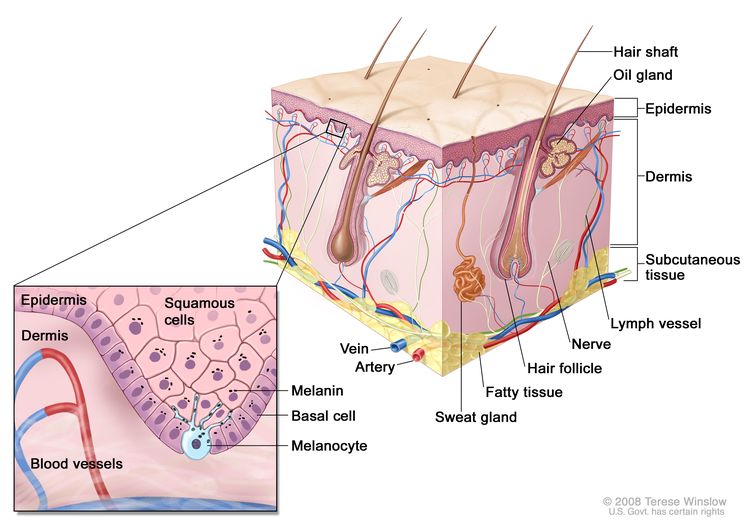}
    \caption{The anatomy of the skin \cite{SEER}}
    \label{fig:skin}
\end{figure}

\paragraph{} The skin, the body's largest organ, consists of several lays. The two main ones are the epidermis and the dermis, as shown in figure~\ref{fig:skin}. Skin cancer begins in the epidermis, which consists of three types of cells: squamous cells, basal cells and melanocytes. Melanocytes are cells that can make melanin, which is the pigment giving the skin its color. \par
Two different cancers can start in the skin: non-melanoma and melanoma, each representing about half of skin cancers. Non-melanoma skin cancer forms in the lower part of the epidermis or in squamous cells, but not in melanocytes. Melanoma forms in melanocytes and is more likely to spread out to other parts of the body. It can start in the skin, but also in mucous membranes such as parts of the eye. In this thesis, we focus on melanoma. \par

Medical literature \cite{SEER} identifies several risk factors for melanoma, including:
\begin{itemize}
    \item Fair complexion
    \item Exposure to natural or artificial sunlight
    \item Exposure to certain environmental factors (radiation, solvents, \dots)
    \item History of blistering sunburns
    \item Presence of several large or many small moles
    \item Family history of unusual moles or melanoma
    \item Weakened immune system
    \item Changes in genes linked to melanoma
\end{itemize} \par

Melanomas are mainly characterized by the location, thickness and ulceration \label{ulceration} (ie whether it has broken through the skin) of the tumor, the speed at which cancer cells divide, its spread to lymph nodes and other parts of the body. All of these factors can impact the severity of the condition. The TNM staging system is an internationally recognized standard for classifying cancers:
\begin{itemize}
    \item T describes the size of the primary tumor, as shown in figure~\ref{fig:skin T}:
    \begin{itemize}
        \item Tis: in situ
        \item T1: less than 1 mm thick
        \item T2: between 1 and 2 mm thick
        \item T3: between 2 and 4 mm thick
        \item T4: more than 4 mm thick
    \end{itemize}
\begin{figure}[H]
    \centering
    \includegraphics[scale=0.6]{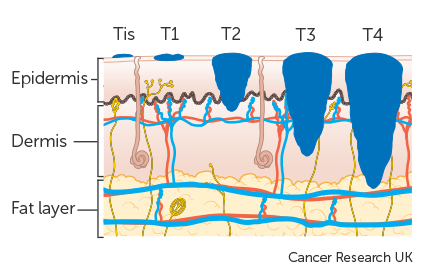}
    \caption{Tumor thickness \cite{TNM20}}
    \label{fig:skin T}
\end{figure}

    \item N describes the spread to regional lymph nodes:
    \begin{itemize}
        \item N0: no melanoma cells in the nearby lymph nodes
        \item N1: melanoma cells in one lymph node or in-transit, satellite or microsatellite metastases
        \item N2: melanoma cells in 2 or 3 lymph nodes or in one lymph node and in-transit, satellite or microsatellite metastases
        \item N3: melanoma cells in 4 or more lymph nodes or in 2 or 3 lymph nodes and in-transit, satellite or microsatellite metastases or in any number of lymph nodes stuck to each other (matted)
    \end{itemize}
    \item M describes the presence of metastasis (M0 for non-metastatic and M1 for metastatic).
\end{itemize}
The more well-known overall stage grouping describes the progression of a cancer thanks to five categories:
\begin{itemize}
    \item Stage 0: also called in situ, when abnormal melanocytes are found. They can become cancer and spread.
    \item Stage I: cancer has formed and is localized.
    \item Stage II: the cancer is locally advanced, in early stages.
    \item Stage III: the cancer is locally advanced, in late stages.
    \item Stage IV: the cancer has spread to other parts of the body which may be distant to the origin site, it is metastatic.
\end{itemize}
The link between the two staging systems is straightforward, as seen in table~\ref{tab:stage TNM}.
\begin{table}[H]
    \centering
    \begin{tabular}{|c|c|c|c|}
        \hline
        Stage & T & N & M \\ \hline
        0 & 0 & 0 & 0 \\ \hline
        I & 1-2 & 0 & 0 \\ \hline
        II & 3-4 & 0 & 0 \\ \hline
        III & 1-4 & 1-3 & 0 \\ \hline
        IV & 1-4 & 1-3 & 1 \\ \hline
    \end{tabular}
    \caption{Staging and its relation to the TNM system}
    \label{tab:stage TNM}
\end{table}

\subsection{Database presentation and mapping}

\paragraph{Presentation} In order to model the mortality of melanoma of the skin cancer patients, we used the public research SEER database. SEER is the Surveillance, Epidemiology, and End Results program of the National Cancer Institute of the United States that collects and publishes cancer information about around 48\% of the American population. This data collection process dates back to 1973, with around 400,000 new cases collected yearly in the most recent years. It is the largest and one of the most reliable cancer database, which makes it a dependable source of information for all types of studies. It is also representative of the US population in terms of measures of poverty and education. \par
The variables describe:
\begin{itemize}
    \item the patient: ID number, sex, age at diagnosis, year of diagnosis, origin, marital status at diagnosis, \dots
    \item the cancer (characteristics at diagnosis): type, site of origin, tumor size, spread to lymph nodes, metastatic state, stage, \dots
\end{itemize}

\begin{figure}[H]
    \centering
    \includegraphics[scale=0.6]{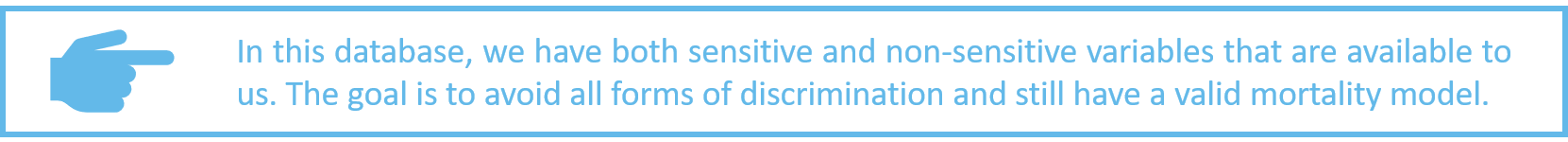}
\end{figure}

\paragraph{Mapping} This database is a huge source of information: there are 5,075,266 observations and 164 variables. Melanoma of the skin cancers represent about 4\% of all observations. Along the years, the way cancers are described has drastically changed, mostly because of the changes in classification of diseases standards. We therefore had to preprocess the data by studying the significance of variables in order to have the same standards throughout the years. For example, we had to regroup 8 different variables to determine the size of the tumor. \par
\begin{figure}[H]
    \centering
    \captionsetup{justification=centering}
    \begin{subfigure}[b]{0.6\textwidth}
        \centering
        \includegraphics[width=\textwidth]{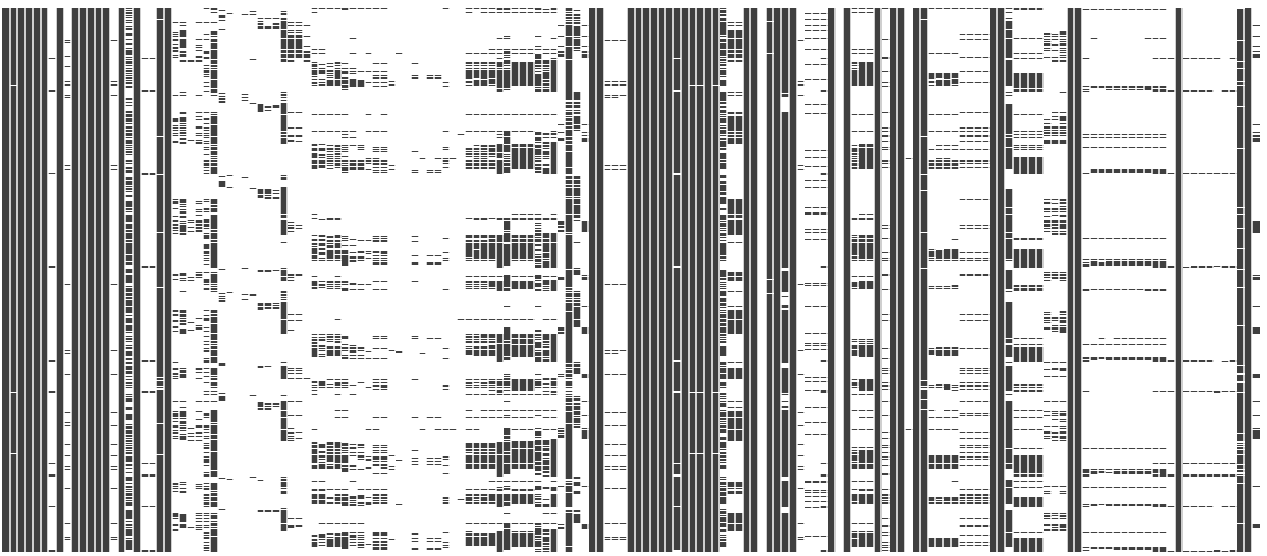}
        \caption{Raw: 5,075,266 rows, 164 columns}
    \end{subfigure}
    \\
    \begin{subfigure}[b]{0.6\textwidth}
        \centering
        \includegraphics[width=\textwidth]{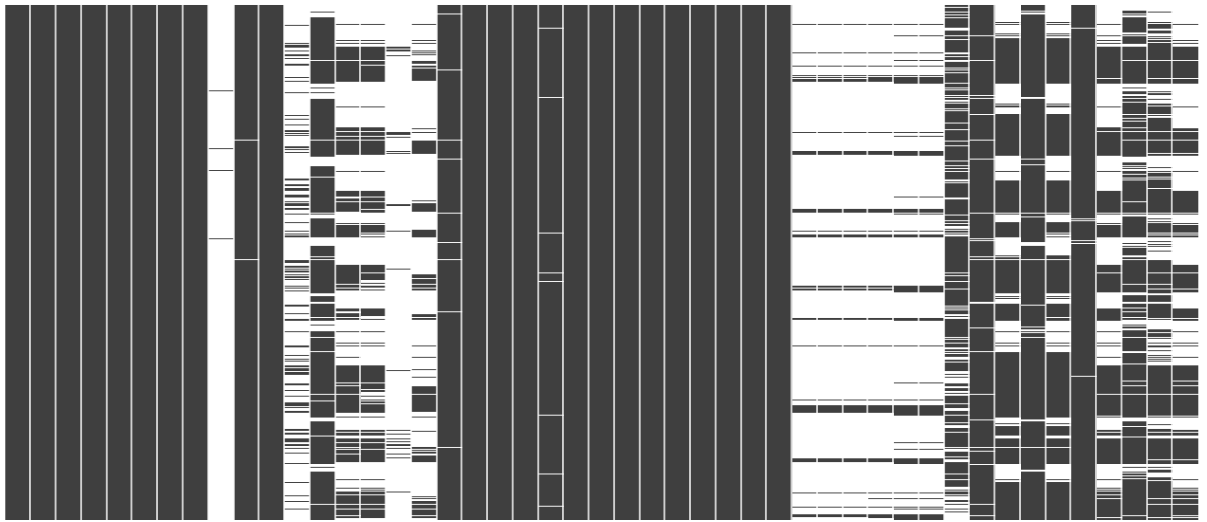}
        \caption{After mapping: 177,960 rows, 47 columns}
    \end{subfigure}
    \caption{Missing values (in white) in the raw and mapped databases}
    \label{fig:missings values raw and clean}
\end{figure}

 Figure~\ref{fig:missings values raw and clean} shows how much preprocessing had to be done: the white spaces represent the missing values. In the raw data, we had more than 5 million rows and 164 columns. After keeping only patients with melanoma of the skin, mapping and deleting variables that had nothing to do with skin cancer (about breast cancer for example), we have 177,960 rows and 47 columns. \par

\paragraph{Missing values and initial variable selection} After the mapping step, we still have many missing values. As we saw in section~\ref{constraints}, there are many constraints to be taken into account when choosing which variables to keep for our model.
\begin{itemize}
    \item Medical constraints:\\
    The probability of dying from cancer greatly depends on the metastatic state, so we cannot do without it. We remove individuals with missing metastatic state from the database. \\
    We also need to have the information on whether the individual is alive or not at the end of the observation period, so we remove individuals with missing information. \\
    Some variables are not medically relevant, for example the type of reporting source: if the information about an individual was collected by a hospital, a lab or another medical facility. So we delete this type of variable. \\

\begin{figure}[H]
    \centering
    \includegraphics[scale=0.6]{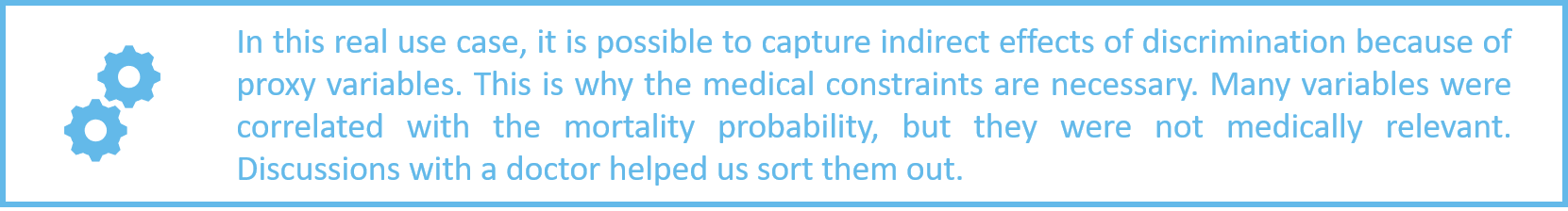}
\end{figure}

    As mortality rates evolve and change over the years, we need to set our time window so that the rates are still valid. Figure~\ref{fig:Yr_dx_count} shows the year of diagnosis distribution. We already have only recent years, mainly because we deleted individuals with unspecified metastatic state, and this variable was not collected in the past.

\begin{figure}[H]
    \centering
    \includegraphics[scale=0.5]{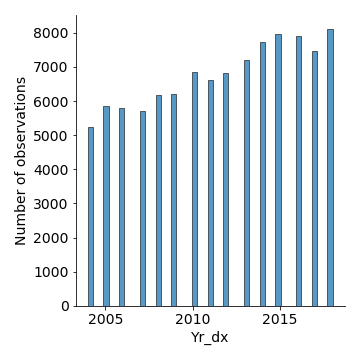}
    \caption{Number of new diagnoses per year}
    \label{fig:Yr_dx_count}
\end{figure}

    \item Underwriting constraints: \\
    We only keep individuals who are between 18 and 80 years old, as they are the ones that are covered by most markets. \\
    Normally, underwriting constraints impose, depending on the local regulation, the deletion of sensitive variables, like the origin of the individual. But this thesis focuses on the subject of discrimination, so we have to keep these variables to be able to measure bias.
    \item Modeling constraints: we do not keep variables to are strongly correlated with each other. This is the case for variables representing redundant information: we have 3 different variables for age at diagnosis, 4 different variables for origin, \dots We only keep one of each, the most complete one.
\end{itemize} \par
For the remaining missing values, we decided to delete the columns with almost only missing values, as they will not be useful for the model. Then, for numerical variables, the median of the series is assigned to any missing value. For categorical variables, we replace the missing values by a category `Missing'. In the end, we are left with 101,797 rows and 28 columns. \par

\subsection{Specificity of survival analysis}

\paragraph{} Most of the time, survival duration is observed partially. This can be due to the occurrence of the event of interest - in our case, death - outside the observation period, or to other events that result in the individual leaving the study (eg lapses, hospital transfers, recording systems failure). This censoring and truncation characterize survival data. If these effects were ignored, the probability of the event of interest would be underestimated. Another mistake would be to remove individuals for which the observation is incomplete from the study, because it would once again lead to biased estimations. \par
To deal with these issues, we either need to use specific models or modify the structure of the data to use standard models. We will use this second approach, which transforms the data by separating it into small time intervals. This allows to model the number of deaths by standard Machine Learning techniques using the exposure to risk as weights, taking into account the censoring. \par

\subsubsection{Exposure}

A first step is therefore to compute the exposure to risk of each individual. It can be computed differently depending on the hypothesis made on mortality. We compute the initial exposure which represents the time each individual was exposed to risk in the interval. It is called initial because it is based on the information at the beginning of the interval. \par
The individual initial exposure $e_{i,j}$ for individual i in time interval $[\tau_j,\tau_{j+1}]$ takes value:
\begin{itemize}
    \item 1 if the individual is alive during the entire time interval
    \item 1 if the individual dies during the time interval
    \item the fraction of the time interval he was observed if the individual is not observed during the entire time interval
\end{itemize}
Formally, denoting
\begin{itemize}
    \item[] $c_{i,j}$ the censoring time of individual i in time interval $[\tau_j,\tau_j+1]$
    \item[] $t_{i,j}$ the death time of individual i in time interval $[\tau_j,\tau_j+1]$
    \item[] $w_j$ the number of individuals withdrawn from the study in time interval $[\tau_j,\tau_j+1]$
    \item[] $l_j$ the number of individuals that are alive during the entire time interval $[\tau_j,\tau_j+1]$
    \item[] $d_j$ the number of individuals that died during the time interval $[\tau_j,\tau_j+1]$
\end{itemize}
we can write the individual initial exposure as:
\begin{equation*}
    \begin{split}
        e_{i,j} & =\left\{
        \begin{array}{ll}
            1 & \mbox{if } t_{i,j}>1 \mbox{ and } c_{i,j}>1  \\
            1 & \mbox{if } t_{i,j} <1 \\
            c_{i,j} & \mbox{if } c_{i,j}<1
        \end{array}
        \right. \\
         & =\underbrace{\mathbb{1}_{t_{i,j}>1}\times\mathbb{1}_{c_{i,j}>1}+\mathbb{1}_{t_{i,j}<1}}_{1-\mathbb{1}_{c_{i,j}<1}}+c_{i,j}\mathbb{1}_{c_{i,j}<1}
    \end{split}
\end{equation*}
The global initial exposure for all individuals in time interval $[\tau_j,\tau_j+1]$ is
\begin{equation*}
    \begin{split}
        E_j & =\sum_{i=1}^{l_j} e_{i,j} \\
         & =\sum_{i=1}^{l_j} (1-\mathbb{1}_{c_{i,j}<1}+c_{i,j}\mathbb{1}_{c_{i,j}<1}) \\
         & =l_j-w_j+\sum_{i=1}^{w_j}c_{i,j}
    \end{split}
\end{equation*}
The number of deaths is the sum of the number of observed deaths and expected deaths (from censored individuals). Writing down
\begin{itemize}
    \item[] $q_j=\mathbb{P}(T<\tau_j+1|T>\tau_j)$ the mortality rate in time interval $[\tau_j,\tau_j+1]$
    \item[] $_{c_{i,j}}q_j=\mathbb{P}(T<\tau_j+c_{i,j}|T>\tau_j)$ the mortality rate in time interval $[\tau_j,c_{i,j}]$
\end{itemize}
we can compute
\begin{equation*}
    \begin{split}
        d_j & =(l_j-w_j)q_j+\sum_{i=1}^{w_j} {}_{c_{i,j}}q_j \\
         & =l_j q_j-\sum_{i=1}^{w_j} {}_{1-c_{i,j}}q_{j+c_{i,j}}
    \end{split}
\end{equation*}
The Balducci hypothesis supposes that mortality rates decrease over the interval and are defined as:
\begin{equation*}
    \begin{split}
        {}_{1-c_{i,j}}q_{j+c_{i,j}} & =\mathbb{P}(T_i\leq\tau_j+1|T_i>\tau_j+c_{i,j}) \\
         & =(1-c_{i,j})\mathbb{P}(T_i\leq\tau_j+1|T_i>\tau_j) \\
         & =(1-c_{i,j})q_j
    \end{split}
\end{equation*}
So we can write
\begin{equation*}
    d_j=l_jq_j-q_j\sum_{i=1}^{w_j} (1-c_{i,j})
\end{equation*}
Solving the formula for $q_j$ gives us:
\begin{equation*}
        \hat{q}_j=\frac{d_j}{l_j-\sum_{i=1}^{w_j} (1-c_{i,j})}=\frac{d_j}{l_j-w_j+\sum_{i=1}^{w_j} c_{i,j}}
\end{equation*}
\begin{equation*}
    \boxed{\hat{q}_j=\frac{d_j}{E_j}}
\end{equation*}
We obtained the estimation for mortality rates on time interval $[\tau_j,\tau_j+1]$, corrected for censoring, using the Balducci hypothesis. Although it is generally not verified because mortality rates increase with time, the errors can be ignored as withdrawals are small compared to the population.

\begin{figure}[H]
    \centering
    \includegraphics[scale=0.6]{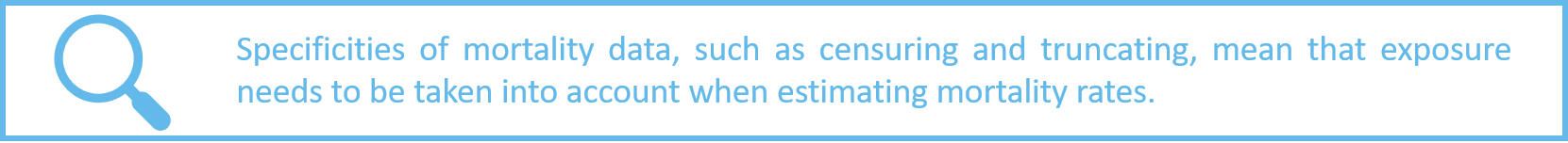}
\end{figure}

\subsubsection{Five-year mortality rates: a first look into the influence of each variable}

\paragraph{} In order to better understand the influence of certain factors on the mortality of melanoma of the skin patients and to compare our data with medical literature, we will compute the five-year mortality rates depending on these factors. In relation to the previous section, we need to compute the number of deaths and the global initial exposure in time interval $[0,5]$ (in years). For the number of deaths, we need to separate the cases of non-metastatic and metastatic patients, because metastasis implies the spread of the cancer to other sites, meaning that the cause of death can be a different cancer or disease that was caused by the initial cancer. For non-metastatic patients, the number of deaths will be computed with the number of deaths caused by skin melanoma and for metastatic patients, all causes of death will be taken into account. For individual i, we compute the death variable:
\begin{equation*}
    \begin{split}
        d_{i,5}^{M0}= & \left\{
        \begin{array}{ll}
            1 & \mbox{if Dead (due to skin melanoma) and survival < 5 years} \\
            0 & \mbox{else}
        \end{array}
        \right. \\
        d_{i,5}^{M1}= & \left\{
        \begin{array}{ll}
            1 & \mbox{if Dead (all causes) and survival < 5 years} \\
            0 & \mbox{else}
        \end{array}
        \right.
    \end{split}
\end{equation*}

Then, we compute the individual initial exposure as presented in the last section:
\begin{equation*}
    \begin{split}
        e_{i,5} & =\left\{
        \begin{array}{ll}
            1 & \mbox{if } t_{i,5}>1 \mbox{ and } c_{i,5}>1  \\
            1 & \mbox{if } t_{i,5} <1 \\
            c_{i,5} & \mbox{ if } c_{i,5}<1
        \end{array}
        \right. \\
         & =\left\{
        \begin{array}{ll}
            1 & \mbox{ if Vital\_status=0 or Survival\_years} \geq 5 \mbox{ Years} \\
            \frac{\mbox{Survival\_years}}{5} & \mbox{ else}
        \end{array}
        \right.
    \end{split}
\end{equation*}
Then, the estimation of the 5-year mortality rate is:
\begin{equation*}
    \hat{q}_5=\frac{\sum_{i=1}^{l_5} d_{i,5}}{\sum_{i=1}^{l_5} e_{i,5}}
\end{equation*}

We will compare the estimated 5-year mortality rates depending on the following factors: metastatic state (M), tumor size (T), spread to regional lymph nodes (N), stage, gender, origin and age. As we stated earlier, we need to separate the cases of non-metastatic and metastatic cancers, but without taking into account the metastatic state and looking at deaths caused by melanoma of the skin only, the 5-year mortality rate is of 7.59\%, ie the 5-year survival rate is of 92.41\% which is very close to the survival rate of 93.7\% given by the National Cancer Institute. \par

\noindent \begin{center}
    \fbox{
\begin{minipage}{0.9\textwidth}
   For the following, it is important to note that looking at mortality rates in different categories is tricky as groups may have different age distributions. Mortality rates by sex might capture correlation with age, if for example there are more observations of one gender than the other in some age groups.
\end{minipage}
}
\end{center}

The five-year mortality rate by metastatic state is given in table~\ref{tab:qx M}. As expected, the mortality rate for metastatic patients is considerably higher than for non-metastatic patients. \par

\begin{table}[H]
    \centering
    \begin{tabular}{c|c|c|}
        \cline{2-3}
         & Melanoma of the skin & All causes \\ \hline
        \multicolumn{1}{|c|}{M0} & 5.28\% & \cellcolor[HTML]{9B9B9B} \\ \hline
        \multicolumn{1}{|c|}{M1} & \cellcolor[HTML]{9B9B9B} & 83.00\% \\ \hline
    \end{tabular}
    \caption{Five-year mortality rate by presence of metastasis}
    \label{tab:qx M}
\end{table}

We expect mortality rates to go up with larger tumors. In the non-metastatic case, except for missing and T0 tumor sizes, the thicker the tumor, the higher the mortality rate. For a tumor size T1, the 5-year mortality rate is of 1.23\%, going up to 28.13\% for tumor size T4. The high mortality rate for T0 can be explained by the small number of observations: there are only 496 records of it, whereas for T1 for example there are 67,243 observations. A small number of observations results in a high variance in the estimation. Individuals with missing tumor sizes have a five-year mortality rate which is close to individuals with T2 tumors. In the metastatic case, we have the same problem for individuals with missing and T0 tumor sizes. Another unexpected result is that $\hat{q}_5^{M1,T1}>\hat{q}_5^{M1,T2}$0. This last result can come from the small sample sizes that result in a great estimation variance: there are respectively 202 and 154 individuals with tumors sizes T1 and T2. All mortality rates for metastatic cancers are in a smaller interval than for non-metastatic cancers. This might be interpreted as a lower predictive importance of the size of the primary tumor for metastatic cancers, and this might be due to the fact that these cancers have already spread to other sites. \par

\begin{table}[H]
    \centering
    \begin{tabular}{cc|c|c|}
            \cline{3-4}
         &  & Melanoma of the skin & All causes \\ \hline
        \multicolumn{1}{|c|}{} & Missing & 8.41\% & \cellcolor[HTML]{9B9B9B}{\color[HTML]{9B9B9B} } \\ \cline{2-4} 
        \multicolumn{1}{|c|}{} & T0 & 32.56\% & \cellcolor[HTML]{9B9B9B}{\color[HTML]{9B9B9B} } \\ \cline{2-4} 
        \multicolumn{1}{|c|}{} & T1 & 1.23\% & \cellcolor[HTML]{9B9B9B}{\color[HTML]{9B9B9B} } \\ \cline{2-4} 
        \multicolumn{1}{|c|}{} & T2 & 7.00\% & \cellcolor[HTML]{9B9B9B}{\color[HTML]{9B9B9B} } \\ \cline{2-4} 
        \multicolumn{1}{|c|}{} & T3 & 16.53\% & \cellcolor[HTML]{9B9B9B}{\color[HTML]{9B9B9B} } \\ \cline{2-4} 
        \multicolumn{1}{|c|}{\multirow{-6}{*}{M0}} & T4 & 28.13\% & \cellcolor[HTML]{9B9B9B} \\ \hline
        \multicolumn{1}{|c|}{} & Missing & \cellcolor[HTML]{9B9B9B} & 85.93\% \\ \cline{2-4} 
        \multicolumn{1}{|c|}{} & T0 & \cellcolor[HTML]{9B9B9B} & 82.09\% \\ \cline{2-4} 
        \multicolumn{1}{|c|}{} & T1 & \cellcolor[HTML]{9B9B9B} & 79.87\% \\ \cline{2-4} 
        \multicolumn{1}{|c|}{} & T2 & \cellcolor[HTML]{9B9B9B} & 75.54\% \\ \cline{2-4} 
        \multicolumn{1}{|c|}{} & T3 & \cellcolor[HTML]{9B9B9B} & 82.90\% \\ \cline{2-4} 
        \multicolumn{1}{|c|}{\multirow{-6}{*}{M1}} & T4 & \cellcolor[HTML]{9B9B9B} & 83.52\% \\ \hline
    \end{tabular}
    \caption{Five-year mortality rate by tumor size}
    \label{tab:qx T}
\end{table}

The spread to regional lymph nodes indicates how far a cancer has spread. The mortality rates are expected to go up with a more extensive spread. Only stages III and IV exhibit regional lymph node spread. For non-metastatic cases, we observe as expected an increase in mortality rates with greater spread. For the metastatic case, we have the same issues as in the previous paragraphs: mortality rates do not behave as expected. This can be due either to the little number of observations or the fact that with metastasis, the cancer has necessarily spread to regional lymph nodes, so the information collection process was faulty.

\begin{table}[H]
    \centering
    \begin{tabular}{cc|c|c|}
        \cline{3-4}
         &  & Melanoma of the skin & All causes \\ \hline
        \multicolumn{1}{|c|}{} & Missing & 6.02\% & \cellcolor[HTML]{9B9B9B}{\color[HTML]{9B9B9B} } \\ \cline{2-4} 
        \multicolumn{1}{|c|}{} & N0 & 3.30\% & \cellcolor[HTML]{9B9B9B}{\color[HTML]{9B9B9B} } \\ \cline{2-4} 
        \multicolumn{1}{|c|}{} & N1 & 23.59\% & \cellcolor[HTML]{9B9B9B}{\color[HTML]{9B9B9B} } \\ \cline{2-4} 
        \multicolumn{1}{|c|}{} & N2 & 33.11\% & \cellcolor[HTML]{9B9B9B}{\color[HTML]{9B9B9B} } \\ \cline{2-4} 
        \multicolumn{1}{|c|}{\multirow{-5}{*}{M0}} & N3 & 47.55\% & \cellcolor[HTML]{9B9B9B}{\color[HTML]{9B9B9B} } \\ \hline
        \multicolumn{1}{|c|}{} & Missing & \cellcolor[HTML]{9B9B9B} & 86.49\% \\ \cline{2-4} 
        \multicolumn{1}{|c|}{} & N0 & \cellcolor[HTML]{9B9B9B} & 78.99\% \\ \cline{2-4} 
        \multicolumn{1}{|c|}{} & N1 & \cellcolor[HTML]{9B9B9B} & 85.06\% \\ \cline{2-4} 
        \multicolumn{1}{|c|}{} & N2 & \cellcolor[HTML]{9B9B9B} & 74.21\% \\ \cline{2-4} 
        \multicolumn{1}{|c|}{\multirow{-5}{*}{M1}} & N3 & \cellcolor[HTML]{9B9B9B} & 84.38\% \\ \hline
    \end{tabular}
    \caption{Five-year mortality rate by spread to regional lymph nodes}
    \label{tab:qx N}
\end{table}

Looking at the stage of the cancer, it is common knowledge that more advanced stages imply higher mortality rates. This is what we observe here. Non-metastatic cancers can have grades ranging from I to III, and stage IV is defined by the presence of metastases. There are no missing stage information for individuals with stage IV cancers, because we imputed the missing values: remembering table~\ref{tab:stage TNM}, metastatic cancers are stage IV, which is how the information was completed for the variable.

\begin{table}[H]
    \centering
    \begin{tabular}{cc|c|c|}
        \cline{3-4}
         &  & Melanoma of the skin & All causes \\ \hline
        \multicolumn{1}{|c|}{} & Missing & 5.36\% & \cellcolor[HTML]{9B9B9B}{\color[HTML]{9B9B9B} } \\ \cline{2-4} 
        \multicolumn{1}{|c|}{} & I & 1.44\% & \cellcolor[HTML]{9B9B9B}{\color[HTML]{9B9B9B} } \\ \cline{2-4} 
        \multicolumn{1}{|c|}{} & II & 14.79\% & \cellcolor[HTML]{9B9B9B}{\color[HTML]{9B9B9B} } \\ \cline{2-4} 
        \multicolumn{1}{|c|}{\multirow{-4}{*}{M0}} & III & 29.87\% & \cellcolor[HTML]{9B9B9B}{\color[HTML]{9B9B9B} } \\ \hline
        \multicolumn{1}{|c|}{M1} & IV & \cellcolor[HTML]{9B9B9B} & 83.00\% \\ \hline
    \end{tabular}
    \caption{Five-year mortality rate by stage}
    \label{tab:qx stage}
\end{table}

Medically speaking, skin cancer behaves the exact same way for both genders. Mortality rates for the general population in the US are higher for men than women, which can be explained by numerous factors such as behavior differences. In insurance, gender is a sensitive variable and risk selection or pricing cannot discriminate by gender, so the same mortality rates have to be used for both men and women. In the case of melanoma of the skin, mortality rates are also higher for men. In the case of non-metastatic cancers, the five-year mortality rate for men is 2.56 points higher than for women and for metastatic cancers, it is 2.83 points higher. \par

\begin{table}[H]
    \centering
    \begin{tabular}{cc|c|c|}
        \cline{3-4}
         &  & Melanoma of the skin & All causes \\ \hline
        \multicolumn{1}{|c|}{} & Women & 3.82\% & \cellcolor[HTML]{9B9B9B} \\ \cline{2-4} 
        \multicolumn{1}{|c|}{\multirow{-2}{*}{M0}} & Men & 6.38\% & \cellcolor[HTML]{9B9B9B} \\ \hline
        \multicolumn{1}{|c|}{} & Women & \cellcolor[HTML]{9B9B9B} & 81.09\% \\ \cline{2-4} 
        \multicolumn{1}{|c|}{\multirow{-2}{*}{M1}} & Men & \cellcolor[HTML]{9B9B9B} & 83.92\% \\ \hline
    \end{tabular}
    \caption{Five-year mortality rate by gender}
    \label{tab:qx gender}
\end{table}

When looking at the origin variable provided by the SEER database, which classifies individuals into five categories (Hispanic, American Indian/Alaska Native, Asian or Pacific Islander, Black, White), we observe great disparity. The five-year mortality rates vary greatly between the categories, both for non-metastatic and metastatic cancers. As we saw previously, skin cancer is more common in individuals with fair complexions because of the lower melanin production. As a result, individuals with darker complexions are less likely to get this type of cancer and it often goes undetected for a longer time, which explains higher mortality rates. It could also be due to other factors such as socioeconomic status, which is still very related to origin in the US, where this data comes from. Another reason for these gaps is the variance of the estimation, as not all categories contain enough observations to have a robust estimation. It is interesting to notice that individuals with missing information about race/origin are predicted very low five-year mortality rates compared to all classes. This is caused by the low number of such observations, leading to a less robust estimator. \par

\begin{table}[H]
    \centering
    \begin{tabular}{cc|c|c|}
        \cline{3-4}
         &  & Melanoma of the skin & All causes \\ \hline
        \multicolumn{1}{|c|}{} & Hispanic & 7.34\% & \cellcolor[HTML]{9B9B9B} \\ \cline{2-4} 
        \multicolumn{1}{|c|}{} & Missing & 0.07\% & \cellcolor[HTML]{9B9B9B} \\ \cline{2-4} 
        \multicolumn{1}{|c|}{} & American Indian/AK Native & 9.86\% & \cellcolor[HTML]{9B9B9B} \\ \cline{2-4} 
        \multicolumn{1}{|c|}{} & Asian or Pacific Islander & 11.73\% & \cellcolor[HTML]{9B9B9B} \\ \cline{2-4} 
        \multicolumn{1}{|c|}{} & Black & 18.82\% & \cellcolor[HTML]{9B9B9B} \\ \cline{2-4} 
        \multicolumn{1}{|c|}{\multirow{-6}{*}{M0}} & White & 5.21\% & \cellcolor[HTML]{9B9B9B} \\ \hline
        \multicolumn{1}{|c|}{} & Hispanic & \cellcolor[HTML]{9B9B9B} & 86.53\% \\ \cline{2-4} 
        \multicolumn{1}{|c|}{} & Missing & \cellcolor[HTML]{9B9B9B} & 35.71\% \\ \cline{2-4} 
        \multicolumn{1}{|c|}{} & American Indian/AK Native & \cellcolor[HTML]{9B9B9B} & 88.45\% \\ \cline{2-4} 
        \multicolumn{1}{|c|}{} & Asian or Pacific Islander & \cellcolor[HTML]{9B9B9B} & 93.85\% \\ \cline{2-4} 
        \multicolumn{1}{|c|}{} & Black & \cellcolor[HTML]{9B9B9B} & 81.85\% \\ \cline{2-4} 
        \multicolumn{1}{|c|}{\multirow{-6}{*}{M1}} & White & \cellcolor[HTML]{9B9B9B} & 82.71\% \\ \hline
    \end{tabular}
    \caption{Five-year mortality rates by origin}
    \label{tab:qx origin}
\end{table}

Marital status can impact health behavior, leading to differences in mortality rates. In our data, we indeed have different estimated five-year mortality rates for all marital status classes. The individuals with missing information about their marital status have the lowest mortality rates, for both non-metastatic and metastatic cancers. For both metastatic states, divorced and widowed individuals have the highest five-year mortality rates, but these categories also correspond to an older population. \par

\begin{table}[H]
    \centering
    \begin{tabular}{cc|c|c|}
        \cline{3-4}
         &  & Melanoma of the skin & All causes \\ \hline
        \multicolumn{1}{|c|}{} & Divorced & 9.71\% & \cellcolor[HTML]{9B9B9B} \\ \cline{2-4} 
        \multicolumn{1}{|c|}{} & Married & 5.86\% & \cellcolor[HTML]{9B9B9B} \\ \cline{2-4} 
        \multicolumn{1}{|c|}{} & Missing & 1.75\% & \cellcolor[HTML]{9B9B9B} \\ \cline{2-4} 
        \multicolumn{1}{|c|}{} & Separated & 8.56\% & \cellcolor[HTML]{9B9B9B} \\ \cline{2-4} 
        \multicolumn{1}{|c|}{} & Single & 6.27\% & \cellcolor[HTML]{9B9B9B} \\ \cline{2-4} 
        \multicolumn{1}{|c|}{} & Unmarried & 7.71\% & \cellcolor[HTML]{9B9B9B} \\ \cline{2-4} 
        \multicolumn{1}{|c|}{\multirow{-7}{*}{M0}} & Widowed & 11.17\% & \cellcolor[HTML]{9B9B9B} \\ \hline
        \multicolumn{1}{|c|}{} & Divorced & \cellcolor[HTML]{9B9B9B} & 85.15\% \\ \cline{2-4} 
        \multicolumn{1}{|c|}{} & Married & \cellcolor[HTML]{9B9B9B} & 81.61\% \\ \cline{2-4} 
        \multicolumn{1}{|c|}{} & Missing & \cellcolor[HTML]{9B9B9B} & 77.78\% \\ \cline{2-4} 
        \multicolumn{1}{|c|}{} & Separated & \cellcolor[HTML]{9B9B9B} & 80.51\% \\ \cline{2-4} 
        \multicolumn{1}{|c|}{} & Single & \cellcolor[HTML]{9B9B9B} & 82.94\% \\ \cline{2-4} 
        \multicolumn{1}{|c|}{} & Unmarried & \cellcolor[HTML]{9B9B9B} & 74.38\% \\ \cline{2-4} 
        \multicolumn{1}{|c|}{\multirow{-7}{*}{M1}} & Widowed & \cellcolor[HTML]{9B9B9B} & 90.75\% \\ \hline
    \end{tabular}
    \caption{Five-year mortality rates by marital status}
    \label{tab:qx mar stat}
\end{table}

\begin{figure}[H]
    \centering
    \includegraphics[scale=0.6]{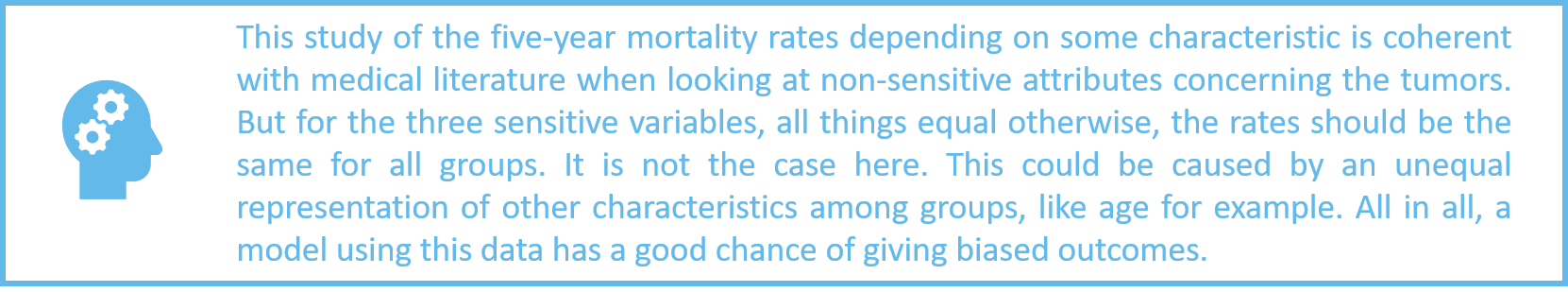}
\end{figure}

\subsubsection{A different data structure needed to use standard models: pseudo table}

\paragraph{} As mentioned previously, time discretization and data structure modification allow the use of standard Machine Learning models to predict the number of death in the desired time interval. This data structure modification is done by creating pseudo tables: for each individual, we create as many rows as the maximal duration. We consider times in years rather than in months, as it seems granular enough for our modeling purposes. In each row, we will have a death variable indicating if the individual died during that time interval and the initial exposure, which will then be used as a weight or an offset when applying the standard models to predict the number of deaths. Another advantage of the approach is that we can have time-varying information about an individual, such as the evolution of his marital status or of the size of his tumor. Unfortunately, in the SEER database, we do not have access to this kind of information, so we can question the relevance of using such variables. \par
The step-by-step process is as follows. We have for each individual i his age at diagnosis, year of diagnosis, survival time in months and a variable indicating if he died because of melanoma of the skin. The first step is to compute the survival time in years and the maximal duration, which is the ceiling of the survival time in years. Then, we create as many rows for each individual i as his/her maximal duration. We can then compute for each of the lines j the duration since diagnosis in years, and the age, year and death due to melanoma in that time interval. Finally, we can compute the individual exposure are we saw in the previous section. For each time interval j: $[\tau_j,\tau_j+1]$ which corresponds to a year, we compute the variables of interest. The duration since diagnosis (in years) is
\begin{equation*}
    \mbox{Duration}_{j}=j-1
\end{equation*}
the death variable (due to melanoma, in the time interval k) is
\begin{equation*}
    d_{i,j}=\left\{
    \begin{array}{ll}
        1 & \mbox{if Death\_melanoma}=1 \mbox{ and } j=\mbox{Max\_duration} \\
        0 & \mbox{ else}
    \end{array}
    \right.
\end{equation*}
the individual initial exposure is
\begin{equation*}
    e_{i,j}=\left\{
    \begin{array}{ll}
        1 & \mbox{if } Y_{i,j}=1 \mbox{ or } j<\mbox{Max\_duration} \\
        S_i-\mbox{Duration}_{j} & \mbox{ else}
    \end{array}
    \right.
\end{equation*}

We will create the pseudo table step-by-step, starting from the raw data from table~\ref{tab:raw survival}:
\begin{itemize}
    \item Individual i=1: \begin{itemize}
        \item For the survival time in years, we just convert the number of months into a number of years: $S_1=25/12=2.08$0.
        \item For the maximal duration in years, we take the ceiling of the survival time in years: $\mbox{Max\_duration}=\lceil S_1 \rceil = \lceil 2.08 \rceil = 3$0.
        \item So we create 3 lines $j=1,2,3$ for this individual: \begin{itemize}
            \item j=1: $\mbox{Duration}_{1}=j-1=0$ \\
            Age=Age\_dx=25 \\
            Year=Year\_dx=2002 \\
            Death\_melanoma $d_{1,1}=0$ because $\mbox{Death\_melanoma}\neq1$ \\
            Exposure $e_{1,1}=1$ because $j=1<3=\mbox{Max\_duration}$
            \item j=2: $\mbox{Duration}_{2}=j-1=1$ \\
            Age=Age\_dx+1=26 \\
            Year=Year\_dx+1=2003 \\
            Death\_melanoma $d_{1,2}=0$ because $\mbox{Death\_melanoma}\neq1$ \\
            Exposure $e_{1,2}=1$ because $j=2<3=\mbox{Max\_duration}$
            \item j=3: $\mbox{Duration}_{3}=j-1=2$ \\
            Age=Age\_dx+2=27 \\
            Year=Year\_dx+2=2004 \\
            Death\_melanoma $d_{1,3}=0$ because $\mbox{Death\_melanoma}\neq1$ \\
            Exposure $e_{1,3}=S_1-D_{1,3}=2.08-2=0.08$ because $Y_{1,3}\neq 1$ and $j=3=\mbox{Max\_duration}$
        \end{itemize}
    \end{itemize}
    \item Individual i=2: \begin{itemize}
        \item $S_2=4/12=0.33$0.
        \item $\mbox{Max\_duration}=\lceil S_2 \rceil = \lceil 0.33 \rceil = 1$0.
        \item So we create 1 line $j=1$ for this individual: \begin{itemize}
            \item j=1: $\mbox{Duration}_{1}=j-1=0$ \\
            Age=Age\_dx=37 \\
            Year=Year\_dx=2004 \\
            Death\_melanoma $d_{2,1}=1$ because $\mbox{Death\_melanoma}=1$ and $j=1=\mbox{Max\_duration}$\\
            Exposure $e_{2,1}=1$ because $Y_{2,1}=1$
        \end{itemize}
    \end{itemize}
    \item Individual i=3: \begin{itemize}
        \item $S_3=58/12=4.83$0.
        \item $\mbox{Max\_duration}=\lceil S_3 \rceil = \lceil 4.83 \rceil = 5$0.
        \item So we create 5 lines $j=1,2,3,4,5$ for this individual: \begin{itemize}
            \item j=1: $\mbox{Duration}_{1}=j-1=0$ \\
            Age=Age\_dx=56 \\
            Year=Year\_dx=2010 \\
            Death\_melanoma $d_{3,1}=0$ because $j=1\neq5=\mbox{Max\_duration}$ \\
            Exposure $e_{3,1}=1$ because $j=1<5=\mbox{Max\_duration}$
            \item j=2: $\mbox{Duration}_{2}=j-1=1$ \\
            Age=Age\_dx+1=57 \\
            Year=Year\_dx+1=2011 \\
            Death\_melanoma $d_{3,2}=0$ because $j=2\neq5=\mbox{Max\_duration}$ \\
            Exposure $e_{3,2}=1$ because $j=2<5=\mbox{Max\_duration}$
            \item j=3: $\mbox{Duration}_{3}=j-1=2$ \\
            Age=Age\_dx+2=58 \\
            Year=Year\_dx+2=2012 \\
            Death\_melanoma $d_{3,3}=0$ because $j=3\neq5=\mbox{Max\_duration}$ \\
            Exposure $e_{3,3}=1$ because $j=3<5=\mbox{Max\_duration}$
            \item j=4: $\mbox{Duration}_{4}=j-1=3$ \\
            Age=Age\_dx+3=59 \\
            Year=Year\_dx+3=2013 \\
            Death\_melanoma $d_{3,4}=0$ because $j=4\neq5=\mbox{Max\_duration}$ \\
            Exposure $e_{3,4}=1$ because $j=4<5=\mbox{Max\_duration}$
            \item j=5: $\mbox{Duration}_{5}=j-1=4$ \\
            Age=Age\_dx+4=60 \\
            Year=Year\_dx+4=2014 \\
            Death\_melanoma $d_{3,5}=1$ because $\mbox{Death\_melanoma}=1$ and $j=5=\mbox{Max\_duration}$ \\
            Exposure $e_{3,5}=1$ because $Y_{3,5}=1$
        \end{itemize}
    \end{itemize}
\end{itemize}
In the end, we get table~\ref{tab:pseudo survival}.

\begin{table}[H]
    \centering
    \begin{tabular}{|c|c|c|c|c|}
        \hline
        i & Age\_dx & Year\_dx & Survival (months) & Death\_melanoma \\ \hline
        1 & 25 & 2002 & 25 & 0 \\ \hline
        2 & 37 & 2004 & 4 & 1 \\ \hline
        3 & 56 & 2010 & 58 & 1 \\ \hline
    \end{tabular}
    \caption{Raw survival data}
    \label{tab:raw survival}
\end{table}

\begin{table}[H]
    \centering
    \makebox[\linewidth]{
    \begin{tabularx}{17,77cm}{|c|c|c|c|c|c|c|c|c|}
        \hline
        i & \begin{tabular}[c]{@{}c@{}}Survival \\ (years)\\$S_j$\end{tabular} & \begin{tabular}[c]{@{}c@{}}Max\_duration\\ (years)\end{tabular} & j & \begin{tabular}[c]{@{}c@{}}$\mbox{Duration}_{j}$\\ (years)\end{tabular} & Age & Year & \begin{tabular}[c]{@{}c@{}}Death\_melanoma\_j\\$d_{i,j}$\end{tabular} & \begin{tabular}[c]{@{}c@{}}Exposure\\ $e_{i,j}$\end{tabular} \\ \hline
        \multirow{3}{*}{1} & \multirow{3}{*}{25/12=2.08} & \multirow{3}{*}{3} & 1 & 0 & 25 & 2002 & 0 & 1 \\ \cline{4-9} 
         &  &  & 2 & 1 & 26 & 2003 & 0 & 1 \\ \cline{4-9} 
         &  &  & 3 & 2 & 27 & 2004 & 0 & 0.08 \\ \hline
        2 & 4/12=0.33 & 1 & 1 & 0 & 37 & 2004 & 1 & 1 \\ \hline
        \multirow{5}{*}{3} & \multirow{5}{*}{58/12=4.83} & \multirow{5}{*}{5} & 1 & 0 & 56 & 2010 & 0 & 1 \\ \cline{4-9} 
         &  &  & 2 & 1 & 57 & 2011 & 0 & 1 \\ \cline{4-9} 
         &  &  & 3 & 2 & 58 & 2012 & 0 & 1 \\ \cline{4-9} 
         &  &  & 4 & 3 & 59 & 2013 & 0 & 1 \\ \cline{4-9} 
         &  &  & 5 & 4 & 60 & 2014 & 1 & 1 \\ \hline
    \end{tabularx}
    }
    \caption{Pseudo table created from the raw survival data}
    \label{tab:pseudo survival}
\end{table}

\begin{figure}[H]
    \centering
    \includegraphics[scale=0.6]{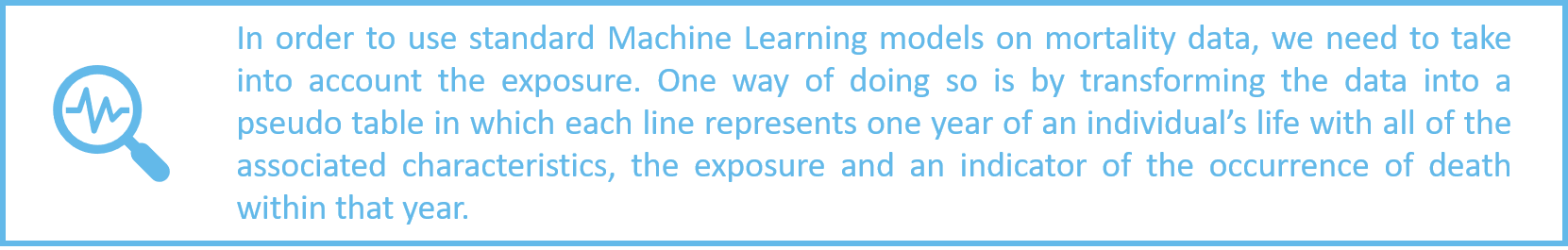}
\end{figure}

\subsection{Descriptive statistics}

\paragraph{} We will perform the descriptive statistics analysis on the pseudo table. As mentioned before, our study is restricted to non-metastatic cancers.

\subsubsection{Variable identification}

\paragraph{} The 28 variables available after mapping and applying underwriting and medical constraints are described in appendix~\ref{appendix:tab:variable description discretized}. They can be categorized into 3 groups: the explanatory variables, the variable of interest and the weight variable. Within the explanatory variables category, we have non-sensitive and sensitive variables. \par

\paragraph{Non-sensitive explanatory variables} They are the ones that are pure descriptors of the medical situation, for example the age, the type of cancer (\texttt{AYA\_site\_recode20}), the time since diagnosis (\texttt{DURATION}) or how far the cancer has spread (\texttt{Extent}). Some of them are continuous and others categorical. \par

\paragraph{Sensitive explanatory variables} They are the ones that can introduce discrimination. They directly concern the individual: sex, origin and marital status. \par
Table~\ref{tab:obs sex} gives the number of observations by sex. The group imbalance ratio is
\begin{equation*}
    \mbox{IR}_{Sex}=\frac{287,820}{249,118}=1.16
\end{equation*}
which is close to 1.

\begin{table}[H]
    \centering
    \begin{tabular}{|c|c|}
        \hline
        Sex & Observations \\ \hline
        Female & 249,118 \\ \hline
        Male & 287,820 \\ \hline
    \end{tabular}
    \caption{Number of observations by sex}
    \label{tab:obs sex}
\end{table}

Table~\ref{tab:obs origin} gives the number of observations by origin. The majority population is the `White' category, with almost 45 times more observations than the second biggest category, `Missing'. \par

\begin{table}[H]
    \centering
    \begin{tabular}{|c|c|}
        \hline
        Origin & Observations \\ \hline
        Hispanic & 11,034 \\ \hline
        Missing & 11,510 \\ \hline
        American Indian/AK Native & 1,157 \\ \hline
        Asian or Pacific Islander & 3,597 \\ \hline
        Black & 1,767 \\ \hline
        White & 507,873 \\ \hline
    \end{tabular}
    \caption{Number of observations by origin}
    \label{tab:obs origin}
\end{table}

Table~\ref{tab:obs mar stat} gives the number of observations by marital status. The majority class is `Married' and the second largest is `Missing'. \par

\begin{table}[H]
    \centering
    \begin{tabular}{|c|c|}
        \hline
        Marital status & Observations \\ \hline
        Divorced & 26,160 \\ \hline
        Married & 272,743 \\ \hline
        Missing & 159,649 \\ \hline
        Separated & 1,946 \\ \hline
        Single & 63,729 \\ \hline
        Unmarried & 674 \\ \hline
        Widowed & 12,037 \\ \hline
    \end{tabular}
    \caption{Number of observations by marital status}
    \label{tab:obs mar stat}
\end{table}

\paragraph{Variable of interest} It is the variable that we will predict with our model: the event of interest is the death due to skin melanoma of the individual in the year (\texttt{Death\_skin}). Table~\ref{tab:obs death skin} gives the number of observations for each value of \texttt{Death\_skin}. There is a great imbalance between classes: the group imbalance ratio is
\begin{equation*}
    \mbox{IR}_{Death\_skin}=\frac{533,167}{3,771}=141.39
\end{equation*}
meaning there are 141.39 more observations for \texttt{Death\_skin}=0 than for \texttt{Death\_skin}=1. \par

\begin{table}[H]
    \centering
    \begin{tabular}{|c|c|}
        \hline
        Death\_skin & Observations \\ \hline
        0 & 533,167 \\ \hline
        1 & 3,771 \\ \hline
    \end{tabular}
    \caption{Number of observations by death\_skin}
    \label{tab:obs death skin}
\end{table}

\paragraph{The weight variable} As mentioned previously, because of the characteristics of survival data, we need to take into account the exposure of individuals to predict mortality. The exposure variable will therefore be used as a weight in the model. \par

\subsubsection{Multivariate analysis}

\paragraph{} We will now study the relationships between variables. For that, we will focus on correlations. As we have already mentioned multiple times before, absence of correlation is not equivalent to independence. \par
As we have many variables and had to turn categorical features into binary data, we end up with a large correlation matrix. The keys points are:
\begin{itemize}
    \item The bright blue colors are negative correlations between different categories of the same variables, for example between \texttt{Sex\_Female} and \texttt{Sex\_Male}.
    \item Generally speaking, there are correlations between \texttt{Tumor}, \texttt{Positive\_Node}, \texttt{Stage}, \texttt{Extent}, \texttt{Ulceration}, \texttt{Mitotic rate}, and \texttt{Surg\_LN/oth/primsite}, which are linked with each other as they are the medical variables that allow the diagnosis.
    \item Time variables are naturally correlated with each other: age with age at diagnosis, year with year of diagnosis and duration.
    \item There are a few light correlations with the variable of interest, all less than 0.13 (in absolute value). 
    \item Concerning the missing values, the dummy variables for Missing values of \texttt{Tumor}, \texttt{Stage} and \texttt{Positive\_Nodes} are strongly positively correlated with each other. It seems coherent, as the stage is defined thanks to the TNM system. Missing categories of other variables are correlated with these variables and with each other: \texttt{Surg\_primsite}, \texttt{Surg\_LN}, \texttt{Surg\_oth}, \texttt{Extent}, \texttt{Ulceration}, and these variables also describe the gravity of the cancer. A Missing origin or marital status is lightly positively correlated with the previously mentioned variables. We can conclude that when there are Missing values for an individual, they are for all or most of the mentioned categories.
\end{itemize}
With this study of correlations, we can conclude that under the modeling constraints, not all variables can be kept.

\begin{figure}[H]
    \centering
    \includegraphics[scale=0.3]{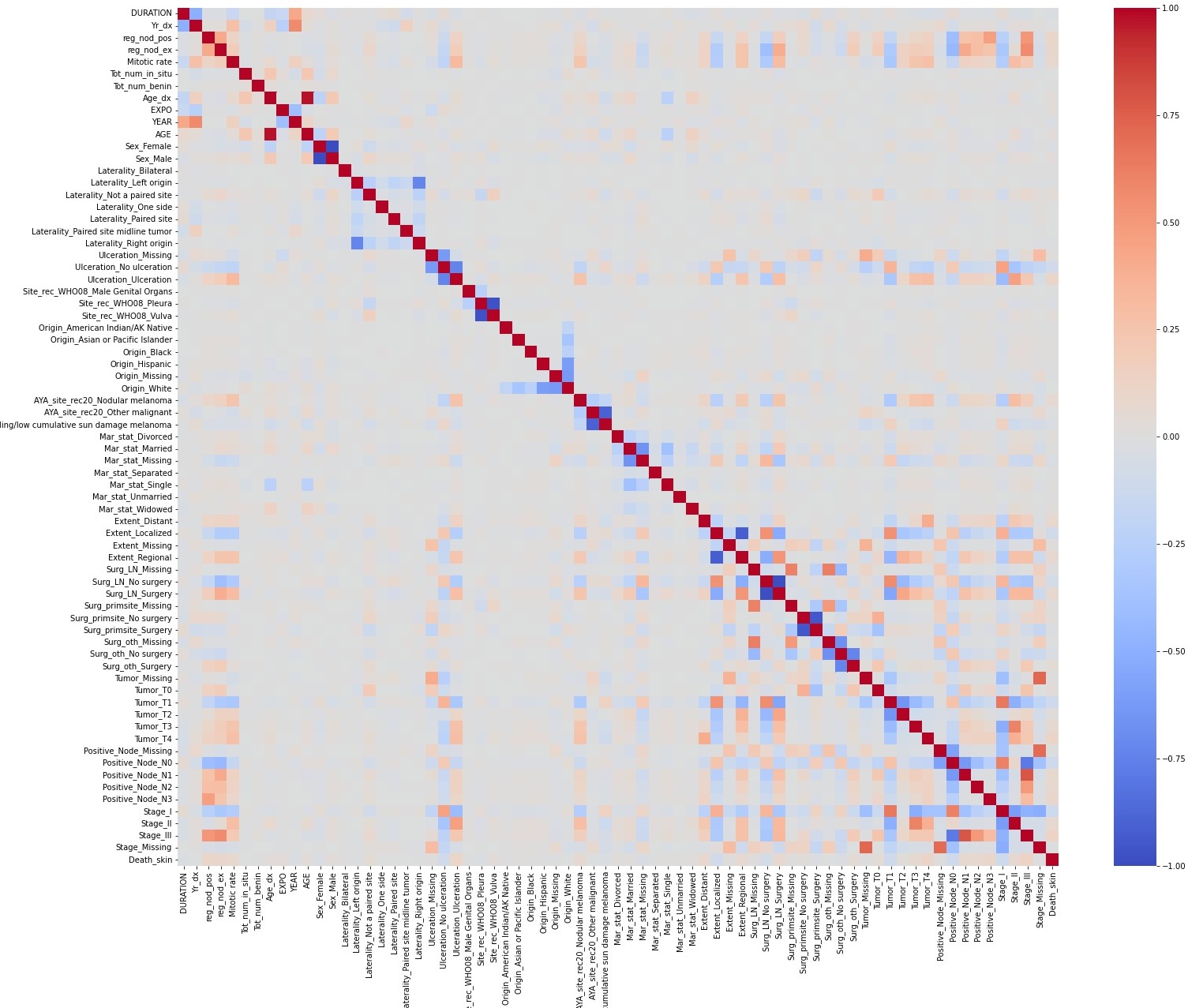}
    \caption{Heatmap of correlations}
    \label{fig:heatmap corr seer}
\end{figure}

\begin{figure}[H]
    \centering
    \includegraphics[scale=0.6]{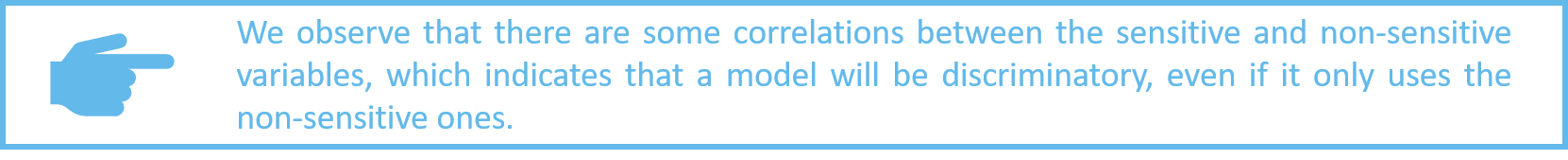}
\end{figure}

\newpage

{\Large\textbf{Discrimination mitigation applied to real mortality data}}

\section{Discrimination mitigation applied to real mortality data}
\subsection{Adapting the logistic regression to survival data}

\paragraph{} If we considered a simple logistic regression to predict the event of death $\delta_{i,j}\sim \mathcal{B}(q_j)$, we would not take into account censoring. Instead, we assign weights to the sample with the individual initial exposure. This way, we estimate the mortality rate $q_j=\mathbb{P}(T\leq\tau_j+1|T>\tau_j)$ with $\hat{q}_j=\frac{\sum_i \delta_{i,j}}{\sum_ie_{i,j}}$0.
\begin{proof}
The weighted likelihood of the model is
\begin{equation*}
    L=\prod_{i=1}^{l_j} q_j^{\delta_{i,j}e_{i,j}}(1-q_j)^{(1-\delta_{i,j})e_{i,j}}
\end{equation*}
The log likelihood to maximize is
\begin{equation*}
    log(L)=\sum_{i=1}^{l_j} \Bigl( \delta_{i,j}e_{i,j}log(q_j)+(1-\delta_{i,j})e_{i,j}log(1-q_j) \Bigr)
\end{equation*}
Deriving with respect to $q_j$, we obtain
\begin{equation*}
    \frac{dlog(L)}{dq_j} = \sum_{i=1}^{l_j} \Bigl( \frac{\delta_{i,j}e_{i,j}}{q_j} - \frac{e_{i,j}-\delta_{i,j}e_{i,j}}{1-q_j} \Bigr)
\end{equation*}
Equalizing with 0, we get
\begin{equation*}
    \begin{split}
         & (1-\hat{q}_j)\sum_{i=1}^{l_j} \delta_{i,j}e_{i,j}=\hat{q}_j\sum_{i=1}^{l_j} \bigl( e_{i,j}-\delta_{i,j}e_{i,j} \bigr) \\
        \mbox{ie } & \hat{q}_j=\frac{\sum_i \delta_{i,j}e_{i,j}}{\sum_i e_{i,j}}=\frac{\sum_i \delta_{i,j}}{\sum_ie_{i,j}} \mbox{ as } \delta=1 \Rightarrow e=1
    \end{split}
\end{equation*}
\end{proof}

\begin{figure}[H]
    \centering
    \includegraphics[scale=0.6]{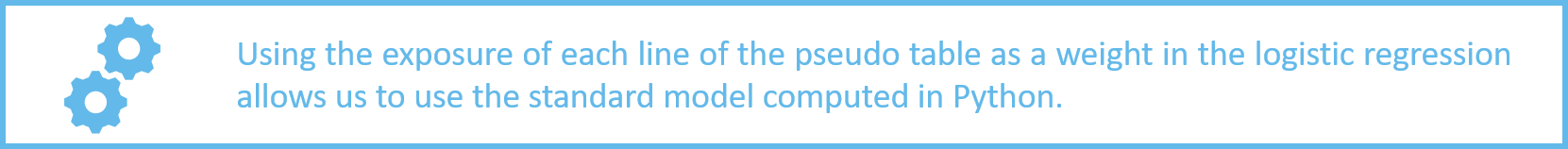}
\end{figure}

\subsection{Variable selection}

\paragraph{} In order to select the variables that are relevant to the model performance, we begin with a model with all variables. Table~\ref{tab:coeff complete} gives the coefficients of the regression, with their standard errors and p-values. We will only keep the variables with a p-value below or equal to 0.005, as it is good practice in modeling. The p-values that are above 0.005 are highlighted in orange. Variables that are not statistically significant are: \begin{itemize}
    \item \texttt{AGE}
    \item \texttt{reg\_nod\_ex}
    \item \texttt{Tot\_num\_benin}
    \item \texttt{AYA\_site\_rec20}
    \item \texttt{Surg\_LN}
    \item \texttt{Surg\_oth}
\end{itemize}
This does not necessarily mean that these variables are not medically relevant. \par

\begin{longtable}[c]{|l|c|c|c|}
    \hline
    Variable & Coefficient & Standard error & P-value \\ \hline
    \endfirsthead
    \endhead
    const & 174.9219 & 13.260 & 0.000 \\ \hline
    DURATION & -0.0191 & 0.003 & 0.000 \\ \hline
    Yr\_dx & -0.0358 & 0.004 & 0.000 \\ \hline
    Age\_dx & 0.0207 & 0.002 & 0.000 \\ \hline
    YEAR & -0.0549 & 0.003 & 0.000 \\ \hline
    AGE & 0.0016 & 0.002 & \cellcolor[HTML]{FFCC67}0.396 \\ \hline
    Sex\_Female & -0.3927 & 0.043 & 0.000 \\ \hline
    reg\_nod\_pos & 0.0289 & 0.007 & 0.000 \\ \hline
    reg\_nod\_ex & 0.0003 & 0.002 & \cellcolor[HTML]{FFCC67}0.849 \\ \hline
    Mitotic rate & 0.0528 & 0.007 & 0.000 \\ \hline
    Tot\_num\_in\_situ & -0.0742 & 0.023 & 0.002 \\ \hline
    Tot\_num\_benin & 0.0777 & 0.170 & \cellcolor[HTML]{FFCC67}0.648 \\ \hline
    Laterality\_Bilateral & 0.6875 & 0.331 & \cellcolor[HTML]{FFCC67}0.038 \\ \hline
    Laterality\_Not a paired site & 0.4738 & 0.061 & 0.000 \\ \hline
    Laterality\_One side & 0.0624 & 0.200 & \cellcolor[HTML]{FFCC67}0.756 \\ \hline
    Laterality\_Paired site & 0.2974 & 0.093 & 0.001 \\ \hline
    Laterality\_Paired site midline tumor & 0.1334 & 0.128 & \cellcolor[HTML]{FFCC67}0.296 \\ \hline
    Laterality\_Right origin & -0.0520 & 0.044 & \cellcolor[HTML]{FFCC67}0.236 \\ \hline
    Ulceration\_Missing & 0.1514 & 0.099 & \cellcolor[HTML]{FFCC67}0.128 \\ \hline
    Ulceration\_Ulceration & 0.5980 & 0.048 & 0.000 \\ \hline
    Site\_rec\_WHO08\_Male Genital Organs & 0.0514 & 0.565 & \cellcolor[HTML]{FFCC67}0.928 \\ \hline
    Site\_rec\_WHO08\_Vulva & 0.5930 & 0.183 & 0.001 \\ \hline
    Origin\_American Indian/AK Native & 0.3195 & 0.293 & \cellcolor[HTML]{FFCC67}0.276 \\ \hline
    Origin\_Asian or Pacific Islander & 0.1691 & 0.156 & \cellcolor[HTML]{FFCC67}0.280 \\ \hline
    Origin\_Black & 0.4672 & 0.167 & 0.005 \\ \hline
    Origin\_Hispanic & 0.2042 & 0.108 & \cellcolor[HTML]{FFCC67}0.060 \\ \hline
    Origin\_Missing & -3.0099 & 1.001 & 0.003 \\ \hline
    AYA\_site\_rec20\_Nodular melanoma & 0.1249 & 0.052 & \cellcolor[HTML]{FFCC67}0.016 \\ \hline
    \begin{tabular}[c]{@{}l@{}}AYA\_site\_rec20\_Superficial spreading\\ /low cumulative sun damage melanoma\end{tabular} & -0.0319 & 0.048 & \cellcolor[HTML]{FFCC67}0.510 \\ \hline
    Mar\_stat\_Divorced & 0.3849 & 0.064 & 0.000 \\ \hline
    Mar\_stat\_Missing & -0.4411 & 0.072 & 0.000 \\ \hline
    Mar\_stat\_Separated & 0.3068 & 0.225 & \cellcolor[HTML]{FFCC67}0.173 \\ \hline
    Mar\_stat\_Single & 0.2578 & 0.055 & 0.000 \\ \hline
    Mar\_stat\_Unmarried & 0.3127 & 0.426 & \cellcolor[HTML]{FFCC67}0.463 \\ \hline
    Mar\_stat\_Widowed & 0.3041 & 0.087 & 0.000 \\ \hline
    Extent\_Distant & 0.7494 & 0.078 & 0.000 \\ \hline
    Extent\_Missing & 0.6607 & 0.171 & 0.000 \\ \hline
    Extent\_Regional & 0.4792 & 0.057 & 0.000 \\ \hline
    Surg\_LN\_Missing & 0.0226 & 0.347 & \cellcolor[HTML]{FFCC67}0.948 \\ \hline
    Surg\_LN\_Surgery & -0.1323 & 0.057 & \cellcolor[HTML]{FFCC67}0.020 \\ \hline
    Surg\_primsite\_Missing & 0.7475 & 0.335 & \cellcolor[HTML]{FFCC67}0.025 \\ \hline
    Surg\_primsite\_No surgery & 0.5662 & 0.136 & 0.000 \\ \hline
    Surg\_oth\_Missing & -0.6858 & 0.474 & \cellcolor[HTML]{FFCC67}0.148 \\ \hline
    Surg\_oth\_Surgery & 0.2390 & 0.095 & \cellcolor[HTML]{FFCC67}0.012 \\ \hline
    Tumor\_Missing & 1.0042 & 0.173 & 0.000 \\ \hline
    Tumor\_T0 & 0.4037 & 0.203 & \cellcolor[HTML]{FFCC67}0.047 \\ \hline
    Tumor\_T2 & 0.9187 & 0.076 & 0.000 \\ \hline
    Tumor\_T3 & 1.0184 & 0.094 & 0.000 \\ \hline
    Tumor\_T4 & 1.2887 & 0.097 & 0.000 \\ \hline
    Positive\_Node\_Missing & 0.9338 & 0.214 & 0.000 \\ \hline
    Positive\_Node\_N1 & -0.1761 & 0.588 & \cellcolor[HTML]{FFCC67}0.764 \\ \hline
    Positive\_Node\_N2 & 0.0468 & 0.590 & \cellcolor[HTML]{FFCC67}0.937 \\ \hline
    Positive\_Node\_N3 & 0.4810 & 0.594 & \cellcolor[HTML]{FFCC67}0.418 \\ \hline
    Stage\_II & 0.5843 & 0.083 & 0.000 \\ \hline
    Stage\_III & 1.5105 & 0.590 & \cellcolor[HTML]{FFCC67}0.011 \\ \hline
    Stage\_Missing & -0.3659 & 0.207 & \cellcolor[HTML]{FFCC67}0.077 \\ \hline
    \captionsetup{justification=centering}
    \caption{Coefficients of the complete model, \\p-values in orange when above the threshold of 0.005}
    \label{tab:coeff complete}\\
\end{longtable}

For our final model, the selected variables need to be statistically relevant, consistent with medical literature and available at the underwriting stage. We have discussed with the oncologist of the team, drawing the following conclusions: \begin{itemize}
    \item For time variables, the most important ones are the current age and duration. As we had strong correlations between age and age at diagnosis, we will only keep age. The year does not influence mortality, as we only have data from recent years.
    \item Laterality does not influence mortality, but it statistically relevant so we will keep the variable in the model.
    \item Fairer complexions are at a higher risk of developing the cancer, but once the cancer has developed, it does not influence mortality, so all things equal otherwise, we should obtain the same mortality rates for all Origins. Unfortunately, and especially in the US, Origin is very often a proxy for social economic status, which influences access to healthcare for example, which in turn has consequences on mortality risks.
    \item Stage is a combination of the TNM staging variables, so there are strong correlations between them, and there is no use in keeping them all. So we will not keep the \texttt{Stage} variable.
\end{itemize}
To conclude, we will keep the \texttt{AGE} and \texttt{Laterality} variables. We note that we are left with the three sensitive variables, \texttt{Sex}, \texttt{Origin} and \texttt{Mar\_stat}. \par

\begin{figure}[H]
    \centering
    \includegraphics[scale=0.6]{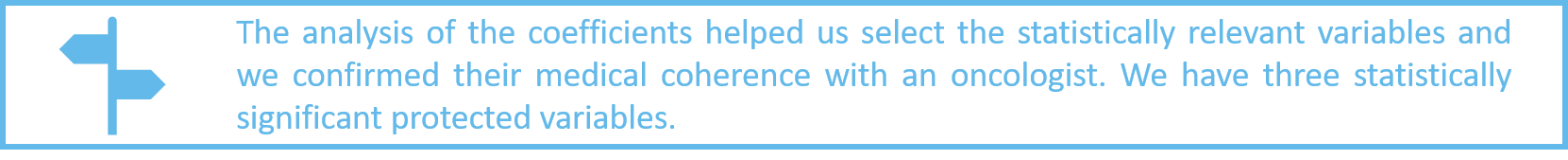}
\end{figure}

\subsection{Regression model with no pre-processing step}

\paragraph{} As we have done in the simulated case, we begin by applying the regression model without any particular modifications. This will allow a comparison with other methods. As we are left with many variables, the coefficients of the regression are given in appendix~\ref{appendix:tab:coeff selected}. \par
Figure~\ref{fig:seer coeff sel} gives a visual understanding of the coefficients. We have an intercept and only kept the least represented classes for the categorical variables, so the coefficients for the remaining categorical variables represent the relative risk compared to the most represented class. For example, for the variable \texttt{Sex}, as we had more observations for Male, we only kept the variable \texttt{Sex\_Female} in our model. The coefficient attributed to this variable by the logistic regression indicates how much more chances of dying a Female has, compared to a Male. For numerical variables, the coefficients indicate how much the chances of dying change for an increase of one in the variable value. We have also included the standard errors, represented by the black lines.
\begin{itemize}
    \item \texttt{DURATION}: the higher the duration, the lower the probability of dying.
    \item \texttt{AGE}: the older the individual, the higher his probability of dying.
    \item \texttt{Sex}: compared to being a Male, being a Female decreases the probability of dying.
    \item \texttt{reg\_nod\_pos}: the higher the number of positive regional nodes, the higher the probability of dying.
    \item \texttt{Mitotic rate}: the higher the density of mitoses, the higher the probability of dying.
    \item \texttt{Laterality}: compared to the Left origin category, a Right origin lowers the probability of death and other categories increases it.
    \item \texttt{Ulceration}: compared to No ulceration, other categories increases the death probability, Ulceration more than Missing.
    \item \texttt{Site\_rec\_WHO08}: compared to tumors originating on the Pleura, tumors originated on the vulva and on male genital organs increase the probability of dying.
    \item \texttt{Origin}: compared to White, Missing largely decreases the probability of dying. Other categories increase it, in the increasing order we have Asian or Pacific Islander, Hispanic, American Indian/AK Native and Black. 
    \item \texttt{Marital status}: compared to Married, Missing decreases the probability of death. Other categories increase it, in the increasing order we have Unmarried, Single, Separated, Widowed and Divorced. We must take into account the standard errors, which are larger than the coefficients for Unmarried and Separated.
    \item \texttt{Extent}: compared to Localized, all categories increase the probability of dying, in the increasing order we have Missing, Regional and Distant.
    \item \texttt{Surg\_primsite}: compared to Surgery, other categories increase the probability of dying, No surgery more than Missing.
    \item \texttt{Tumor}: compared to T1, larger tumors increase the probability of dying in size order. Missing increases the same way as T2. But we have an anomaly for T0 as it seems it increases the probability of dying compared to T1. This might be due to the small sample size, as we saw when we were looking at the five-year mortality rates.
    \item \texttt{Positive\_Node}: compared to N0, larger numbers of positives nodes increase the probability of dying. Missing increases it, but not at much as the larger sizes.
\end{itemize}
The coefficients are coherent with medical findings: a higher number of positive regional nodes, the presence of ulceration, a further extent, and a larger tumor all point towards a severe cancer, which implies lower survival odds.

\begin{figure}[H]
    \centering
    \includegraphics[scale=0.7]{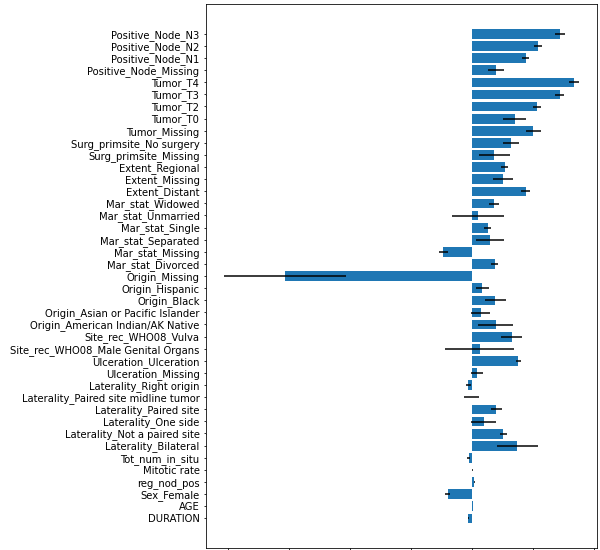}
    \caption{Coefficients and their standard errors for the model with selected variables}
    \label{fig:seer coeff sel}
\end{figure}

\begin{figure}[H]
    \centering
    \includegraphics[scale=0.6]{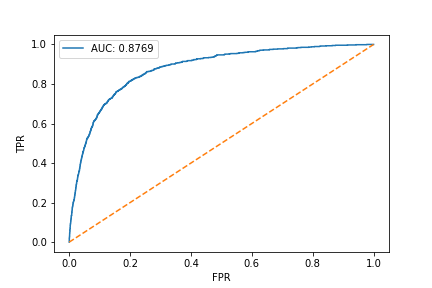}
    \caption{ROC curve for the model with selected variables}
    \label{fig:ROC selected var}
\end{figure}

\paragraph{Performance evaluation} Figure~\ref{fig:ROC selected var} gives the ROC curve. The performance of the model is almost the same as with all variables, with an AUC of 0.8769. \par

In order to compute fairness metrics, we need to classify individuals into two categories $\hat{Y}=0$ or $1$, $\hat{Y}$ being the event of death due to skin melanoma, 1 if the event occurred ie the individual died of this cause and 0 otherwise. As the probabilities of death are all very close to 0, we cannot use the standard threshold of 0.5 to classify individuals. We will select the threshold by looking at the true percentage of dead individuals. In the entire dataset, we have about 0.7\% of dead individuals, as we saw in the section on descriptive statistics. We will therefore look for a threshold $\tau$ such that
\begin{equation*}
    \hat{Y}=1 \mbox{ if } \mathbb{P}(\hat{Y}=1)\geq \tau \mbox{ and } 0 \mbox{ otherwise, giving 0.7\% of individuals with }\hat{Y}=1
\end{equation*}
We find $\tau=0.1088$0. This value gives the confusion matrix as shown in figure~\ref{tab:seer confusion matrix sel variables}. We have, as planned, about 0.7\% individuals classified with $\hat{Y}=1$0. The class $\hat{Y}=0$ is the positive class here. \par

\begin{table}[H]
    \begin{center}
        \begin{tabular}{ccccc}
            \cline{2-3}
            \multicolumn{1}{c|}{}  & \multicolumn{1}{c|}{$\hat{Y}$=0} & \multicolumn{1}{c|}{$\hat{Y}$=1} &  &  \\ \cline{1-3}
            \multicolumn{1}{|c|}{$Y=0$} & \multicolumn{1}{c|}{106,236}  & \multicolumn{1}{c|}{699}  &  &  \\ \cline{1-3}
            \multicolumn{1}{|c|}{$Y=1$} & \multicolumn{1}{c|}{662}  & \multicolumn{1}{c|}{95}  &  &  \\ \cline{1-3}
        \end{tabular}
    \end{center}
    \caption{Confusion matrix (model with selected variables)}
    \label{tab:seer confusion matrix sel variables}
\end{table}

\paragraph{Fairness evaluation} We computed the acceptance rate and the true and false positive rates, and looked at them globally, by sex, by origin and by marital status. To conduct an extensive analysis, we should also look at subgroups, crossing information about the three sensitive attributes, like we did in the simulated case, but it becomes intricate as we have many different categories, and we risk having too few observations in one of the subcategories. \par
\begin{itemize}
    \item Globally, we have values close to 100\% for the acceptance and true positive rates, due to the numerous classifications into the positive class compared to the negative class. The false positive rate is quite high, as we have 6 times more false positives than true negatives.
    
\begin{table}[H]
    \centering
    \begin{tabular}{c|c}
        (\%) & Global \\ \hline
        AR & 99.26 \\
        TPR & 99.35 \\
        FPR & 87.45
    \end{tabular}
    \caption{Global fairness metrics (model with selected variables)}
    \label{tab:seer metrics sel variables global}
\end{table}

    \item By sex \begin{itemize}
        \item The acceptance rates are very close, only 0.76 points apart, but higher for Female.
        \item The true positive rates are very close for both genders, only 0.69 points apart, but higher for Female.
        \item The false positive rates are 3.87 points part, and higher for Female.
    \end{itemize}
    All in all, looking at the three definitions of fairness, the Male sex is always disadvantaged by this model, although the gaps between metrics are not large.

\begin{table}[H]
    \centering
    \begin{tabular}{c|cc|c}
        (\%) & Female & Male & Difference \\ \hline
        AR & 99.67 & 98.91 & 0.76 \\
        TPR & 99.71 & 99.03 & 0.69\\
        FPR & 90.18 & 86.30 & 3.87
    \end{tabular}
    \caption{Fairness metrics by sex (model with selected variables)}
    \label{tab:seer metrics sel variables sex}
\end{table}

    \item By origin \begin{itemize}
        \item The acceptance rates are between 92.53\% and 100\%. The lowest is for Black and the highest is for Missing.
        \item The true positive rates are between 93.21\% and 100\%. The lowest is for Black and the highest for Missing.
        \item The false positive rates are very distant. The lowest FPR is for Missing at 0\%, then 89.20\% for White. The highest is for American Indian/Alaska Native at 100.00\%.
    \end{itemize}
    It is interesting to note that the Missing category only has true positive classifications, which explains the values for the fairness metrics. Going back on the regression weights of figure~\ref{fig:seer coeff sel}, the \texttt{Origin\_Missing} variable has a colossal negative coefficient compared to other dummy variables, and especially compared to other variables for \texttt{Origin}. Whenever the origin of an individual was not collected, his probability of dying within the year is null. We can wonder about the data collection process: perhaps individuals with zero to no chance of dying were not as `interesting' for the database. \\
    Black is the most disadvantaged group under the statistical parity and equal opportunity definitions. For the equalized odds definition, Black has the lowest TPR but White has the lowest FPR, so we cannot conclude on the most disadvantaged group.
\begin{table}[H]
    \centering
    \begin{tabular}{c|ccccccc}
        (\%)& White & Hispanic & Black & \begin{tabular}[c]{@{}c@{}}Asian or\\ Pacific Islander\end{tabular} & \begin{tabular}[c]{@{}c@{}}American Indian\\ /AK Native\end{tabular} & Missing & \multicolumn{1}{|c}{Var} \\ \hline
        AR & 99.33 & 97.70 & 92.53 & 96.63 & 98.72 & 100.00 & \multicolumn{1}{|c}{$6.10\mathrm{e}{-4}$} \\
        TPR & 99.40 & 97.94 & 93.21& 97.53 & 100.00 & 100.00 & \multicolumn{1}{|c}{$5.52\mathrm{e}{-4}$} \\
        FPR & 89.20 & 97.94 & 93.21 & 97.53 & 100.00 & 0.00 & \multicolumn{1}{|c}{$8.35\mathrm{e}{-2}$}
    \end{tabular}
    \caption{Fairness metrics by origin (model with selected variables)}
    \label{tab:seer metrics sel variables race}
\end{table}

    \item By marital status \begin{itemize}
        \item The acceptance rate is lowest for Separated and highest for Missing, then Married.
        \item The true positive rates are between 97.54\% and 99.99\%, lowest for Widowed and highest for Missing, then Married.
        \item The false positive rate is lowest for Widowed, at 78.79\% and highest for Separated and Unmarried, at 100\%.
    \end{itemize}
    The Separated class is the most disadvantaged under the statistical parity definition, Widowed under the equal opportunity and equalized odds definitions.
\end{itemize}

\begin{table}[H]
    \centering
    \begin{tabular}{c|cccccccc}
         (\%) & Married & Missing & Single & Widowed & Divorced & Separated & Unmarried & \multicolumn{1}{|c}{Var} \\ \hline
        AR & 99.23 & 99.99 & 98.82 & 97.29 & 97.47 & 95.75 & 98.44 & \multicolumn{1}{|c}{$1.72\mathrm{e}{-4}$} \\
        TPR & 99.33 & 99.99 & 98.95 & 97.54 & 97.69 & 95.71 & 98.43 & \multicolumn{1}{|c}{$1.71\mathrm{e}{-4}$} \\
        FPR & 88.16 & 97.10 & 84.30 & 78.79 & 82.19 & 100.00 & 100.00 & \multicolumn{1}{|c}{$6.76\mathrm{e}{-3}$}
    \end{tabular}
    \caption{Fairness metrics by marital status (model with selected variables)}
    \label{tab:seer metrics sel variables mar stat}
\end{table}

To conclude, the most disadvantaged group is not the same depending on the definition of fairness that is used.

\begin{figure}[H]
    \centering
    \includegraphics[scale=0.6]{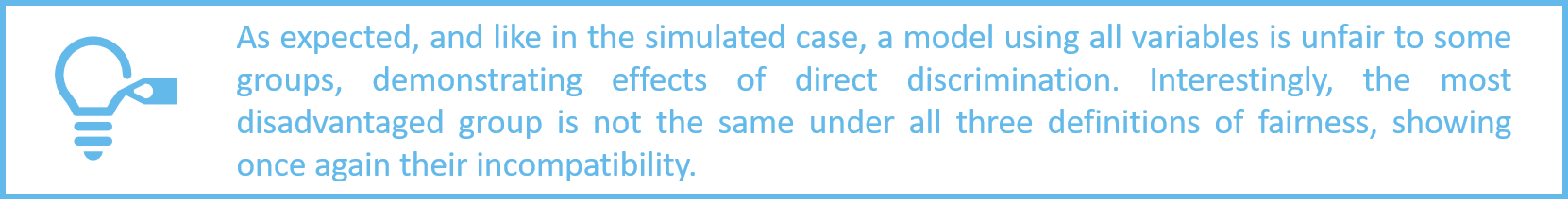}
\end{figure}

\subsection{Removing protected variables to avoid direct discrimination}

\paragraph{} We will now compare the model with the protected variables to the model without protected variables. As we saw in the simulated case, removing the protected variables is not the perfect solution as it does not prevent discrimination. Simply ignoring the sensitive attributes gives larger weights to variables that are correlated with them. \par
Here, we will apply a regression model to all variables except the sensitive ones. The coefficients are given in appendix~\ref{appendix:tab:coeff without A}. \par
Figure~\ref{fig:seer coeff without A} shows the importance of each variable. The main difference that we can observe, compared to the model with all variables, is that the variable \texttt{Site\_rec\_WHO08\_Male Genital Organs} now has a 
negative coefficient, with a consequent standard error. Other than that, the weights stayed in the same order, with some amplitude variations.

\begin{figure}[H]
    \centering
    \captionsetup{justification=centering}
    \includegraphics[scale=0.6]{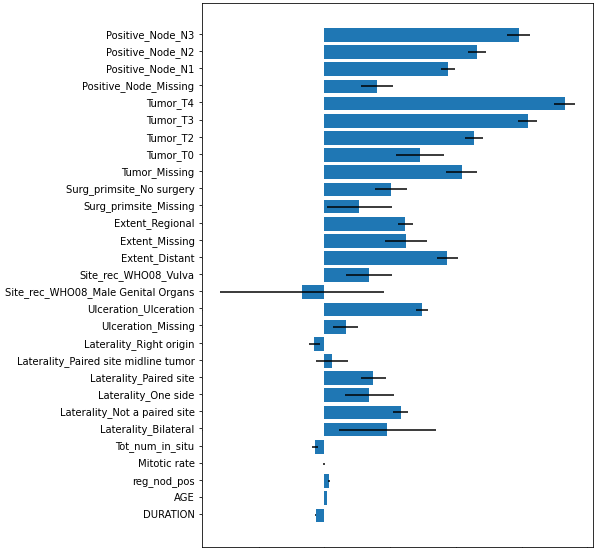}
    \caption{Coefficients and their standard errors (black lines), model without protected variables}
    \label{fig:seer coeff without A}
\end{figure}

\begin{figure}[H]
    \centering
    \includegraphics[scale=0.6]{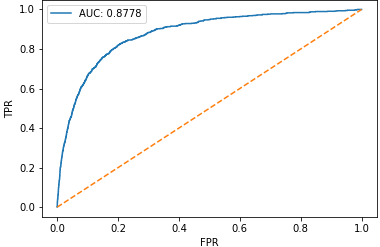}
    \caption{ROC curve for the model without the protected variables}
    \label{fig:ROC without A}
\end{figure}

\paragraph{Performance evaluation} Figure~\ref{fig:ROC without A} gives the ROC curve. The model with all selected variables had an AUC of 0.8769, and without protected variables the model has an AUC of 0.8778, which is slightly higher. \par

\paragraph{Fairness evaluation}
\begin{itemize}
    \item Globally \begin{itemize}
        \item[] The acceptance rate, true positive rate and false positive rate have slightly decreased compared to the model using all variables.
    \end{itemize}

\begin{table}[H]
    \centering
    \begin{tabular}{ccccc}
        \multicolumn{2}{c}{With all variables} &  & \multicolumn{2}{c}{Without protected variables} \\
        \multicolumn{1}{c|}{(\%)} & Global &  & \multicolumn{1}{c|}{(\%)} & Global \\ \cline{1-2} \cline{4-5} 
        \multicolumn{1}{c|}{AR} & 99.26 &  & \multicolumn{1}{c|}{AR} & 99.20 \\
        \multicolumn{1}{c|}{TPR} & 99.35&  & \multicolumn{1}{c|}{TPR} & 99.29 \\
        \multicolumn{1}{c|}{FPR} & 87.45 &  & \multicolumn{1}{c|}{FPR} & 86.26
    \end{tabular}
    \caption{Global fairness metrics}
    \label{tab:seer metrics without A global}
\end{table}

    \item By sex \begin{itemize}
        \item The acceptance rates difference has decreased, and Female kept the higher value.
        \item The true positive rates for Female and Male are closer than in the model with all variables. The Female category still has a higher TPR.
        \item The false positive rates are slightly closer than previously. The Female group still has a higher FPR.
    \end{itemize}
    Male is still disadvantaged under all three fairness definitions, although the gaps in metrics have narrowed.

\begin{table}[H]
    \centering
    \begin{tabular}{ccccccccc}
        \multicolumn{4}{c}{With all variables} &  & \multicolumn{4}{c}{Without protected variables} \\
        \multicolumn{1}{c|}{(\%)} & Female & Male & \multicolumn{1}{|c}{Difference} &  & \multicolumn{1}{c|}{Sex} & Female & Male & \multicolumn{1}{|c}{Difference} \\ \cline{1-4} \cline{6-9} 
        \multicolumn{1}{c|}{AR} & 99.67 & 98.91 & \multicolumn{1}{|c}{0.76} &  & \multicolumn{1}{c|}{AR} & 99.46 & 98.97 & \multicolumn{1}{|c}{0.49} \\
        \multicolumn{1}{c|}{TPR} & 99.71 & 99.03 & \multicolumn{1}{|c}{0.69} &  & \multicolumn{1}{c|}{TPR} & 99.52 & 99.10 & \multicolumn{1}{|c}{0.42} \\
        \multicolumn{1}{c|}{FPR} & 90.18 & 86.30 & \multicolumn{1}{|c}{3.87} &  & \multicolumn{1}{c|}{FPR} & 88.76 & 84.97 & \multicolumn{1}{|c}{3.79}
    \end{tabular}
    \caption{Fairness metrics by sex}
    \label{tab:seer metrics without A sex}
\end{table}

    \item By origin \begin{itemize}
        \item The lowest acceptance rate has increased and the highest has stayed at 100\%, for Missing. Aside from this category, the highest AR is still for White and the lowest still for Black.
        \item Black still has the lowest true positive rate, but it has increased. Missing still has the highest TPR, at 100\%. The second highest is for American Indian/AK Native, which had a TPR of 100\% in the previous model.
        \item Apart from Missing, White had the lowest false positive rate but now Black does. American Indian/AK Native had the highest FPR but now Hispanic does.
    \end{itemize}
    Black is still the most disadvantaged group under the statistical parity and equal opportunity definitions. With the previous model, we could not conclude on the most disadvantaged group under the equalized odds definition, but here without taking into account the value for Missing, Black is the most disadvantaged group.
    
\begin{table}[H]
    \centering
    \begin{tabular}{cccccccc}
        \multicolumn{8}{c}{With all variables} \\
        \multicolumn{1}{c|}{(\%)} & White & Hispanic & Black & \begin{tabular}[c]{@{}c@{}}Asian or\\ Pacific Islander\end{tabular} & \begin{tabular}[c]{@{}c@{}}American Indian\\ /AK Native\end{tabular} & Missing & \multicolumn{1}{|c}{Var} \\ \hline
        \multicolumn{1}{c|}{AR} & 99.33 & 97.70 & 92.53 & 96.63 & 98.72 & 100.00 & \multicolumn{1}{|c}{$6.10\mathrm{e}{-4}$} \\
        \multicolumn{1}{c|}{TPR} & 99.40 & 97.94 & 93.21& 97.53 & 100.00 & 100.00 & \multicolumn{1}{|c}{$5.52\mathrm{e}{-4}$} \\
        \multicolumn{1}{c|}{FPR} & 89.20 & 97.94 & 93.21 & 97.53 & 100.00 & 0.00 & \multicolumn{1}{|c}{$8.35\mathrm{e}{-2}$} \\
         &  &  &  &  &  & \\
        \multicolumn{8}{c}{Without protected variables} \\
        \multicolumn{1}{c|}{(\%)} & White & Hispanic & Black & \begin{tabular}[c]{@{}c@{}}Asian or\\ Pacific Islander\end{tabular} & \begin{tabular}[c]{@{}c@{}}American Indian\\ /AK Native\end{tabular} & Missing & \multicolumn{1}{|c}{Var} \\ \hline
        \multicolumn{1}{c|}{AR} & 99.23 & 98.76 & 93.85 & 96.68 & 99.10 & 100.00 & \multicolumn{1}{|c}{$4.37\mathrm{e}{-4}$} \\
        \multicolumn{1}{c|}{TPR} & 99.31 & 98.88 & 95.32 & 97.05 & 99.54 & 100.00 & \multicolumn{1}{|c}{$2.71\mathrm{e}{-4}$} \\
        \multicolumn{1}{c|}{FPR} & 87.07 & 89.66 & 45.45 & 70.00 & 66.67 & 0.00 & \multicolumn{1}{|c}{$9.28\mathrm{e}{-2}$}
    \end{tabular}
    \caption{Fairness metrics by origin}
    \label{tab:seer metrics without A origin}
\end{table}

    \item By marital status \begin{itemize}
        \item Compared to the previous model, the acceptance rates are closer to each other, with the lowest value going from 95.75\% for Separated to 97.48\% for Unmarried and the highest value going from 99.99\% to 99.89\%, both for Missing.
        \item The true positive rates are also closer to each other. Widowed had the lowest TPR, but now Unmarried does. Missing still has the highest.
        \item The false positive rates are also closer to each other. Widowed still has the lowest FPR and Unmarried still has the highest. Separated had the highest with Unmarried, but now has a lower FPR.
    \end{itemize}
    For statistical parity, we are closer to fairness but the most disadvantaged group is not the same. For equal opportunity and equalized odds, we are a little closer to fairness and Widowed is still the most disadvantaged group.
    
\begin{table}[H]
    \centering
    \begin{tabular}{ccccccccc}
        \multicolumn{9}{c}{With all variables} \\
        \multicolumn{1}{c|}{(\%)} & Married & Missing & Single & Widowed & Divorced & Separated & Unmarried & \multicolumn{1}{|c}{Var} \\ \hline
        \multicolumn{1}{c|}{AR} & 99.23 & 99.99 & 98.82 & 97.29 & 97.47 & 95.75 & 98.44 & \multicolumn{1}{|c}{$1.72\mathrm{e}{-4}$} \\
        \multicolumn{1}{c|}{TPR} & 99.33 & 99.99 & 98.95 & 97.54 & 97.69 & 95.71 & 98.43 & \multicolumn{1}{|c}{$1.71\mathrm{e}{-4}$} \\
        \multicolumn{1}{c|}{FPR} & 88.16 & 97.10 & 84.30 & 78.79 & 82.19 & 100.00 & 100.00 & \multicolumn{1}{|c}{$6.76\mathrm{e}{-2}$} \\
        \multicolumn{9}{c}{ } \\
        \multicolumn{9}{c}{Without protected variables} \\
        \multicolumn{1}{c|}{(\%)} & Married & Missing & Single & Widowed & Divorced & Separated & Unmarried & \multicolumn{1}{|c}{Var} \\ \hline
        \multicolumn{1}{c|}{AR} & 99.01 & 99.89 & 99.06 & 97.74 & 98.14 & 97.75 & 97.48 & \multicolumn{1}{|c}{$6.85\mathrm{e}{-5}$} \\
        \multicolumn{1}{c|}{TPR} & 99.12 & 99.91 & 99.15 & 98.13 & 98.29 & 97.96 & 97.44 & \multicolumn{1}{|c}{$6.23\mathrm{e}{-5}$} \\
        \multicolumn{1}{c|}{FPR} & 85.43 & 92.75 & 87.96 & 78.72 & 87.32 & 85.71 & 100.00 & \multicolumn{1}{|c}{$3.78\mathrm{e}{-3}$}
    \end{tabular}
    \caption{Fairness metrics by marital status}
    \label{tab:seer metrics without A mar stat}
\end{table}
\end{itemize}

To conclude, the performance metrics have not deteriorated too much compared to the model using all variables. For Sex, the fairness is better under all three definitions and the same group remains in a disadvantage position. For Origin, we also have the same most disadvantaged group as in the model using all variables, for some metrics the fairness has improved and for others it has deteriorated. For marital status, fairness has improved but for one of the definitions the most diasadvantaged group is not the same as before.

\begin{figure}[H]
    \centering
    \includegraphics[scale=0.6]{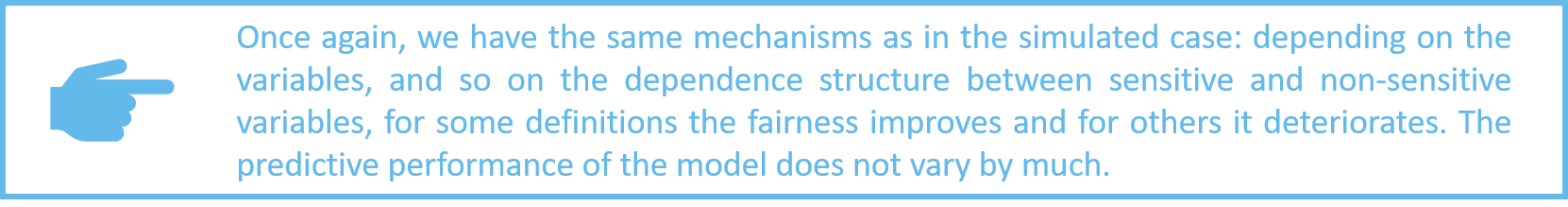}
\end{figure}

\subsection{Transforming the non-protected variables to mitigate indirect discrimination}

\paragraph{} As we have done for the simulated data, we will change the basis formed by the centered variables to obtain transformed non-sensitive explanatory variables uncorrelated with the sensitive ones. We are dealing with 12 sensitive variables, because the sensitive variables \texttt{Sex}, \texttt{Origin} and \texttt{Mar\_stat} have been converted to dummy variables. The procedure is the exact same as in the simulated case, but with more variables.

\begin{figure}[H]
    \centering
    \captionsetup{justification=centering}
    \begin{subfigure}[b]{0.45\textwidth}
        \centering
        \includegraphics[width=\textwidth]{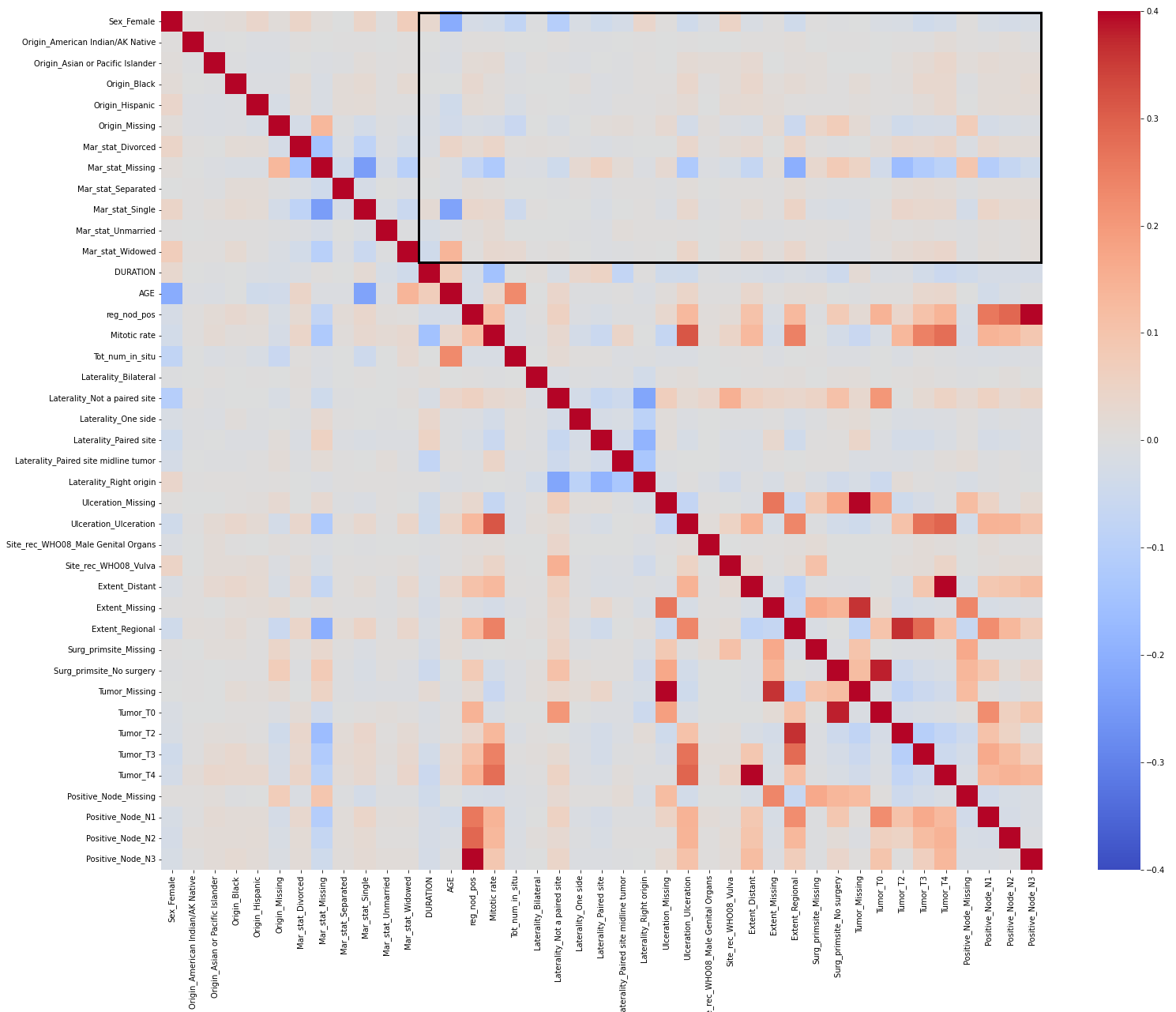}
        \caption{Original data}
    \end{subfigure}
    \begin{subfigure}[b]{0.45\textwidth}
        \centering
        \includegraphics[width=\textwidth]{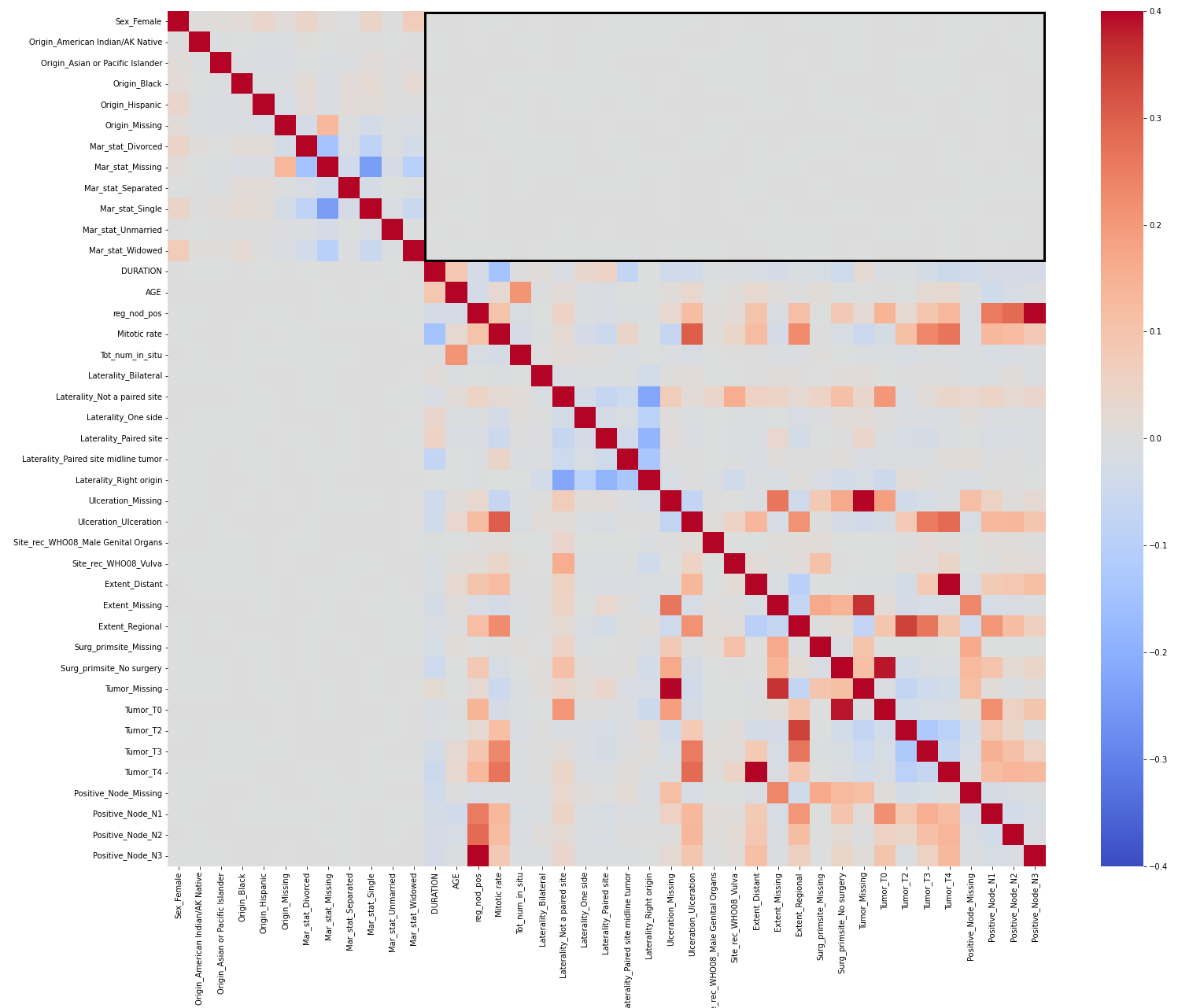}
        \caption{Transformed data}
    \end{subfigure}
    \caption{Heatmap of correlations, before and after change of basis,\\correlations between the sensitive attributes and the others framed in black}
    \label{fig:seer heatmap proj}
\end{figure}

\paragraph{Application to the data} We apply the procedure to all explanatory variables (ie not to the exposure nor the variable of interest) and obtain the projected non-sensitive variables that are uncorrelated to the sensitive ones. Figure~\ref{fig:seer heatmap proj} shows the correlation matrix before and after orthogonal projection. The correlations between the sensitive and other variables are framed in blacked. On the left, we can see a few correlations (blue and red boxes) and on the right, they are all null (only grey boxes). As expected, these results are the same as in the simulated case. 

\begin{figure}[H]
    \centering
    \includegraphics[scale=0.6]{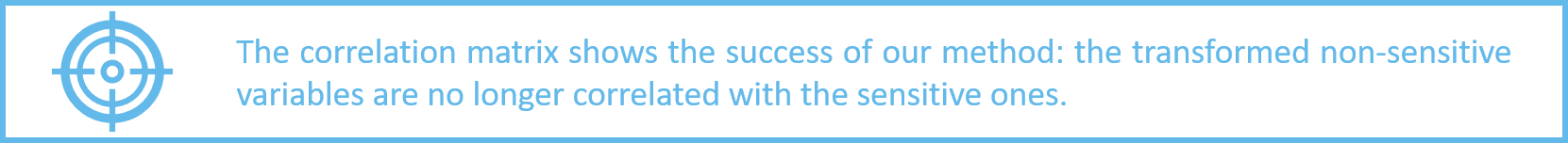}
\end{figure}

\paragraph{Applying the model} We will now apply the logistic regression model to the projected non-sensitive variables, keeping the exposure as weights, to predict the probabilities of death due to melanoma of the skin. We will compare the performance and fairness metrics of this model to the ones of the model without protected variables. The coefficients are presented in appendix~\ref{appendix:tab:coeff proj}. \par
Figure~\ref{fig:seer coeff proj} helps us visualize the coefficients and their standard errors. The most noticeable difference with the coefficients of the model without protected variables concerns \texttt{Site\_rec\_WHO08\_Male Genital Organs} which still has a large standard error but a positive coefficient instead of a negative one. We have the oppposite effect for \texttt{Laterality\_Paired site midline tumor}. \par

\begin{figure}[H]
    \centering
    \includegraphics[scale=0.7]{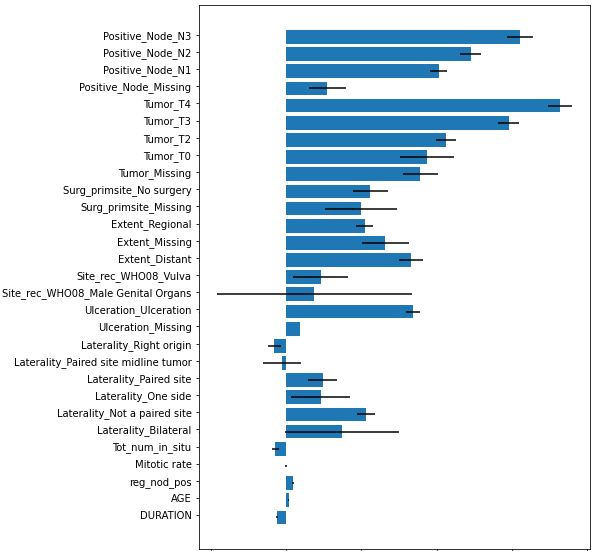}
    \caption{Coefficients and their standard errors (model with transformed variables)}
    \label{fig:seer coeff proj}
\end{figure}

\begin{figure}[H]
    \centering
    \includegraphics[scale=0.6]{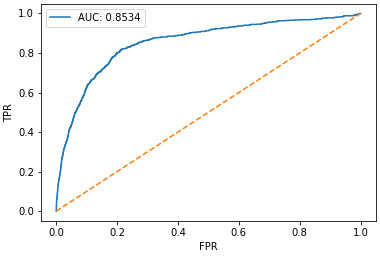}
    \caption{ROC curve for the model with transformed variables}
    \label{fig:ROC proj}
\end{figure}

\paragraph{Performance evaluation} Looking at the ROC curve in figure~\ref{fig:ROC proj}, the model performances have once again downgraded: we went from an AUC of 0.8778 (model without the protected variables) to an AUC of 0.8534, which still indicates good predictive performance. \par

\begin{figure}[H]
    \centering
    \includegraphics[scale=0.6]{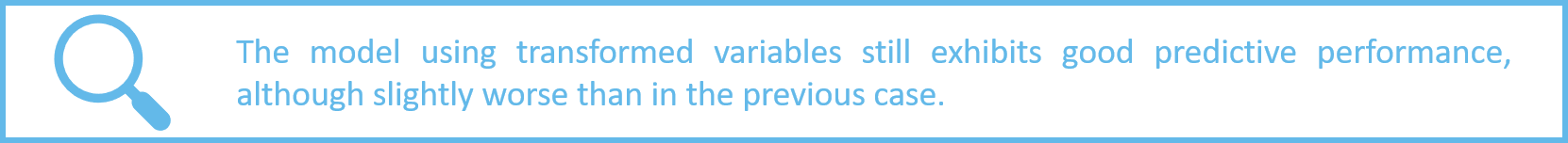}
\end{figure}

\paragraph{Fairness evaluation}
\begin{itemize}
    \item Globally, the AR, TPR and FPR are higher than for the model without protected variables. The AR and FPR are 0.1 point higher and the FPR 0.7 point higher.
    
\begin{table}[H]
    \centering
    \begin{tabular}{ccccc}
        \multicolumn{2}{c}{Without protected variables}  &  & \multicolumn{2}{c}{Transformed variables} \\
        \multicolumn{1}{c|}{(\%)} & Global  &  & \multicolumn{1}{c|}{(\%)} & Global \\ \cline{1-2} \cline{4-5} 
        \multicolumn{1}{c|}{AR} & 99.20 &  & \multicolumn{1}{c|}{AR} & 99.30 \\
        \multicolumn{1}{c|}{TPR} & 99.29 &  & \multicolumn{1}{c|}{TPR} & 99.39 \\
        \multicolumn{1}{c|}{FPR} & 86.26 &  & \multicolumn{1}{c|}{FPR} & 86.92
              
    \end{tabular}
    \caption{Global fairness metrics}
    \label{tab:seer metrics proj global}
\end{table}
    
    \item By sex \begin{itemize}
        \item The acceptance rates are now only 0.03 point apart.
        \item The true positive rates are only 0.04 point apart.
        \item The false positive rates are further apart and have changed signs.
    \end{itemize}
    We have approximately reached statistical parity when looking at the Sex variable: both groups are treated fairly by the model. We have also approximately reached equal opportunity. As the FPR are now 9.02 points apart, we are further away from the definition of equalized odds, with now the Female category being disadvantaged by the model.
\begin{table}[H]
    \centering
    \begin{tabular}{ccccccccc}
        \multicolumn{4}{c}{Without protected variables} &  & \multicolumn{4}{c}{Transformed variables} \\
        \multicolumn{1}{c|}{(\%)} & Female & Male & \multicolumn{1}{|c}{Difference} &  & \multicolumn{1}{c|}{Sex} & Female & Male & \multicolumn{1}{|c}{Difference} \\ \cline{1-4} \cline{6-9} 
        \multicolumn{1}{c|}{AR} & 99.46 & 98.97 & \multicolumn{1}{|c}{0.49} &  & \multicolumn{1}{c|}{AR} & 99.32 & 99.29 & \multicolumn{1}{|c}{0.03} \\
        \multicolumn{1}{c|}{TPR} & 99.52 & 99.10 & \multicolumn{1}{|c}{0.42} &  & \multicolumn{1}{c|}{TPR} & 99.41 & 99.37 & \multicolumn{1}{|c}{0.04} \\
        \multicolumn{1}{c|}{FPR} & 88.76 & 84.97 & \multicolumn{1}{|c}{3.79} &  & \multicolumn{1}{c|}{FPR} & 80.75 & 89.77 & \multicolumn{1}{|c}{-9.02}
    \end{tabular}
    \caption{Fairness metrics by sex}
    \label{tab:seer metrics proj sex}
\end{table}

    \item By origin \begin{itemize}
        \item The acceptances rates are now very close to each other. The order is the same except for Black and Asian that switched places (the two lowest acceptance rates). The variance in acceptances rates is of $1.82\mathrm{e}{-5}$0.
        \item The true positive rates are closer to each other than in the case without protected variables. The order is the same except for the American Indian/AK Native that went from second highest TPR to second lowest. The variance in true positive rates is of $1.23\mathrm{e}{-5}$0.
        \item The false positive rates are further apart than before, with a variance of 0.12. The order has been changed: Hispanic had the highest FPR but now has the fourth highest. Black and American Indian/AK Native respectively had the fifth and fourth highest but now have the same highest. Missing still has the lowest.
    \end{itemize}
    We have approximately reached statistical parity and equal opportunity when looking at the Origin variable, as we have very close acceptance and true positive rates. For equalized odds, we are further away from fairness as the false positive rates are further apart.

\begin{table}[H]
    \centering
    \begin{tabular}{cccccccc}
        \multicolumn{8}{c}{Without protected variables} \\
        \multicolumn{1}{c|}{(\%)} & White & Hispanic & Black & \begin{tabular}[c]{@{}c@{}}Asian or\\ Pacific Islander\end{tabular} & \begin{tabular}[c]{@{}c@{}}American Indian\\ /AK Native\end{tabular} & Missing & \multicolumn{1}{|c}{Var} \\ \hline
        \multicolumn{1}{c|}{AR} & 99.23 & 98.76 & 93.85 & 96.68 & 99.10 & 100.00 & \multicolumn{1}{|c}{$4.37\mathrm{e}{-4}$} \\
        \multicolumn{1}{c|}{TPR} & 99.31 & 98.88 & 95.32 & 97.05 & 99.54 & 100.00 & \multicolumn{1}{|c}{$2.71\mathrm{e}{-4}$} \\
        \multicolumn{1}{c|}{FPR} & 87.07 & 89.66 & 45.45 & 70.00 & 66.67 & 0.00 & \multicolumn{1}{|c}{$9.28\mathrm{e}{-2}$} \\
         &  &  &  &  &  &  &  \\
        \multicolumn{8}{c}{Transformed variables} \\
        \multicolumn{1}{c|}{(\%)} & White & Hispanic & Black & \begin{tabular}[c]{@{}c@{}}Asian or\\ Pacific Islander\end{tabular} & \begin{tabular}[c]{@{}c@{}}American Indian\\ /AK Native\end{tabular} & Missing & \multicolumn{1}{|c}{Var} \\ \hline
        \multicolumn{1}{c|}{AR} & 99.30 & 98.93 & 98.88 & 98.62 & 99.04 & 99.96 & \multicolumn{1}{|c}{$1.82\mathrm{e}{-5}$} \\
        \multicolumn{1}{c|}{TPR} & 99.39 & 99.18 & 98.85 & 99.16 & 99.02 & 99.96 & \multicolumn{1}{|c}{$1.23\mathrm{e}{-5}$} \\
        \multicolumn{1}{c|}{FPR} & 87.45 & 76.92 & 100.00 & 63.64 & 100.00 & 0.00 & \multicolumn{1}{|c}{$1.18\mathrm{e}{-1}$}
    \end{tabular}
    \caption{Fairness metrics by origin}
    \label{tab:seer metrics proj race}
\end{table}

    \item By marital status \begin{itemize}
        \item The acceptance rates are now almost equal, with a variance of $1.83\mathrm{e}{-5}$0.
        \item The same goes for the true positive rates, with a variance of $1.70\mathrm{e}{-5}$0.
        \item The false positive rates are a lot further apart.
    \end{itemize}
    We have approximately reached statistical parity and equal opportunity. For equalized odds, we are further away.
\end{itemize}

\begin{table}[H]
    \centering
    \begin{tabular}{ccccccccc}
        \multicolumn{9}{c}{Without protected variables} \\
        \multicolumn{1}{c|}{(\%)} & Married & Missing & Single & Widowed & Divorced & Separated & Unmarried & \multicolumn{1}{|c}{Var} \\ \hline
        \multicolumn{1}{c|}{AR} & 99.01 & 99.89 & 99.06 & 97.74 & 98.14 & 97.75 & 97.48 & \multicolumn{1}{|c}{$6.85\mathrm{e}{-5}$} \\
        \multicolumn{1}{c|}{TPR} & 99.12 & 99.91 & 99.15 & 98.13 & 98.29 & 97.96 & 97.44 & \multicolumn{1}{|c}{$6.23\mathrm{e}{-5}$} \\
        \multicolumn{1}{c|}{FPR} & 85.43 & 92.75 & 87.96 & 78.72 & 87.32 & 85.71 & 100.00 & \multicolumn{1}{|c}{$3.78\mathrm{e}{-3}$} \\
         &  &  &  &  &  &  &  &  \\
        \multicolumn{9}{c}{With transformed variables} \\
        \multicolumn{1}{c|}{(\%)} & Married & Missing & Single & Widowed & Divorced & Separated & Unmarried & \multicolumn{1}{|c}{Var} \\ \hline
        \multicolumn{1}{c|}{AR} & 99.12 & 99.77 & 99.06 & 99.19 & 99.07 & 98.40 & 98.47 & \multicolumn{1}{|c}{$1.83\mathrm{e}{-5}$} \\
        \multicolumn{1}{c|}{TPR} & 99.23 & 99.79 & 99.18 & 99.42 & 99.21 & 98.66 & 98.47 & \multicolumn{1}{|c}{$1.70\mathrm{e}{-5}$} \\
        \multicolumn{1}{c|}{FPR} & 86.13 & 92.86 & 84.40 & 86.96 & 90.12 & 75.00 & 0.00 & \multicolumn{1}{|c}{$9.31\mathrm{e}{-2}$}
    \end{tabular}
    \caption{Fairness metrics by marital status}
    \label{tab:seer metrics proj mar stat}
\end{table}

\begin{figure}[H]
    \centering
    \includegraphics[scale=0.6]{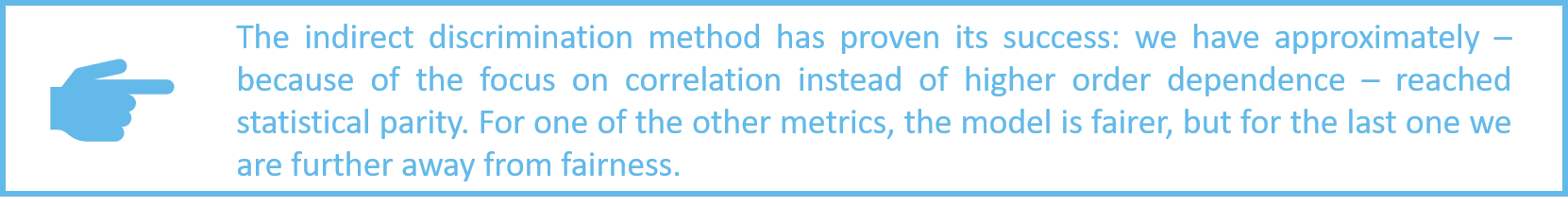}
\end{figure}

\subsection{Conclusion on the methods}

\paragraph{} Just like in the simulated case, we compared the performance and fairness of the logistic regression model using three different types of explanatory variables: all variables, only non-sensitive variables, and transformed non-sensitive variables. Visual results by sensitive variables can be found in appendices~\ref{appendix:fig:metrics by sex seer}, \ref{appendix:fig:metrics by origin seer} and \ref{appendix:fig:metrics by marital status seer}. \par
The model using all variables has the best results in terms of AUC, with only non-sensitive variables we have a decrease in AUC and finally with transformed non-sensitive variables the AUC is the lowest. This all shows the trade-off between performance and fairness. \par
In terms of fairness, the model using all variables treats unfairly the different protected groups. By Sex, Male is disadvantaged under all three definitions of fairness. By Origin, Black is the most disadvantaged group under the statistical parity and equal opportunity definitions. For equalized odds, we cannot conclude. By Marital status, Separated is the most disadvantaged category under the statistical parity definition and Widowed under the two others. \par
When we remove the protected variables, thus avoiding direct discrimination, the results depend on the relationships the protected variables have with the other variables. By sex, fairness improves under all definitions, but Male is still disadvantaged. By Origin, fairness improves under the statistical parity and equal opportunity definitions, with Black still being the most disadvantaged group. Under the equalized odds definition, one involved metric improves but the other does not. By Marital status, fairness improves under all definitions. For statistical parity, Unmarried is now the most disadvantaged group instead of Separated. For the other definitions, Widowed is still the most disadvantaged group. \par
Finally, applying the change of basis method transformed the non-sensitive variables and we now have a null correlation with the sensitive variables. Applying the model, we have the expected results: we are now very close to respecting statistical parity. Looking at the two other definitions, we are closer to equal opportunity. Looking at the metrics for equalized odds, the fairness is worse, so we can conclude that all fairness definitions are not compatible with each other. \par

\newpage
{\Large \textbf{Conclusion}}
\section*{\centering Conclusion}
\addcontentsline{toc}{section}{Conclusion}
\vspace{4cm}
\paragraph{} The goal of this thesis was to mitigate indirect discrimination when using Machine Learning models for insurance mortality data. We have provided the reader with an innovative method based on mathematical concepts of linear algebra to achieve this goal. Looking first at a simulated case allowed us to grasp the intricate effects of discrimination mechanisms, and gave us insights as to how to tackle the real use case. \par
We have made a few approximations, as reminded throughout the thesis. Firstly, we have chosen to focus on only one definition of fairness, statistical parity. Although it is the most popular in research articles, we are not certain that it will be the one chosen by regulators and policy makers in their control for fairness. Secondly, statistical parity is an independence condition, and we have approximated dependence with its first-order component, correlation. Our method is therefore a first step towards fairness, but might not be sufficient if variables have intricate and complex relationships with each other. Although, as we have mentioned, if regulators decide to use such a metric to control fairness, they will need to take into account acceptability threshold around the goal value of the metric, because of its statistical nature. Lastly, with our transformation of the non-sensitive variables, we have interpretability issues as the new vectors are a combination of the old ones. Explaining to an insured or even to other less technical employees of the insurance company, why premiums vary may be tricky as they are based on the mortality rates that come from the use of the transformed variables. \par
Further developments around this subject should thus focus on expanding the method to all forms of dependence instead of correlation only, and even looking at ways of adapting the method to other fairness definitions. Other improvements may include adaptation to regression problems, which are also encountered by insurers. \par

\clearpage

\newpage
\appendix

\section{Reminder on confidence intervals}
\label{appendix:confidence intervals}
\paragraph{} As fairness metrics are statistical, confidence intervals have to be taken into account. A legal decision revolving around a metric cannot overlook this aspect and if a value is given as a fairness standard, there has to be some interval within which is it acceptable to be.\par
Suppose we have a sample $(X_1,\dots,X_n)$ following a normal distribution $\mathcal{N}(\mu,\sigma^2)$0. The goal is to provide a confidence interval for $\mu$0. With $\hat{\mu}$ and $\hat{\sigma}$ the unbiased mean and standard deviation estimators:
\begin{equation*}
    \begin{split}
        \hat{\mu} & =\overline{X}=\frac{1}{n}\sum_{i=1}^n X_i \\
        \hat{\sigma} & =\sqrt{\frac{1}{n-1}\sum_{i=1}^n (X_i-\overline{X})^2}
    \end{split}
\end{equation*}
Then $T=\sqrt{n}\frac{\hat{\mu}-\mu}{\hat{\sigma}}$ follows a Student law with (n-1) degrees of freedom. Choosing the quantile $t$ such that $-t<T<t$, we can write
\begin{equation*}
    \hat{\mu}-\frac{t\hat{\sigma}}{\sqrt{n}} <\mu <\hat{\mu}+\frac{t\hat{\sigma}}{\sqrt{n}}
\end{equation*}
\begin{proof}
\begin{equation*}
    \begin{split}
        \mathbb{P}(-t< T<t) & =\mathbb{P}(T<t)-\mathbb{P}(T<-t) \\
         & \mathbb{P}T<t-\mathbb{P}(T>t) \\
         & =2\mathbb{P}(T<t)-1
    \end{split}
\end{equation*}
So
\begin{equation*}
    \mathbb{P}(-t_{\alpha/2}<T<t_{\alpha/2})=2(1-\alpha/2)-1=1-\alpha
\end{equation*}
So the confidence interval is
\begin{equation*}
    -t_{\alpha/2}<T<t_{\alpha/2} \iff \hat{\mu}-t_{\alpha/2}\frac{\hat{\sigma}}{\sqrt{n}}<\mu<\hat{\mu}+t_{\alpha/2}\frac{\hat{\sigma}}{\sqrt{n}}
\end{equation*}
\end{proof}
The formula can be approximated thanks to the asymptotic distribution of $\hat{\mu}$ which is normally distributed: $\hat{\mu}\sim\mathcal{N}(\mu,\sigma^2/n)$0. Therefore,
\begin{equation*}
    \hat{\mu}-|z_{\alpha/2}|\frac{\hat{\sigma}}{\sqrt{n}} <\mu <\hat{\mu}+|z_{\alpha/2}|\frac{\hat{\sigma}}{\sqrt{n}}
\end{equation*}
with $z_{\alpha/2}$ the standard normal quantile of level $\alpha/2$0. For $\alpha=5\%,|z_{2.5\%}|=1.96$0.

\newpage

\section{The Gram-Schmidt process}
\label{appendix:gram schmidt}
\paragraph{} The Gram-Schmidt process \cite{horn13} is a method for orthonormalizing a set of vectors in an inner product space. The process takes a finite and linearly independent set of vectors $(u_1,\dots,u_n)$ and produces a set of orthogonal vectors $(v_1,\dots,v_n)$0. The process works as follows:
\begin{equation*}
    \begin{split}
        v_1 & =u_1 \\
        v_2 & =u_2-\frac{\langle u_2,v_1 \rangle}{\langle v_1,v_1 \rangle}v_1  \\
        \dots & \\
        v_n & =u_n-\sum_{k=1}^{n-1} \frac{\langle u_n,v_k \rangle}{\langle v_k,v_k \rangle}v_k
    \end{split}
\end{equation*} \par
A first idea was therefore to use this process. The issue with this method is that we obtain a transformation of all variables, both sensitive and non-sensitive, except for the first one as $v_1=u_1$, and they are all orthogonal to each other, which is not what we are looking for.

\newpage

\section{Variable description (pseudo database)}
\label{appendix:tab:variable description discretized}

\begin{tabular}{ll}
    \texttt{AGE}: & Age of individual in time interval \\
     & Numerical (int) \\
     & 18-80 \\
\end{tabular} \\
\\
\begin{tabular}{ll}
    \texttt{Age\_dx}: & Age of individual at diagnosis \\
     & Numerical (int) \\
     & 18-80 \\
\end{tabular} \\
\\
\begin{tabular}{ll}
    \texttt{AYA\_site\_rec20}: & Type of melanoma \\
     & Categorical \\
     & Nodular melanoma, Superficial spreading/low cumulative sun damage \\
     & melanoma, Other malignant \\
\end{tabular} \\
\\
\begin{tabular}{ll}
    \texttt{DURATION}: & Time since diagnosis (in years) \\
     & Numerical (int) \\
     & $\mathbb{N}$ \\
\end{tabular} \\
\\
\begin{tabular}{ll}
    \texttt{Death\_skin}: & Individual died of skin melanoma in time interval \\
     & Numerical (int) \\
     & $\{0, 1\}$ \\
\end{tabular} \\
\\
\begin{tabular}{ll}
    \texttt{EXPO}: & Exposure of individual in time interval \\
     & Numerical (float) \\
     & $[0,1]$ \\
\end{tabular} \\
\\
\begin{tabular}{ll}
    \texttt{Extent}: & How far the tumor has spread at diagnosis \\
     & Categorical \\
     & In situ, Localized, Regional, Distant, Missing \\
\end{tabular} \\
\\
\begin{tabular}{ll}
    \texttt{Laterality} & Side of the body tumor originated on \\
     & Categorical \\
     & Right origin, Left origin, Bilateral, \dots \\
\end{tabular} \\
\\
\begin{tabular}{ll}
    \texttt{Mar\_stat} & Marital status of individual at diagnosis \\
     & Categorical \\
     & Married, Single, Divorced, \dots \\
\end{tabular} \\
\\
\begin{tabular}{ll}
    \texttt{Metastasis}: & Metastatic state at diagnosis \\
     & Categorical \\
     & M0, M1 \\
\end{tabular} \\
\\
\begin{tabular}{ll}
    \texttt{Mitotic rate}: & Number of mitoses per mm$^2$ at diagnosis \\
     & Numerical (int) \\
     & 0-11 \\
\end{tabular} \\
\\
\begin{tabular}{ll}
    \texttt{Positive\_Node}: & Spread to regional lymph nodes at diagnosis \\
     & Categorical \\
     & N0, N1, N2, N3, Missing \\
\end{tabular} \\
\\
\begin{tabular}{ll}
    \texttt{Origin}: & Origin of individual \\
     & Categorical \\
     & Hispanic, White, Asian, \dots \\
\end{tabular} \\
\\
\begin{tabular}{ll}
    \texttt{reg\_nod\_ex}: & Number of regional lymph nodes that were removed and examined \\
     & Numerical (int) \\
     & $\mathbb{N}$ \\
\end{tabular} \\
\\
\begin{tabular}{ll}
    \texttt{reg\_nod\_pos}: & Number of regional lymph nodes that contain metastases at diagnosis \\
     & Numerical (int) \\
     & $\mathbb{N}$ \\
\end{tabular} \\
\\
\begin{tabular}{ll}
    \texttt{Sex}: & Sex of the patient \\
     & Categorical \\
     & Male, Female \\
\end{tabular} \\
\\
\begin{tabular}{ll}
    \texttt{Site\_rec\_WHO08}: & Origin site of primary tumor \\
     & Categorical \\
     & Pleura, Vulva, Male Genital Organs \\
\end{tabular} \\
\\
\begin{tabular}{ll}
    \texttt{Stage}: & Cancer stage at diagnosis \\
     & Categorical \\
     & I, II, III, IV
\end{tabular}
\\
\begin{tabular}{ll}
    \texttt{Surg\_LN}: & Surgery of lymph nodes \\
     & Categorical \\
     & Surgery, No surgery \\
\end{tabular} \\
\\
\begin{tabular}{ll}
    \texttt{Surg\_oth}: & Surgery of other sites \\
     & Categorical \\
     & Surgery, No surgery \\
\end{tabular} \\
\\
\begin{tabular}{ll}
    \texttt{Surg\_primsite}: & Surgery of primary site \\
     & Categorical \\
     & Surgery, No surgery \\
\end{tabular} \\
\\
\begin{tabular}{ll}
    \texttt{Tot\_num\_benin}: & Total number of benign tumors at diagnosis \\
     & Numerical (int) \\
     & $\mathbb{N}$ \\
\end{tabular} \\
\\
\begin{tabular}{ll}
    \texttt{Tot\_num\_in\_situ}: & Total number of in situ tumors at diagnosis \\
     & Numerical (int) \\
     & $\mathbb{N}$ \\
\end{tabular} \\
\\
\begin{tabular}{ll}
    \texttt{Tumor}: & Tumor size at diagnosis \\
     & Categorical \\
     & T0, T1, T2, T3, T4 \\
\end{tabular} \\
\\
\begin{tabular}{ll}
    \texttt{Ulceration}: & Presence of ulceration (break on the skin) at diagnosis \\
     & Categorical \\
     & No ulceration, Ulceration, Missing \\
\end{tabular} \\
\\
\begin{tabular}{ll}
    \texttt{YEAR}: & Year of time interval \\
     & Numerical (int) \\
     & 2004-2018 \\
\end{tabular} \\
\\
\begin{tabular}{ll}
    \texttt{Yr\_dx}: & Year of diagnosis \\
     & Numerical (int) \\
     & 2004-2018 \\
\end{tabular} \\

\newpage

\section{Coefficients of the model with all selected variables}
\label{appendix:tab:coeff selected}

\begin{longtable}[c]{|l|c|c|c|}
    \hline
    Variable & Coefficient & Standard error & P-value \\ \hline
    \endfirsthead
    \endhead
    const & -7.1992 & 0.118 & 0.000 \\ \hline
    DURATION & -0.0618 & 0.007 & 0.000 \\ \hline
    AGE & 0.0211 & 0.002 & 0.000 \\ \hline
    Sex\_Female & -0.3976 & 0.043 & 0.000 \\ \hline
    reg\_nod\_pos & 0.0369 & 0.007 & 0.000 \\ \hline
    Mitotic rate & 0.0043 & 0.006 & 0.494 \\ \hline
    Tot\_num\_in\_situ & -0.0540 & 0.023 & 0.020 \\ \hline
    Laterality\_Bilateral & 0.7446 & 0.331 & 0.025 \\ \hline
    Laterality\_Not a paired site & 0.5161 & 0.060 & 0.000 \\ \hline
    Laterality\_One side & 0.1908 & 0.200 & 0.339 \\ \hline
    Laterality\_Paired site & 0.3998 & 0.093 & 0.000 \\ \hline
    Laterality\_Paired site midline tumor & -0.0055 & 0.126 & 0.965 \\ \hline
    Laterality\_Right origin & -0.0595 & 0.044 & 0.175 \\ \hline
    Ulceration\_Missing & 0.0898 & 0.099 & 0.362 \\ \hline
    Ulceration\_Ulceration & 0.7611 & 0.046 & 0.000 \\ \hline
    Site\_rec\_WHO08\_Male Genital Organs & 0.1275 & 0.561 & 0.820 \\ \hline
    Site\_rec\_WHO08\_Vulva & 0.6528 & 0.176 & 0.000 \\ \hline
    Origin\_American Indian/AK Native & 0.3888 & 0.292 & 0.183 \\ \hline
    Origin\_Asian or Pacific Islander & 0.1445 & 0.156 & 0.355 \\ \hline
    Origin\_Black & 0.3856 & 0.168 & 0.021 \\ \hline
    Origin\_Hispanic & 0.1710 & 0.108 & 0.114 \\ \hline
    Origin\_Missing & -3.0668 & 1.001 & 0.002 \\ \hline
    Mar\_stat\_Divorced & 0.3712 & 0.064 & 0.000 \\ \hline
    Mar\_stat\_Missing & -0.4660 & 0.071 & 0.000 \\ \hline
    Mar\_stat\_Separated & 0.2985 & 0.225 & 0.185 \\ \hline
    Mar\_stat\_Single & 0.2562 & 0.055 & 0.000 \\ \hline
    Mar\_stat\_Unmarried & 0.0923 & 0.426 & 0.828 \\ \hline
    Mar\_stat\_Widowed & 0.3620 & 0.087 & 0.000 \\ \hline
    Extent\_Distant & 0.8813 & 0.077 & 0.000 \\ \hline
    Extent\_Missing & 0.5145 & 0.167 & 0.002 \\ \hline
    Extent\_Regional & 0.5354 & 0.055 & 0.000 \\ \hline
    Surg\_primsite\_Missing & 0.3650 & 0.251 & 0.146 \\ \hline
    Surg\_primsite\_No surgery & 0.6385 & 0.126 & 0.000 \\ \hline
    Tumor\_Missing & 1.0028 & 0.122 & 0.000 \\ \hline
    Tumor\_T0 & 0.6995 & 0.183 & 0.000 \\ \hline
    Tumor\_T2 & 1.0626 & 0.066 & 0.000 \\ \hline
    Tumor\_T3 & 1.4361 & 0.071 & 0.000 \\ \hline
    Tumor\_T4 & 1.6748 & 0.078 & 0.000 \\ \hline
    Positive\_Node\_Missing & 0.3923 & 0.133 & 0.003 \\ \hline
    Positive\_Node\_N1 & 0.8841 & 0.055 & 0.000 \\ \hline
    Positive\_Node\_N2 & 1.0769 & 0.066 & 0.000 \\ \hline
    Positive\_Node\_N3 & 1.4402 & 0.086 & 0.000 \\ \hline
    \caption{Coefficients of the model with selected variables}
    \label{tab:coeff selected}\\
\end{longtable}

\newpage
\section{Coefficients of the model without protected variables}
\label{appendix:tab:coeff without A}

\begin{longtable}[c]{|l|c|c|c|}
    \hline
    Variable & Coefficient & Standard error & P-value \\ \hline
    \endfirsthead
    \endhead
    const & -7.4933 & 0.108 & 0.000 \\ \hline
    DURATION & -0.0662 & 0.007 & 0.000 \\ \hline
    AGE & 0.0225 & 0.002 & 0.000 \\ \hline
    reg\_nod\_pos & 0.0384 & 0.007 & 0.000 \\ \hline
    Mitotic rate & -0.0019 & 0.006 & 0.761 \\ \hline
    Tot\_num\_in\_situ & -0.0680 & 0.023 & 0.004 \\ \hline
    Laterality\_Bilateral & 0.4795 & 0.369 & 0.194 \\ \hline
    Laterality\_Not a paired site & 0.5809 & 0.059 & 0.000 \\ \hline
    Laterality\_One side & 0.3412 & 0.187 & 0.067 \\ \hline
    Laterality\_Paired site & 0.3723 & 0.093 & 0.000 \\ \hline
    Laterality\_Paired site midline tumor & 0.0578 & 0.124 & 0.642 \\ \hline
    Laterality\_Right origin & -0.0745 & 0.044 & 0.088 \\ \hline
    Ulceration\_Missing & 0.1616 & 0.095 & 0.090 \\ \hline
    Ulceration\_Ulceration & 0.7454 & 0.046 & 0.000 \\ \hline
    Site\_rec\_WHO08\_Male Genital Organs & -0.1704 & 0.624 & 0.785 \\ \hline
    Site\_rec\_WHO08\_Vulva & 0.3408 & 0.176 & 0.053 \\ \hline
    Extent\_Distant & 0.9361 & 0.078 & 0.000 \\ \hline
    Extent\_Missing & 0.6202 & 0.157 & 0.000 \\ \hline
    Extent\_Regional & 0.6152 & 0.056 & 0.000 \\ \hline
    Surg\_primsite\_Missing & 0.2659 & 0.247 & 0.281 \\ \hline
    Surg\_primsite\_No surgery & 0.5068 & 0.122 & 0.000 \\ \hline
    Tumor\_Missing & 1.0455 & 0.118 & 0.000 \\ \hline
    Tumor\_T0 & 0.7275 & 0.182 & 0.000 \\ \hline
    Tumor\_T2 & 1.1378 & 0.066 & 0.000 \\ \hline
    Tumor\_T3 & 1.5450 & 0.071 & 0.000 \\ \hline
    Tumor\_T4 & 1.8266 & 0.078 & 0.000 \\ \hline
    Positive\_Node\_Missing & 0.3997 & 0.124 & 0.001 \\ \hline
    Positive\_Node\_N1 & 0.9418 & 0.055 & 0.000 \\ \hline
    Positive\_Node\_N2 & 1.1609 & 0.066 & 0.000 \\ \hline
    Positive\_Node\_N3 & 1.4787 & 0.087 & 0.000 \\ \hline
    \caption{Coefficients of the model without protected variables}
    \label{tab:seer coef without protected} \\
\end{longtable}

\newpage

\section{Coefficients of the model with transformed variables}
\label{appendix:tab:coeff proj}

\begin{longtable}[c]{|l|c|c|c|}
    \hline
    Variable & Coefficient & Standard error & P-value \\ \hline
    \endfirsthead
    \endhead
    const & -5.7105 & 0.031 & 0.000 \\ \hline
    DURATION & -0.0617 & 0.007 & 0.000 \\ \hline
    AGE & 0.0169 & 0.002 & 0.000 \\ \hline
    reg\_nod\_pos & 0.0425 & 0.007 & 0.000 \\ \hline
    Mitotic rate & 0.0006 & 0.006 & 0.930 \\ \hline
    Tot\_num\_in\_situ & -0.0732 & 0.024 & 0.002 \\ \hline
    Laterality\_Bilateral & 0.3711 & 0.379 & 0.327 \\ \hline
    Laterality\_Not a paired site & 0.5309 & 0.061 & 0.000 \\ \hline
    Laterality\_One side & 0.2291 & 0.194 & 0.237 \\ \hline
    Laterality\_Paired site & 0.2427 & 0.095 & 0.011 \\ \hline
    Laterality\_Paired site midline tumor & -0.0276 & 0.128 & 0.829 \\ \hline
    Laterality\_Right origin & -0.0791 & 0.043 & 0.067 \\ \hline
    Ulceration\_Missing & 0.0913 & 0.097 & 0.348 \\ \hline
    Ulceration\_Ulceration & 0.8420 & 0.047 & 0.000 \\ \hline
    Site\_rec\_WHO08\_Male Genital Organs & 0.1864 & 0.648 & 0.774 \\ \hline
    Site\_rec\_WHO08\_Vulva & 0.2293 & 0.182 & 0.207 \\ \hline
    Extent\_Distant & 0.8314 & 0.079 & 0.000 \\ \hline
    Extent\_Missing & 0.6594 & 0.155 & 0.000 \\ \hline
    Extent\_Regional & 0.5229 & 0.056 & 0.000 \\ \hline
    Surg\_primsite\_Missing & 0.4959 & 0.237 & 0.037 \\ \hline
    Surg\_primsite\_No surgery & 0.5580 & 0.117 & 0.000 \\ \hline
    Tumor\_Missing & 0.8921 & 0.118 & 0.000 \\ \hline
    Tumor\_T0 & 0.9362 & 0.178 & 0.000 \\ \hline
    Tumor\_T2 & 1.0608 & 0.066 & 0.000 \\ \hline
    Tumor\_T3 & 1.4789 & 0.071 & 0.000 \\ \hline
    Tumor\_T4 & 1.8200 & 0.078 & 0.000 \\ \hline
    Positive\_Node\_Missing & 0.2745 & 0.122 & 0.024 \\ \hline
    Positive\_Node\_N1 & 1.0127 & 0.056 & 0.000 \\ \hline
    Positive\_Node\_N2 & 1.2251 & 0.067 & 0.000 \\ \hline
    Positive\_Node\_N3 & 1.5559 & 0.087 & 0.000 \\ \hline
    \caption{Coefficients of the model with transformed variables}
    \label{tab:seer proj coef} \\
\end{longtable}

\newpage
\section{Fairness metrics by sex}
\label{appendix:fig:metrics by sex seer}

\begin{figure}[H]
    \centering
    \begin{subfigure}[b]{0.3\textwidth}
        \centering
        \includegraphics[width=0.6\textwidth]{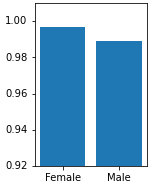}
        \caption{All variables}
    \end{subfigure}
    \begin{subfigure}[b]{0.3\textwidth}
        \centering
        \includegraphics[width=0.6\textwidth]{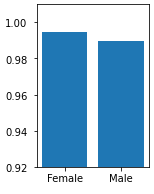}
        \caption{Without protected}
    \end{subfigure}
    \begin{subfigure}[b]{0.3\textwidth}
        \centering
        \includegraphics[width=0.6\textwidth]{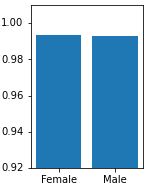}
        \caption{Transformed variables}
    \end{subfigure}
    \caption{Acceptance rates $\frac{\mbox{TP}+\mbox{FP}}{\mbox{TP}+\mbox{TN}+\mbox{FP}+\mbox{FN}}$ by sex}
    \label{fig:seer AR sex}
\end{figure}

\begin{figure}[H]
    \centering
    \begin{subfigure}[b]{0.3\textwidth}
        \centering
        \includegraphics[width=0.6\textwidth]{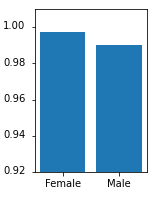}
        \caption{All variables}
    \end{subfigure}
    \begin{subfigure}[b]{0.3\textwidth}
        \centering
        \includegraphics[width=0.6\textwidth]{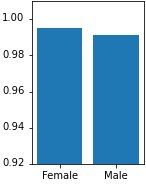}
        \caption{Without protected}
    \end{subfigure}
    \begin{subfigure}[b]{0.3\textwidth}
        \centering
        \includegraphics[width=0.6\textwidth]{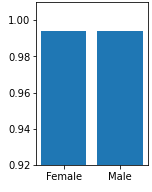}
        \caption{Transformed variables}
    \end{subfigure}
    \caption{True positive rates $\frac{\mbox{TP}}{\mbox{TP}+\mbox{FN}}$ by sex}
    \label{fig:seer TPR sex}
\end{figure}

\begin{figure}[H]
    \centering
    \begin{subfigure}[b]{0.3\textwidth}
        \centering
        \includegraphics[width=0.6\textwidth]{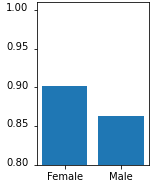}
        \caption{All variables}
    \end{subfigure}
    \begin{subfigure}[b]{0.3\textwidth}
        \centering
        \includegraphics[width=0.6\textwidth]{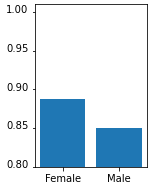}
        \caption{Without protected}
    \end{subfigure}
    \begin{subfigure}[b]{0.3\textwidth}
        \centering
        \includegraphics[width=0.6\textwidth]{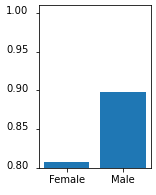}
        \caption{Transformed variables}
    \end{subfigure}
    \caption{False positive rates $\frac{\mbox{FP}}{\mbox{FP}+\mbox{TN}}$ by sex}
    \label{fig:seer FPR sex}
\end{figure}

\newpage

\section{Fairness metrics by origin}
\label{appendix:fig:metrics by origin seer}

\begin{figure}[H]
    \centering
    \begin{subfigure}[b]{0.3\textwidth}
        \centering
        \includegraphics[width=\textwidth]{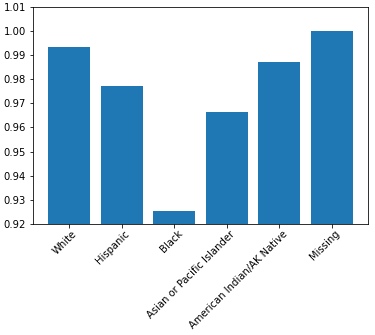}
        \caption{All variables}
    \end{subfigure}
    \begin{subfigure}[b]{0.3\textwidth}
        \centering
        \includegraphics[width=\textwidth]{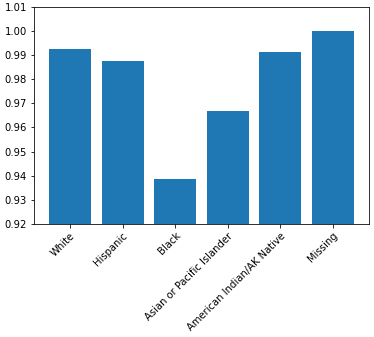}
        \caption{Without protected}
    \end{subfigure}
    \begin{subfigure}[b]{0.3\textwidth}
        \centering
        \includegraphics[width=\textwidth]{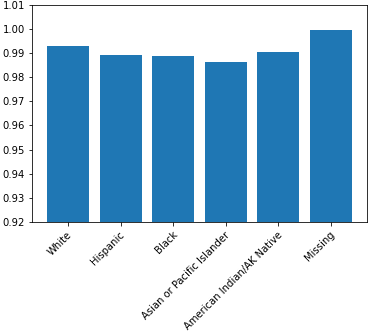}
        \caption{Transformed variables}
    \end{subfigure}
    \caption{Acceptance rates $\frac{\mbox{TP}+\mbox{FP}}{\mbox{TP}+\mbox{TN}+\mbox{FP}+\mbox{FN}}$ by origin}
    \label{fig:seer AR origin}
\end{figure}

\begin{figure}[H]
    \centering
    \begin{subfigure}[b]{0.3\textwidth}
        \centering
        \includegraphics[width=\textwidth]{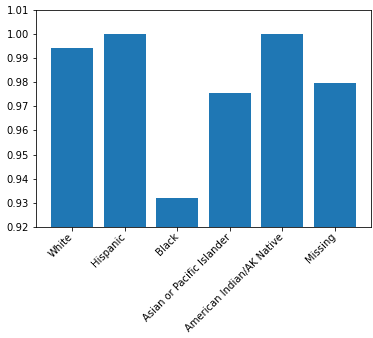}
        \caption{All variables}
    \end{subfigure}
    \begin{subfigure}[b]{0.3\textwidth}
        \centering
        \includegraphics[width=\textwidth]{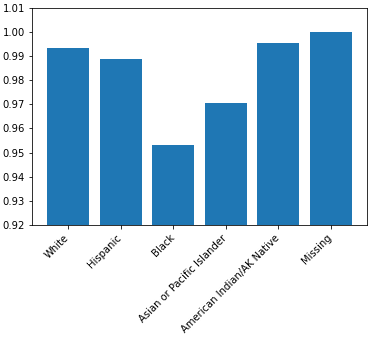}
        \caption{Without protected}
    \end{subfigure}
    \begin{subfigure}[b]{0.3\textwidth}
        \centering
        \includegraphics[width=\textwidth]{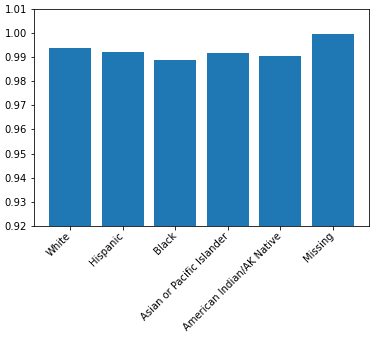}
        \caption{Transformed variables}
    \end{subfigure}
    \caption{True positive rates $\frac{\mbox{TP}}{\mbox{TP}+\mbox{FN}}$ by origin}
    \label{fig:seer TPR origin}
\end{figure}

\begin{figure}[H]
    \centering
    \begin{subfigure}[b]{0.3\textwidth}
        \centering
        \includegraphics[width=\textwidth]{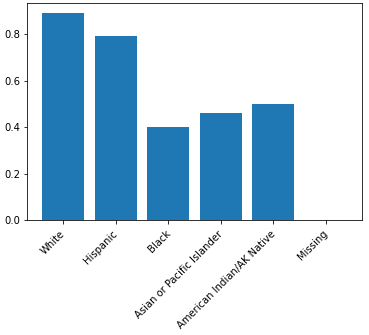}
        \caption{All variables}
    \end{subfigure}
    \begin{subfigure}[b]{0.3\textwidth}
        \centering
        \includegraphics[width=\textwidth]{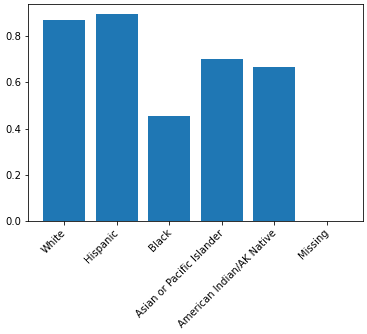}
        \caption{Without protected}
    \end{subfigure}
    \begin{subfigure}[b]{0.3\textwidth}
        \centering
        \includegraphics[width=\textwidth]{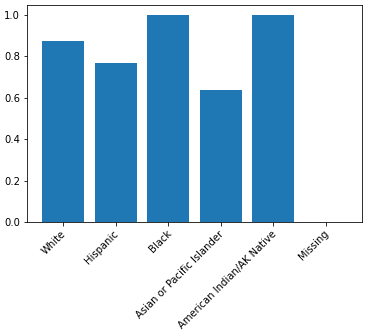}
        \caption{Transformed variables}
    \end{subfigure}
    \caption{False positive rates $\frac{\mbox{FP}}{\mbox{FP}+\mbox{TN}}$ by origin}
    \label{fig:seer FPR origin}
\end{figure}

\newpage
\section{Fairness metrics by marital status}
\label{appendix:fig:metrics by marital status seer}

\begin{figure}[H]
    \centering
    \begin{subfigure}[b]{0.3\textwidth}
        \centering
        \includegraphics[width=\textwidth]{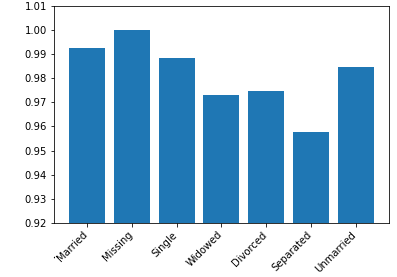}
        \caption{All variables}
    \end{subfigure}
    \begin{subfigure}[b]{0.3\textwidth}
        \centering
        \includegraphics[width=\textwidth]{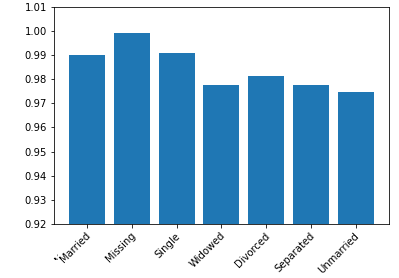}
        \caption{Without protected}
    \end{subfigure}
    \begin{subfigure}[b]{0.3\textwidth}
        \centering
        \includegraphics[width=\textwidth]{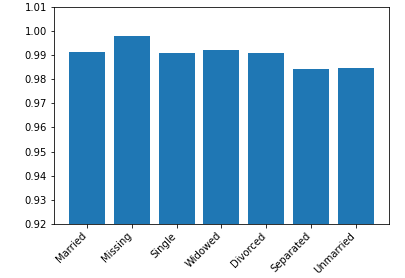}
        \caption{Transformed variables}
    \end{subfigure}
    \caption{Acceptance rates $\frac{\mbox{TP}+\mbox{FP}}{\mbox{TP}+\mbox{TN}+\mbox{FP}+\mbox{FN}}$ by marital status}
    \label{fig:seer AR}
\end{figure}

\begin{figure}[H]
    \centering
    \begin{subfigure}[b]{0.3\textwidth}
        \centering
        \includegraphics[width=\textwidth]{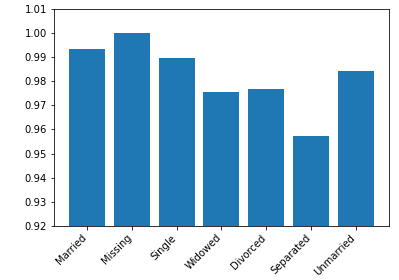}
        \caption{All variables}
    \end{subfigure}
    \begin{subfigure}[b]{0.3\textwidth}
        \centering
        \includegraphics[width=\textwidth]{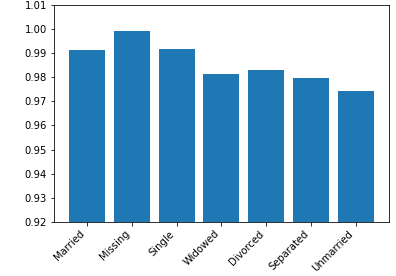}
        \caption{Without protected}
    \end{subfigure}
    \begin{subfigure}[b]{0.3\textwidth}
        \centering
        \includegraphics[width=\textwidth]{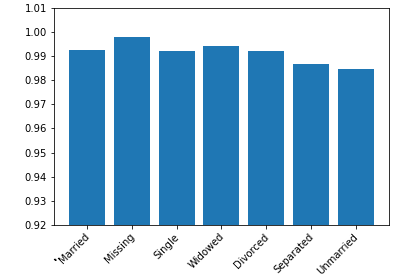}
        \caption{Transformed variables}
    \end{subfigure}
    \caption{True positive rates $\frac{\mbox{TP}}{\mbox{TP}+\mbox{FN}}$ by marital status}
    \label{fig:seer TPR}
\end{figure}

\begin{figure}[H]
    \centering
    \begin{subfigure}[b]{0.3\textwidth}
        \centering
        \includegraphics[width=\textwidth]{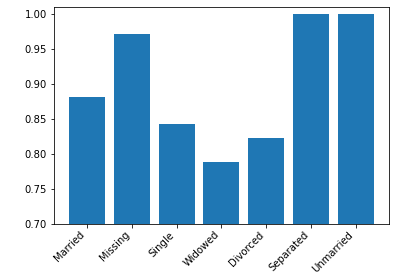}
        \caption{All variables}
    \end{subfigure}
    \begin{subfigure}[b]{0.3\textwidth}
        \centering
        \includegraphics[width=\textwidth]{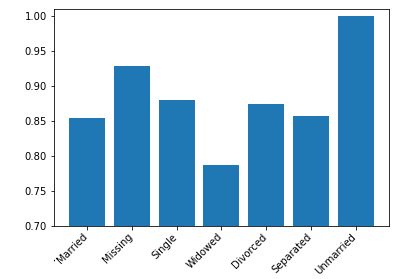}
        \caption{Without protected}
    \end{subfigure}
    \begin{subfigure}[b]{0.3\textwidth}
        \centering
        \includegraphics[width=\textwidth]{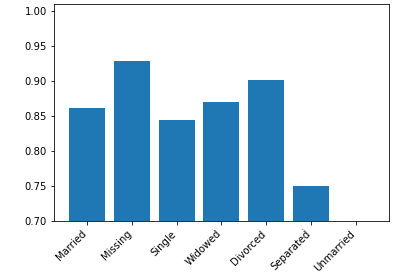}
        \caption{Transformed variables}
    \end{subfigure}
    \caption{False positive rates $\frac{\mbox{FP}}{\mbox{FP}+\mbox{TN}}$ by marital status}
    \label{fig:seer FPR}
\end{figure}


\nocite{*}
\begin{thebibliography}{57}
\providecommand{\natexlab}[1]{#1}
\providecommand{\url}[1]{\texttt{#1}}
\expandafter\ifx\csname urlstyle\endcsname\relax
  \providecommand{\doi}[1]{doi: #1}\else
  \providecommand{\doi}{doi: \begingroup \urlstyle{rm}\Url}\fi

\bibitem[civ()]{civil_code}
French civil code.
\newblock Légifrance.
\newblock URL
  \url{https://www.legifrance.gouv.fr/codes/texte_lc/LEGITEXT000006070721}.
\newblock [Online; accessed July 7, 2022].

\bibitem[ins()]{ins_code}
French insurance code.
\newblock Légifrance.
\newblock URL
  \url{https://www.legifrance.gouv.fr/codes/id/LEGITEXT000006073984/}.
\newblock [Online; accessed July 7, 2022].

\bibitem[pen()]{penal_code}
French penal code.
\newblock Légifrance.
\newblock URL
  \url{https://www.legifrance.gouv.fr/codes/id/LEGITEXT000006070719/}.
\newblock [Online; accessed July 6, 2022].

\bibitem[sen()]{senatebill}
Senate bill 21-169.
\newblock URL
  \url{https://www.leg.colorado.gov/sites/default/files/2021a_169_signed.pdf}.
\newblock signed on July 6, 2021.

\bibitem[ACPR(2022)]{acpr_govAI}
ACPR.
\newblock Governance of artificial intelligence in finance, 06 2022.
\newblock Discussion papers publication.

\bibitem[Alves et~al.(2022)Alves, Bernier, Couceiro, Makhlouf, Palamidessi, and
  Zhioua]{alves22}
G.~Alves, F.~Bernier, M.~Couceiro, K.~Makhlouf, C.~Palamidessi, and S.~Zhioua.
\newblock Survey on fairness notions and related tensions.
\newblock 06 2022.

\bibitem[Avraham et~al.(2013)Avraham, Logue, and Schwarcz]{avraham13}
R.~Avraham, K.~Logue, and D.~Schwarcz.
\newblock Understanding insurance anti-discrimination laws.
\newblock \emph{Law \& Economics Working Papers}, 52, 2013.

\bibitem[Barocas and Selbst(2016)]{barocas16}
S.~Barocas and A.~Selbst.
\newblock Big data's disparate impact.
\newblock \emph{104 California Law Review 671}, 09 2016.

\bibitem[Barry and Charpentier(2022)]{barry22}
L.~Barry and A.~Charpentier.
\newblock The fairness of machine learning in insurance: New rags for an old
  man?
\newblock \emph{arXiv e-prints}, 05 2022.

\bibitem[Bellamy et~al.(2018)Bellamy, Dey, Hind, S.Hoffman, Houde, Kanna,
  Lohia, Martino, Mehta, Mojsilovic, Nagar, Ramamurthy, Richards, Saha,
  Sattigeri, Singh, Varshney, and Zhang]{AIF360}
R.~Bellamy, K.~Dey, M.~Hind, S.Hoffman, S.~Houde, K.~Kanna, P.~Lohia,
  J.~Martino, S.~Mehta, A.~Mojsilovic, S.~Nagar, K.~Natesan Ramamurthy,
  J.~Richards, D.~Saha, P.~Sattigeri, M.~Singh, K.~Varshney, and Y.~Zhang.
\newblock Ai fairness 360: An extensible toolkit for detecting, understanding,
  and mitigating unwanted algorithmic bias.
\newblock \emph{CoRR}, 10 2018.

\bibitem[Benson et~al.(2016)Benson, III, and Wikipedia]{cognitive_bias_codex}
B.~Benson, J.~Manoogian III, and Wikipedia.
\newblock The cognitive bias codex, 2016.
\newblock URL
  \url{https://fr.wikipedia.org/wiki/Fichier:The_Cognitive_Bias_Codex_-_180%2B_biases,_designed_by_John_Manoogian_III_(jm3).png}.
\newblock [Online; accessed July 5, 2022].

\bibitem[Binns(2019)]{binns19}
R.~Binns.
\newblock On the apparent conflict between individual and group fairness.
\newblock \emph{CoRR}, 12 2019.

\bibitem[Bird et~al.(2020)Bird, Dud{\'i}k, Edgar, Horn, Lutz, Milan, Sameki,
  Wallach, and Walker]{fairlearn}
S.~Bird, M.~Dud{\'i}k, R.~Edgar, B.~Horn, R.~Lutz, V.~Milan, M.~Sameki,
  H.~Wallach, and K.~Walker.
\newblock Fairlearn: A toolkit for assessing and improving fairness in {AI}.
\newblock Technical report, Microsoft, 05 2020.
\newblock URL
  \url{https://www.microsoft.com/en-us/research/publication/fairlearn-a-toolkit-for-assessing-and-improving-fairness-in-ai/}.

\bibitem[Blodgett and O'Connor(2017)]{blodgett17}
S.~Blodgett and B.~O'Connor.
\newblock Racial disparity in natural language processing: A case study of
  social media african-american english.
\newblock \emph{CoRR}, 2017.

\bibitem[Blyth(2022)]{scor22}
P.~Blyth.
\newblock Borderline malignancy / low malignant potential cancers.
\newblock \emph{SCOR Expert Views}, 10 2022.

\bibitem[Boczar et~al.(2021)Boczar, Avila, Carter, Moore, Giardi, Guliyeva,
  Bruce, McLeod, and Forte]{boczar21}
D.~Boczar, F.~Avila, R.~Carter, P.~Moore, D.~Giardi, G.~Guliyeva, C.~Bruce,
  C.~McLeod, and A.~Forte.
\newblock Using facial recognition tools for health assessment.
\newblock \emph{Plastic surgical nursing : official journal of the American
  Society of Plastic and Reconstructive Surgical Nurses}, 41:\penalty0
  112–116, 2021.

\bibitem[Boyd et~al.(2014)Boyd, Levy, and Marwick]{boyd14}
D.~Boyd, K.~Levy, and A.~Marwick.
\newblock The networked nature of algorithmic discrimination.
\newblock \emph{Collected Essays, New America}, pages 53--57, 10 2014.

\bibitem[Butucea(2022)]{butucea22}
C.~Butucea.
\newblock Introduction à l'apprentissage statistique.
\newblock ENSAE course, 2022.

\bibitem[Calvo(2022)]{calvo22}
M.~Calvo.
\newblock Comment izzy constat dépoussière le constat amiable.
\newblock \emph{L'Argus de l'Assurance}, 06 2022.

\bibitem[Charpentier(2010)]{charpentier10}
A.~Charpentier.
\newblock Copules et risques corrélés.
\newblock https://freakonometrics.github.io/teaching/, 11 2010.

\bibitem[Charpentier(2022)]{charpentier22}
A.~Charpentier.
\newblock \emph{Insurance : Discrimination, Biases \& Fairness}.
\newblock Opinions \& debates n°25 edition, 07 2022.

\bibitem[Chawla et~al.(2002)Chawla, Bowler, Hall, and Kegelmeyer]{chawla02}
N.~V. Chawla, K.~W. Bowler, L.~O. Hall, and W.~P. Kegelmeyer.
\newblock Smote: Synthetic minority over-sampling technique.
\newblock \emph{Journal Of Artificial Intelligence Research}, 16:\penalty0
  321--357, 2002.

\bibitem[Chen et~al.(2019)Chen, Kallus, Mao, Svacha, and Udell]{chen19}
J.~Chen, N.~Kallus, X.~Mao, G.~Svacha, and M.~Udell.
\newblock Fairness under unawareness.
\newblock \emph{Proceedings of the Conference on Fairness, Accountability, and
  Transparency}, 01 2019.

\bibitem[Commission()]{eu_ai_framework}
European Commission.
\newblock Regulatory framework proposal on artificial intelligence.
\newblock URL
  \url{https://digital-strategy.ec.europa.eu/en/policies/regulatory-framework-ai}.
\newblock Consulted on June 14, 2022.

\bibitem[Cottet(2022)]{cottet22}
V.~Cottet.
\newblock Statistique mathématique.
\newblock ENSAE course, 2022.

\bibitem[DARES(2020)]{dares20}
DARES.
\newblock Quelles sont les conditions d'emploi des salariés à temps partiel ?
\newblock \emph{DARES analyses}, 25, 08 2020.

\bibitem[Faure(2020)]{faure20}
S.~Faure.
\newblock Secteurs féminisés : la parité s’éloigne encore.
\newblock \emph{INSEE Analyses Centre-Val de Loire}, 10, 07 2020.

\bibitem[Fermanian(2022)]{fermanian}
J.-D. Fermanian.
\newblock Copulas and applications.
\newblock Lectures at ENSAE, 2022.

\bibitem[Frees and Huang(2021)]{frees21}
E.~Frees and F.~Huang.
\newblock The discriminating (pricing) actuary.
\newblock \emph{North American Actuarial Journal}, pages 1--23, 03 2021.

\bibitem[Gaddis(2017)]{gaddis17}
S.~Gaddis.
\newblock How black are lakisha and jamal? racial perceptions from names used
  in correspondence audit studies.
\newblock \emph{Sociological Science}, 4\penalty0 (19):\penalty0 469--489,
  2017.

\bibitem[GDPR.eu({\natexlab{a}})]{gdpr_article_17}
GDPR.eu.
\newblock Article 17 gdpr, {\natexlab{a}}.
\newblock URL \url{https://gdpr.eu/article-17-right-to-be-forgotten/}.
\newblock Consulted on April 25, 2022.

\bibitem[GDPR.eu({\natexlab{b}})]{gdpr_article_22}
GDPR.eu.
\newblock Article 22 gdpr, {\natexlab{b}}.
\newblock URL
  \url{https://gdpr.eu/article-22-automated-individual-decision-making/}.
\newblock Consulted on April 25, 2022.

\bibitem[Grari(2022)]{grari22}
V.~Grari.
\newblock \emph{Adversarial mitigation to reduce unwanted biases in machine
  learning}.
\newblock PhD thesis, 06 2022.

\bibitem[Hara et~al.(2014)Hara, Sun, Moore, Jacobs, and Froehlich]{hara14}
K.~Hara, J.~Sun, R.~Moore, D.~Jacobs, and J.~Froehlich.
\newblock Tohme: Detecting curb ramps in google street view using
  crowdsourcing, computer vision, and machine learning.
\newblock \emph{Proceedings of the 27th Annual ACM Symposium on User Interface
  Software and Technology}, page 189–204, 2014.

\bibitem[Horn and Johnson(2013)]{horn13}
R.~Horn and C.~Johnson.
\newblock \emph{Matrix Analysis, second edition}, volume 466.
\newblock Cambridge University Press, 2013.

\bibitem[Institute(2022)]{SEER}
National~Cancer Institute.
\newblock Seer cancer stat facts: Melanoma of the skin, 2022.
\newblock URL \url{https://seer.cancer.gov/statfacts/html/melan.html}.

\bibitem[Jean et~al.(2016)Jean, Burke, Xie, Davis, Lobell, and Ermon]{jean16}
N.~Jean, M.~Burke, S.~Xie, W.~Davis, D.~Lobell, and S.~Ermon.
\newblock Combining satellite imagery and machine learning to predict poverty.
\newblock \emph{Science}, 353:\penalty0 790 -- 794, 2016.

\bibitem[Kamiran and Calders(2009)]{kamiran19a}
F.~Kamiran and T.~Calders.
\newblock Classifying without discriminating.
\newblock \emph{Proceedings 2nd IEEE International Conference on Computer,
  Control and Communication (IC4 2009, Karachi, Pakistan, February 17-18,
  2009)}, pages 1--6, 2009.

\bibitem[Kamiran and Calders(2012)]{kamiran11}
F.~Kamiran and T.~Calders.
\newblock Data preprocessing techniques for classification without
  discrimination.
\newblock \emph{Knowledge and Information Systems}, 33, 10 2012.

\bibitem[Krafcheck(2021)]{krafcheck21}
E.~Krafcheck.
\newblock The insurance industry’s renewed focus on disparate impacts and
  unfair discrimination.
\newblock Milliman, 09 2021.

\bibitem[Kuschke(2012)]{test-achats}
B.~Kuschke.
\newblock Association belge des consommateurs test-achats asbl, vann van vugt,
  charles basselier v conseil des ministres, case c-236/09 ecj, gender equality
  in insurance.
\newblock \emph{De Jure Law Journal}, 2012.

\bibitem[Larson et~al.(2016)Larson, Mattu, Kirchner, and Angwin]{propublica}
J.~Larson, S.~Mattu, L.~Kirchner, and J.~Angwin.
\newblock How we analyzed the compas recidivism algorithm.
\newblock ProPublica, 05 2016.
\newblock URL
  \url{https://www.propublica.org/article/how-we-analyzed-the-compas-recidivism-algorithm}.

\bibitem[Martin and Varner(2017)]{martin17}
L.~Martin and K.~Varner.
\newblock Race, residential segregation, and the death of democracy: Education
  and myth of postracialism.
\newblock \emph{Democracy and Education}, 25\penalty0 (1), 2017.

\bibitem[Mehrabi et~al.(2019)Mehrabi, Morstatter, Saxena, and
  Galstyan]{mehrabi19}
N.~Mehrabi, F.~Morstatter, N.~Saxena, and A.~Galstyan.
\newblock A survey on bias and fairness in machine learning.
\newblock \emph{ACM Computing Surveys}, 54\penalty0 (6), 08 2019.

\bibitem[Miller(2009)]{miller09}
M.~Miller.
\newblock Disparate impact and unfairly discriminatory insurance rates.
\newblock \emph{Casualty Actuarial Society E-Forum}, 2009.

\bibitem[Molnar()]{molnar}
C.~Molnar.
\newblock \emph{Interpretable Machine Learning, A Guide for Making Black Box
  Models Explainable}.
\newblock March 2022 version.

\bibitem[Narayanan(2018)]{narayanan_yt}
A.~Narayanan.
\newblock Tutorial: 21 fairness definitions and their politics.
\newblock YouTube [online], 03 2018.
\newblock URL \url{https://www.youtube.com/watch?v=jIXIuYdnyyk}.
\newblock FAT Conference.

\bibitem[Observatory(2020)]{WHO20}
The Global~Cancer Observatory.
\newblock Melanoma of skin.
\newblock World Health Organization, 2020.
\newblock URL
  \url{https://gco.iarc.fr/today/data/factsheets/cancers/16-Melanoma-of-skin-fact-sheet.pdf}.

\bibitem[Petkantchin(2010)]{petkantchin10}
V.~Petkantchin.
\newblock Eu anti-discrimination policy's impact on insurance risk management:
  A parallel with the us sub-prime crisis.
\newblock \emph{Pensions: An International Journal}, 15:\penalty0 155--160,
  2010.

\bibitem[Reardon and Bischoff(2011)]{reardon11}
S.~Reardon and K.~Bischoff.
\newblock Income inequality and income segregation.
\newblock \emph{American Journal of Sociology}, 116\penalty0 (4):\penalty0
  1092--1153, 2011.

\bibitem[Robineau(2010)]{robineau10}
M.~Robineau.
\newblock Sélection des risques et discriminations en droit des assurances.
\newblock \emph{Les cahiers de droit de la santé}, pages 177--191, 2010.

\bibitem[Rugh and Massey(2014)]{rugh14}
J.~Rugh and D.~Massey.
\newblock Segregation in post-civil rights america: Stalled integration or end
  of the segregated century?
\newblock \emph{Du Bois Review: Social Science Research on Race}, 11\penalty0
  (2):\penalty0 205–232, 2014.

\bibitem[Schlesinger et~al.(2018)Schlesinger, O'Hara, and
  Taylor]{schlesinger18}
A.~Schlesinger, K.~O'Hara, and A.~Taylor.
\newblock Let's talk about race: Identity, chatbots, and ai.
\newblock \emph{Proceedings of the 2018 CHI Conference on Human Factors in
  Computing Systems}, page 1–14, 04 2018.

\bibitem[Squires and Chadwick(2006)]{squires06}
G.~Squires and J.~Chadwick.
\newblock Linguistic profiling a continuing tradition of discrimination in the
  home insurance industry?
\newblock \emph{Urban Affairs Review}, 41:\penalty0 400--415, 01 2006.

\bibitem[Stead(2020)]{stead20}
S.~Stead.
\newblock Disparate impact and unfair discrimination in insurance are not the
  same thing.
\newblock \emph{Federation of Regulatory Counsel Journal}, 31\penalty0 (3),
  2020.

\bibitem[UK(2020)]{TNM20}
Cancer~Research UK.
\newblock Tnm staging, 05 2020.
\newblock URL
  \url{https://www.cancerresearchuk.org/about-cancer/melanoma/stages-types/tnm-staging}.

\bibitem[Williams et~al.(2018)Williams, Brooks, and Shmargad]{williams18}
B.~Williams, C.~Brooks, and Y.~Shmargad.
\newblock How algorithms discriminate based on data they lack: Challenges,
  solutions, and policy implications.
\newblock \emph{Journal of Information Policy}, 09 2018.

\end{thebibliography}
\end{document}